\documentclass[a4paper,12pt]{scrartcl} 
\title{\cdsplit \ and \hpdsplit: 
       efficient conformal regions  in high dimensions}

\author{Rafael Izbicki,
Gilson Shimizu,
Rafael B. Stern}

\usepackage[T1]{fontenc}
\usepackage{nameref}
\usepackage{algorithm, algorithmic}
\usepackage[colorlinks=true, urlcolor=blue, citecolor=blue, linkcolor=blue]{hyperref}

\usepackage{amsfonts, amsmath, amssymb, amsthm, thmtools}
\usepackage{mathrsfs}
\usepackage{cleveref}

\declaretheorem[name=Theorem, refname={Theorem, Theorems}, Refname={Theorem, Theorems}]{theorem_2}
\declaretheorem[name=Assumption, refname={Assumption, Assumptions}, Refname={Assumption, Assumptions}, sibling=theorem_2]{assumption_2}
\declaretheorem[name=Definition, refname={Definition, Definitions}, Refname={Definition, Definitions}, sibling=theorem_2]{definition_2}
\declaretheorem[name=Example, refname={Example, Examples}, Refname={Example, Examples}, sibling=theorem_2]{example}
\declaretheorem[name=Lemma, refname={Lemma, Lemmas}, Refname={Lemma, Lemmas}, sibling=theorem_2]{lemma_2}

\crefname{section}{section}{sections}
\Crefname{section}{Section}{Sections}
\crefname{table}{table}{tables}
\Crefname{table}{Table}{Tables}

\usepackage{color, enumitem, float, fourier, layout, multirow, setspace, verbatim}
\setlist[enumerate]{leftmargin=*}
\usepackage[colorinlistoftodos]{todonotes}
\usepackage{listings}
\usepackage{caption}
\usepackage{epigraph}
\usepackage{subfig}
\usepackage{ulem}
\usepackage{makecell}

\usepackage[round]{natbib}

\usepackage{graphicx}
\graphicspath{{figures/}{../figures/}}

\let\vec\mathbf

\let\vec\mathbf

\def\C{{\vec{C}}}

\def\x{{\vec{x}}}
\def\X{{\vec{X}}}
\def\w{{\vec{w}}}

\def\D{{\mathbb D}}
\def\E{{\mathbb E}}
\def\I{{\mathbb{I}}}

\def\P{{\mathbb P}}

\def\Re{{\mathbb R}}

\def\Ca{{C_{\alpha}}}
\def\Cas{{C_{\alpha}^*}}
\def\qa{{q_\alpha}}
\def\hqa{{\hat{q}_\alpha}}

\def\hg{{\widehat{g}}}
\def\hF{{\widehat{F}}}
 
\def\hH{{\hat{H}}}

\def\sA{{\mathcal{A}}}
\def\sX{{\mathcal{X}}}
\def\sY{{\mathcal{Y}}}

\usepackage{adjustbox}
\usepackage{multirow}

\def\cdsplit{{\texttt{CD-split}}}
\def\regsplit{{\texttt{Reg-split}}}
\def\quantilesplit{{\texttt{Quantile-split}}}
\def\cdsplitp{{\texttt{CD-split}$^+$}}
\def\hpdsplit{{\texttt{HPD-split}}}
\def\flexcode{{\texttt{FlexCode}}}

\def\distsplit{{\texttt{Dist-split}}}

\def\hf{{\widehat{f}}}
\def\hF{{\widehat{F}}}

\def\figSize{{0.4\textwidth}}

\usepackage{makecell}

\renewcommand{\algorithmicrequire}{\textbf{\small Input:}}
\renewcommand{\algorithmicensure}{\textbf{\small Output:}}
\definecolor{darkgreen}{rgb}{0.3, 0.5, 0.0}
\newcommand{\codecomment}[1]{\textbf{\color{darkgreen}\ // #1}}

\usepackage{mathtools}
\usepackage{etextools}
\usepackage{ifthen}
\DeclarePairedDelimiter{\ceil}{\lceil}{\rceil}

\renewcommand{\emph}[1]{\textit{#1}}

\begin{document}

\maketitle

\begin{abstract}%
 Conformal methods create prediction bands that 
 control average coverage assuming solely i.i.d. data.
 Although the literature has mostly focused on
 prediction intervals, more general regions can
 often better represent uncertainty.
 For instance, a bimodal target is
 better represented by the union of two intervals.
 Such prediction regions are obtained by
 \cdsplit , which combines 
 the split method and 
 a data-driven partition of
 the feature space which 
 scales to high dimensions. \cdsplit \ however contains many tuning parameters, and their role is not clear.
 In this paper, we provide new insights on \cdsplit \ by exploring its  theoretical properties. In particular,
 we show that \cdsplit \ converges 
 asymptotically to the oracle 
 highest predictive density set and 
 satisfies local and asymptotic conditional validity.  
 We also present  simulations
 that show how to tune \cdsplit.
 Finally, we  introduce \hpdsplit, 
 a variation of \cdsplit \ that requires less tuning, 
 and show that it shares  the same theoretical guarantees as \cdsplit.
 In a wide variety of our simulations,
 \cdsplit \ and \hpdsplit \ have better conditional coverage
 and yield smaller prediction regions than
 other methods.
\end{abstract}


\section{Goals in Conformal Prediction}
\label{sec:goals}

Most supervised machine learning methods yield
a point estimate for a target, $Y \in \sY$,
based on features, $\X \in \sX$.
However, it is often more informative to
present prediction bands, that is,
a subset of $\sY$ with 
plausible values for $Y$ \citep{Neter1996}. 
A particular way of constructing 
prediction bands is through
\emph{conformal predictions}
\citep{Vovk2005,Vovk2009}.
Conformal predictions generate 
a predictive region for a future target, $Y_{n+1}$,
based on features, $\X_{n+1}$, and
past observations $(\X_1,Y_1),\ldots,(\X_n,Y_n)$.
An appealing property of the conformal methodology is that
it controls the \emph{marginal coverage} of
prediction bands assuming
solely exchangeable\footnote{
 The assumption of i.i.d. is 
 a special case of exchangeability.
} data \citep{Kallenberg2006}:
\begin{definition_2}
 \label{def:marginal_validity}
 A conformal prediction, $\Ca(\X_{n+1})$, 
 satisfies \emph{marginal validity} if
 \begin{align*}
  \P\left(Y_{n+1} \in C(\X_{n+1}) \right) 
  &\geq 1-\alpha,
  &\text{where } 1-\alpha 
  \text{is a coverage level.}
 \end{align*}
\end{definition_2}
Besides marginal validity one might also 
wish for stronger guarantees.
For instance, one might desire
adequate coverage for each new instance and
not solely on average across instances.
This property is named conditional validity:
\begin{definition_2}
 \label{def:conditional_validity}
 A conformal prediction satisfies 
 \emph{conditional validity} if,
 \begin{align*}
  \P(Y_{n+1}\in \Ca(\textbf{X}_{n+1})
  |\textbf{X}_{n+1}=\textbf{x}_{n+1})
  &\geq 1-\alpha,
  & \text{for every  } \ \x_{n+1} \in \sX.
 \end{align*}
\end{definition_2}
Unfortunately, conditional validity can be
obtained only under strong assumptions 
about the distribution of $(\X,Y)$
\citep{Vovk2012,Lei2014,Barber2019}.
Given this result, effort has been focused
on obtaining  intermediate conditions,
such as local validity:
\begin{definition_2}
 \label{def:local_validity}
 A conformal prediction satisfies 
 \emph{local validity} if,
 \begin{align*}
  \P(Y_{n+1}\in \Ca(\textbf{X}_{n+1})
  |\textbf{X}_{n+1}\in A) &\geq 1-\alpha,
  & \text{ for every } A 
  \text{ in a partition of } \ \sX.
 \end{align*}
\end{definition_2}
Current methods that 
obtain local validity 
compute conformal regions using
only training instances that fall in $A$
\citep{Lei2014,Barber2019,Guan2019}.
However, these methods do not 
scale to high-dimensional settings because 
it is challenging to create $A$ that 
is large enough so that 
many training instances fall in $A$,
and yet small enough so that
local validity is close to
conditional validity.

Another alternative is to obtain
conditional validity at the specified level as
the sample size increases \citep{Lei2018}:
\begin{definition_2}
 \label{def:as_conditional_validity}
 A conformal prediction satisfies 
 \emph{asymptotic conditional validity} if,
 there exist random sets, $\Lambda_n$, such that 
 $\P(\X_{n+1} \in \Lambda_n|\Lambda_n) 
 = 1-o_{\P}(1)$ and
 \begin{align*}
 \sup_{\x_{n+1} \in \Lambda_n}\big
 |\P(Y_{n+1} \in C(\textbf{X}_{n+1})
 |\textbf{X}_{n+1}=\x_{n+1}) 
 - (1-\alpha) \big| 
 = o(1).
\end{align*}
\end{definition_2}
In a regression context in which 
$\sY = \mathbb{R}$, \citet{Lei2018} obtains 
asymptotic conditional coverage under
assumptions such as  $Y = \mu(\X) + \epsilon$,
where  $\epsilon$ is independent of $\X$ and
has density symmetric around 0.
Also, asymptotic conditional coverage was
obtained under weaker conditions with
methods based on quantile regression
\citep{Sesia2019,Romano2019}
and cumulative distribution function (cdf) estimators
\citep{chernozhukov2019distributional,Izbicki2020}.

\begin{figure}
 \centering
 \includegraphics[width=0.45\textwidth]{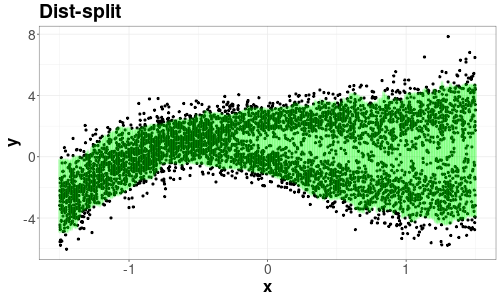}%
 \includegraphics[width=0.45\textwidth]{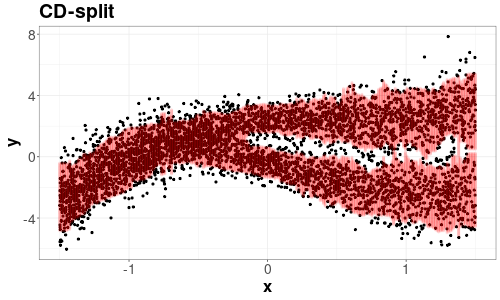}
 \caption{Comparison between 
 a conformal method based on intervals (left) 
 and \cdsplit \ when $Y|\X$ is bimodal.}
 \label{fig:bimodal}
\end{figure}

Besides validity,
it is also desirable to obtain
small prediction regions.
For instance, some of the methods in the last paragraph
converge to the interval with
the smallest length among
the ones with adequate conditional coverage.
However, even the oracle interval can be large.
For instance, \Cref{fig:bimodal} presents 
a case in which, for large values of $\X$,
$Y$ is bimodal. In this case, the
interval-based conformal method in the left
provides large prediction bands, since
it must include the low density region
between the modes of the distribution.

Multimodal data such as in
\Cref{fig:bimodal} often occurs 
in applications \citep{hyndman1996estimating,de2003conditional,Dutordoir2018}.
For instance, theory predicts that
the density of the redshift of a galaxy
based on its photometric features is  
highly asymmetrical and often multimodal
\citep{schmidt2020evaluation}. 
Taking these multimodalities 
into account is necessary to 
obtain reliable
cosmological inferences 
\citep{wittman2009lies},
as further discussed in \Cref{sec:photo-z}.

Whenever relevant multimodalities exist,
a conformal method should
be able to yield regions
which are not intervals.
Moreover, ideally it should 
converge to the 
highest predictive density set (hpd), which
can be considerably smaller than
the smallest interval:
\begin{definition_2}[Convergence to the highest predictive density set]
 \label{def:hpd_conv}
 The highest predictive density set,
 $\Cas(\x)$, is the region with 
 the smallest Lebesgue measure with
 the specified $1-\alpha$ coverage:
 \begin{align*}
  \Cas(\x) 
  &:= \left\{y:f(y|\x) > \qa(\x)\right\},
  & \text{where $\qa(\x)$ is
 the $\alpha$ quantile of $f(Y|\X)$
 given that $\X=\x$.}
 \end{align*}
 A conformal prediction method converges to
 the highest predictive density set if:
 \begin{align*}
  \P(Y_{n+1} \in \Cas(\X_{n+1}) 
  \triangle \Ca(\X_{n+1})) 
  &= o(1),
  & \text{where $A \triangle B :=
  (A \cap B^c) \cup (B \cap A^c)$}
 \end{align*}
\end{definition_2}

Conformal methods that converge to 
the highest predictive density set satisfy 
asymptotic conditional coverage,
as proved in \cref{sec:hpd_to_as}:

\begin{theorem_2}
 \label{thm:triangle_as_equiv}
 If a conformal method converges to
 the highest predictive density set, then
 it satisfies asymptotic conditional coverage.
\end{theorem_2}

To the best of our knowledge, 
\cdsplit \ \citep{Izbicki2020}, 
is the only conformal method that has 
the goal of approximating the highest predictive density sets in high-dimensional feature spaces.
As a result, \cdsplit\ attains prediction regions that
are considerably smaller than
the ones obtained from interval-based methods,
as illustrated in \Cref{fig:bimodal}. 
However, \citet{Izbicki2020} shows 
only empirically that 
\cdsplit \ converges to the 
highest predictive density sets.

In this paper, we further investigate
conformal methods that converge to
the oracle highest predictive density set.
With respect to \cdsplit, we prove that
it satisfies local and asymptotic conditional validities,
and also converges to the oracle highest density set.
We also perform many simulation studies which 
guide the choice in \cdsplit 's tuning parameters.
Next, we introduce a new method, \hpdsplit, which
requires less tuning parameters than \cdsplit, while
having similar theoretical guarantees
and similar performance in simulations.
In order to obtain such properties, 
\hpdsplit \ uses an informative conformity score which,
under weak assumptions, is approximately
independent of the features.

\Cref{sec:review} reviews 
some conformal prediction
strategies that appear in the literature.
\Cref{sec:conformal_cde} describes
the proposed conformal
conditional density estimation methods.
\Cref{sec:assumptions} presents
the main theoretical properties of
the proposed methods.
In particular, it is shown that 
all proposed methods
satisfy marginal and
asymptotic conditional validity and
converge to the
highest predictive density set.
\Cref{sec:exp} uses several new experiments
based on simulations to 
show how to tune \cdsplit \ and
to compare \cdsplit \ and \hpdsplit \
to other existing methods.
\Cref{sec:photo-z} demonstrates
the performance of \cdsplit \ and
\hpdsplit \ in an application of
photometric redshift prediction.
All proofs are in the Appendix.

\section{Review of some conformal prediction strategies}
\label{sec:review}

The main goal in conformal predictions is
to use the data to obtain
a valid prediction region, $C(\X_{n+1})$.
A general strategy for obtaining 
such a region is the split method \citep{papadopoulos2002inductive,Vovk2012,Lei2018}.
Under this method, 
the data is divided into two sets:
the training set,
$\D'=\{(\X'_1,Y'_1),\ldots,(\X'_n,Y'_n)\}$,
and the prediction set,
$\D=\{(\X_1,Y_1),\ldots,(\X_m,Y_m)\}$.
We define $m = n$ 
solely to simplify notation.
Next, a conformity score,
$\hg: \sX \times \sY \rightarrow \Re$,
is trained using  $\D'$.
Finally, $\D$
is used for calculating $U_i := \hg(\X_i,Y_i)$,
the \emph{split residuals}.
Since the split residuals are  
assumed to be exchangeable,
the rank of $U_{n+1}$ is uniform
among $\{1,\ldots,n+1\}$ and,
by letting $U_{\lfloor\alpha\rfloor}$ to be
the $\lfloor n\alpha \rfloor$ order statistic
among $U_1,\ldots,U_n$, obtain that
\begin{align*} 
 \P\left(U_{n+1} \geq U_{\lfloor\alpha\rfloor}\right)
 &\geq 1-\alpha \nonumber \\
 \P\left(\hg(\X_{n+1}, Y_{n+1}) \geq U_{\lfloor\alpha\rfloor}\right)
 &\geq 1-\alpha \nonumber \\
 \P\left(Y_{n+1} \in 
 \left\{y: \hg(\X_{n+1}, y) \geq U_{\lfloor\alpha\rfloor}\right\}\right)
 &\geq 1-\alpha.
\end{align*}
That is, $C(\X_{n+1}) = 
\left\{y: \hg(\X_{n+1}, y) 
\geq U_{\lfloor\alpha\rfloor}\right\}$
is a marginally valid prediction region.
However, in order to obtain 
stronger types of validity,
it might be necessary 
to change the definition of the cutoff,
$U_{\lfloor\alpha\rfloor}$, so that
it adapts to the value of $\X_{n+1}$.

One way to obtain this adaptivity is
to calculate the cutoff using solely
the instances in $\D$ with
covariates close to $\X_{n+1}$.
For instance, \citet{Lei2014}
divides $\sX$ in a partition, $\sA$,
and compute $U_{\lfloor\alpha\rfloor}$
using solely the instances that
fall in the same partition element as $\X_{n+1}$,
as in \cref{def:local_split}.
\begin{definition_2}
 \label{def:local_split}
 Let $\sA$ be a partition of $\sX$.
 For each $A \in \sA$, let
 \begin{align*}
  A(\X_{n+1}) = \{(\X,Y) \in \D: 
  \exists A \in \sA \text{ s.t. } 
  \X \in A \text{ and } \X_{n+1} \in A\}   
 \end{align*}
 Let $U_i = \hg(\X_i,Y_i)$ be
 computed for each $(\X_i,Y_i)$ in $A(\X_{n+1})$.
 $U_{\lfloor \alpha \rfloor}(\X_{n+1})$ is the
 $\lfloor |A(\X_{n+1})| \cdot \alpha \rfloor$ 
 order statistic of these $U_i$.
\end{definition_2}
Since the split residuals 
in \cref{def:local_split}
are exchangeable, one still obtains
marginal validity by substituting
$U_{\lfloor\alpha\rfloor}$ for
$U_{\lfloor\alpha\rfloor}(\X_{n+1})$
in the split method.

\begin{theorem_2}
 \label{thm:local_Validity}
 If $\D=\{(\X_1,Y_1),\ldots,(\X_n,Y_n),
 (\X_{n+1},Y_{n+1})\}$ are 
 exchangeable, then
 $$C(\X_{n+1}) = 
 \left\{y: \hg(\X_{n+1}, y) 
 \geq U_{\lfloor\alpha\rfloor}
 (\X_{n+1})\right\}$$ satisfies
 marginal validity and
 local validity with
 respect to $\mathcal{A}$.
\end{theorem_2}

The methods introduced in this paper
are based on the above strategies,
as described in the following section.
\section{Conformal predictions based on 
conditional density estimators}
\label{sec:conformal_cde}

Our proposed methods are based on
conditional density estimators.
In all of them, 
the conformity score, $\hat{g}$,
is a function of an estimator
of the conditional density
of $Y$ given $\X$, $f(y|\x)$.
This conditional density estimator is
denoted by $\hf(y|\x)$.
Since $\hf(y|\x)$ is defined 
no matter whether $Y$ is 
discrete or continuous,
the proposed methods are
applicable both to
conformal regression and to
conformal classification.
When $\sY$ is discrete,
$\hf$ is a conditional probability estimate, $\hat{P}(y|\x)$.

All of the proposed methods use
the random variable $Z:=f(Y|\X)$.
The conditional cumulative 
density function (cdf) of $f(Y|\X)$, 
$H(z|\x)$, and its estimate,
$\hH(z|\x)$, are presented in
\cref{def:split_cdf}:

\begin{definition_2}
 \label{def:split_cdf}
 \label{def:res_cdf}
 $H(z|\x)$ and $\hH(z|\x)$ are,
 respectively, the conditional cdf
 of $f(Y|\X)$ and its estimate:
 \begin{align*}
  H(z|\x) 
  &:= \int_{\{y:f(y|\x) \leq z\}}{f(y|\x)dy} \\
  \hH(z|\x) 
  &:= \int_{\{y:\hf(y|\x) \leq z\}}{\hf(y|\x)dy}
 \end{align*}
\end{definition_2}
Note that $H(z|\x)$ is the
conditional cdf of $f(Y|\X)$, which
is different from $F(y|\x)$,
the conditional cdf of $Y$.
Besides these quantities,
the conditional quantiles of
$f(Y|\X)$, $\qa(\x)$,
and their estimates, $\hqa(\x)$,
are also useful:
\begin{definition_2}
 \label{def:split_quantile}
 \label{def:u_quantile}
 $\qa(\x) := H^{-1}(\alpha|\x)$ is
 the $\alpha$-quantile of $f(Y|\X)$.
 Also, $\hqa(\x) :=  \hH^{-1}(\alpha|\x)$ is
 an estimate of the conditional
 $\alpha$-quantile of $f(Y|\X)$.
\end{definition_2}

Given the above definitions,
it is possible to describe
the proposed methods.
\subsection{\cdsplit}
\label{subsec:cdsplit}

\cdsplit \ uses the split method and
adaptive cutoffs.
In \cdsplit,
the conformity score, $\hg$, is a
conditional density estimator, $\hf(y|\x)$:
\begin{definition_2}
\label{def:cd_split_residuals}
 The \cdsplit \ residual is 
 $U_i := \hf(Y_i|\X_i)$.
\end{definition_2}

\cdsplit \ also uses the
partition method, as outlined in \cref{def:local_split}.
The performance of this method is
highly dependent on the chosen partition.
For instance, if the partition were
defined according to the Euclidean distance
on the feature space, then
\cdsplit \ would not scale to
high-dimensional feature spaces 
\citep{Lei2014,Barber2019,tibshirani2019conformal}.
In these settings small Euclidean neighborhoods 
have few data points and, therefore,
the partition would be composed of 
large neighborhoods. As a result, 
each partition element would contain features
with drastically varying densities, that is,
the method would deviate strongly from
conditional coverage.
\cdsplit \ avoids this problem by
choosing a partition such that, if
$\x_i$ and $\x_j$ fall 
in the same partition element, then
$U_i = \hf(Y_i|\x_i)$ and
$U_j = \hf(Y_j|\x_j)$ have
similar $\alpha$ quantiles.
\cref{def:cd_split_partition} 
formalizes this idea:

\begin{definition_2}[\cdsplit \ partition]
 \label{def:cd_split_partition}
 Let $\mathcal{I}$ be a partition of  
 $\mathbb{R}^+$.
 $\sA$ is a partition of $\sX$ such that
 $\x_i$ and $\x_j$ are in the
 same partition element of $\sA$ 
 if and only if
 $\hat{q}_{\alpha}(\x_i)$ and
 $\hat{q}_{\alpha}(\x_j)$ are 
 in the same partition element of $\mathcal{I}$.
\end{definition_2}

\Cref{def:cd_split_partition} partitions 
the feature space in a way that
is directly related to conditional coverage.
As a result, coarse partitions can 
still deviate weakly from conditional coverage.

By combining the ideas above,
it is possible to formally define \cdsplit.

\begin{definition_2}[\cdsplit]
 \label{def:cd_split}
 Let $U_{\lfloor \alpha \rfloor}(\X_{n+1})$
 (\cref{def:local_split}) be computed 
 using the partition in
 \cref{def:cd_split_partition}.
 The \cdsplit \ conformal prediction,
 $C(\X_{n+1})$ is:
 \begin{align}
  \label{eq:cd_split}
  C(\x_{n+1}) =
  \left\{y:\hf(y|\x_{n+1})
  \geq U_{\lfloor \alpha \rfloor}(\X_{n+1})\right\},
 \end{align}
 \Cref{alg:cd} summarizes the implementation of \cdsplit.
 \begin{algorithm}
  \caption{ \small \cdsplit}
  \label{alg:cd}
  \algorithmicrequire \ {\small Data $(\x_{i},y_{i})$,
  $i=1,...,n$, coverage level $1-\alpha \in (0,1)$,
  algorithm $\mathcal{B}$ for fitting
  conditional density function,
  a partition of $\mathbb{R}^+$, 
  $\mathcal{I}$.} \\
  \algorithmicensure \ {\small Prediction band for
  $\x_{n+1}\in\mathbb{R}^d$}
  \begin{algorithmic}[1]
   \STATE Randomly split $\{1,2,...,n\}$ into 
   two subsets $\D$ and $\D'$
   \STATE Compute $\hat{f} =
   \mathcal{B}(\{(\x_i,y_i):i \in \D' \})$
   \codecomment{Estimate conditional density}
   \STATE Use $\hat{f}$ to 
   compute $\hqa (\x_i)$, 
   a quantile estimate,
   for each $(\x_i, y_i) \in \D$
   \codecomment{(\cref{def:u_quantile})}
   \STATE Determine $A(\x_{n+1})$,
   the set of $(\x_i,y_i) \in \D'$ 
   such that $\hqa(\x_i)$ and 
   $\hqa(\x_{n+1})$ belong to
   the same partition element of $\mathcal{I}$
   \codecomment{(\cref{def:local_split,def:cd_split_partition})}
   \STATE Compute 
   $U_{\lfloor\alpha\rfloor}(\x_{n+1})$,
   the $\alpha$-quantile of
   $\left\{\hf(y_i|\x_i): 
   (\x_i,y_i) \in A(\x_{n+1})\right\}$
   \STATE Build a finite grid 
   over $\mathcal{Y}$ and,
   by interpolation,
   \textbf{return} 
   $\left\{y:\hf(y|\x_{n+1})
   \geq U_{\lfloor\alpha\rfloor}(\x_{n+1})\right\}$
  \end{algorithmic}
 \end{algorithm}
\end{definition_2}

By observing \cref{eq:cd_split},
it is possible to obtain some
intuition on the theoretical properties
of \cdsplit.
If $\hf \approx f$ and
$\sA$ is a sufficiently fine partition,
then for every $\x \in A(\x_{n+1})$,
$\qa(\x) \approx \qa(\x_{n+1})$.
As a result, if there are
many instances in $A(\x_{n+1})$
$U_{\lfloor\alpha\rfloor}(\x_{n+1}) 
\approx \qa(\x_{n+1})$ and
$C(\x_{n+1}) \approx 
\{y: f(y|\x_{n+1}) \geq \qa(\x_{n+1})\}$,
the oracle conformal prediction. 
If the conditional density is well estimated, 
then the conformal band is 
close to the oracle band,
as discussed in \cref{sec:theory}.

Despite the above desirable properties, 
the partition in \cdsplit \ requires
many tuning parameters.
The following subsection
introduces \hpdsplit ,
which avoids the partition method while
yielding conformal regions similar to the ones in \cdsplit .
\subsection{\hpdsplit}
\label{sec:hpd}

Several of the tuning parameters in
\cdsplit \ follow from the fact that
its conformity score is not approximately
independent of the features, $\X$.
Indeed, observe that \cdsplit \ uses $\hat{f}(Y|\X)$ as
a conformity score. Since the
conditional distribution of $f(Y|\X)$ is
generally not independent of $\X$,
the conformity score in \cdsplit \ 
is not approximately independent of $\X$.
As a result, in order to obtain properties
such as asymptotic conditional coverage,
\cdsplit \ relies on
the partition-method in \cref{sec:review},
which increases the number of tuning parameters.
Given the above considerations,
one might imagine that it is possible to
reduce the number of tuning parameters in \cdsplit \ 
by choosing a conformity score that
is approximately independent of $\X$.

\begin{figure}
 \centering
 \includegraphics[trim={0 0 400 0}, clip,scale=0.32]
 {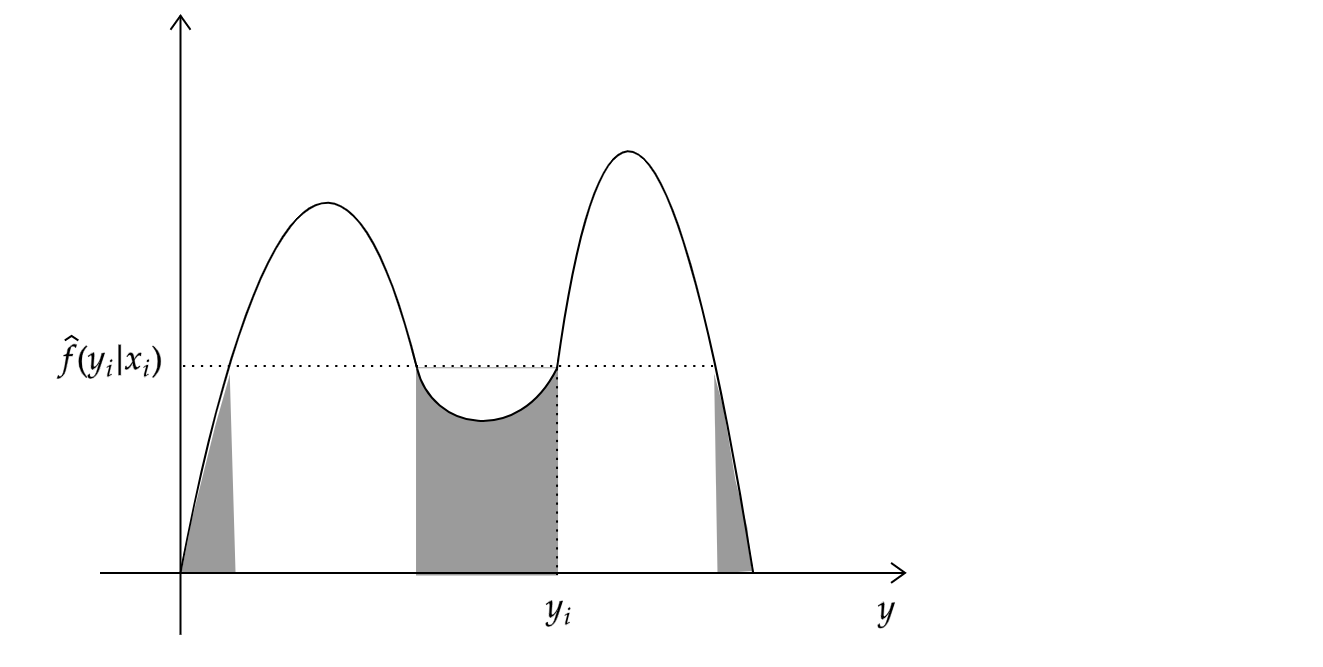}
 \caption{The \hpdsplit \ score of sample $(\x_i,y_i)$ is given by the shaded area on the plot.}
 \label{fig:hpd}
\end{figure}

The idea above is the main intuition for $\hpdsplit$.
In \hpdsplit , the conformity score is
$\hH(\hf(y|\x)|\x)$ instead of $\hf(y|\x)$ (recall \Cref{def:split_cdf}). 
Observe that $H(z|\x)$ is the 
conditional cdf of $f(Y|\X)$ and,
therefore, $H(f(Y|\X)|\X) 
\sim U(0,1)$ given $\X$.
That is, $H(f(Y|\X)|\X)$ is 
independent of $\X$.
Therefore, if $\hf(y|\x)$ converges to $f(y|\x)$, then
one might expect that
the conformity score $\hH(\hf(Y|\X))$ is
approximately independent of $\X$.
As a result, the partition method need not be used.
The \hpdsplit\ residuals are formalized in 
\cref{def:hpd_split_residual}:
\begin{definition_2}[\hpdsplit \ residual]
 \label{def:hpd_split_residual}
 The \hpdsplit \ residuals are given by
 \begin{align*}
  U_i &:=
  \hH\left(\hf(Y_i|\X_i)|\X_i\right)=\int_{\left\{y:\hf(y|\X_i) 
  \leq \hat f(Y_i|\X_i)\right\}}{\hf(y|\X_i)dy}.
 \end{align*}
\end{definition_2}
The \hpdsplit \ residual is 
the area of the curve $\hf$ below $\hf(Y_i|\X_i)$ 
(the shaded area in \cref{fig:hpd}). 
\hpdsplit \ uses this residual in
the standard split method:
\begin{definition_2}[\hpdsplit]
 \label{def:hpd_split}
 The \hpdsplit \ conformal prediction is
 \begin{align}
 \label{eq:hpd_split}
  C(\x_{n+1}) 
  :=&
  \left\{y: 
 \hH\left(\hf(y|\x_{n+1})|\x_{n+1}\right)
 \geq  U_{\lfloor \alpha \rfloor}\right\} \nonumber \\ 
 =& \left\{y: 
 \hf(y|\x_{n+1})   \geq  \hH^{-1}\left( U^{\text{hpd}}_{\lfloor \alpha \rfloor}  \ |\x_{n+1} \right)\right\}.
 \end{align}
\end{definition_2}
\Cref{alg:hpd} summarizes 
 the implementation of \hpdsplit.
 
By observing \cref{eq:hpd_split},
it is possible to obtain some
intuition on the theoretical properties
of \hpdsplit.
If $\hf \approx f$, then
$U_i \approx U(0,1)$ and
$U^{\text{hpd}}_{\lfloor \alpha \rfloor}
\approx \alpha$.
Similarly, since $\hf \approx f$,
$\hH^{-1} \approx H^{-1}$ and
$\hH^{-1}(\alpha|\x) \approx \qa(\x)$.
By combining these approximations
in \cref{eq:hpd_split},
one obtains that
$C(\x_{n+1}) \approx 
\left\{y: f(y|\x_{n+1}) \geq \qa(\x_{n+1})\right\}$,
the oracle highest predictive set.
Indeed, $\hpdsplit$ converges
to the highest predictive set and
satisfies asymptotic 
conditional validity, as
discussed in \cref{sec:theory}.

If $Y$ is discrete, then
\hpdsplit\ is equivalent to 
the adaptive method proposed by  \citet{romano2020classification}. 
The CQC method
\citep{cauchois2021knowing}
is an alternative 
when $Y$ is discrete that 
also satisfies
asymptotic conditional validity.

\begin{algorithm}
 \caption{ \small \hpdsplit}
 \label{alg:hpd}
 \algorithmicrequire \ {\small 
 Data $(\x_{i},Y_{i})$, $i=1,...,n$, 
 coverage level $1-\alpha \in (0,1)$, 
 algorithm $\mathcal{B}$ for 
 fitting conditional density function.} \\ 
 \algorithmicensure \ {\small Prediction band 
 for $\x_{n+1}\in\mathbb{R}^d$}
 \begin{algorithmic}[1]
  \STATE Randomly split $\{1,2,...,n\}$ into 
  two subsets $\D$ and $\D'$
  \STATE Fit $\hat{f} = 
  \mathcal{B}(\{(\x_i,Y_i):i \in \D' \})$
  \codecomment{Estimate conditional density function}
  \STATE Let $\hat{H}$ be an estimate of
   the cdf of the split residuals,
   $\hat{f}(Y|\X)$, obtained by
   numerical integration
   (\cref{def:split_cdf}).
   \codecomment{Estimate of
   split residual cdf}
   \STATE Let $U_{\lfloor\alpha\rfloor}$
   be the $\alpha$-quantile of
   $\left\{\hH\left(\hf(y_i|\x_i)|\x_i\right): 
   i \in \mathbb{D} \right\}$.
  \STATE Build a finite grid 
   over $\mathcal{Y}$ and,
   by interpolation,
   \textbf{return}
  $\left\{y:\hH\left(\hf(y|\x_{n+1})|\x_{n+1}\right)\geq
  U_{\lfloor\alpha\rfloor}\right\}$.
 \end{algorithmic}
\end{algorithm}

Despite the simplicity of \hpdsplit,
the partition method can obtain
better conditional coverage in
a few scenarios. The following section
presents an improved partition,
which is used to define \cdsplitp .

\subsection{\cdsplitp: improved partitions}
\label{sec:cdsplitp}

\cdsplit \ partitions the feature space
based on solely the estimated
conditional $\alpha$ quantiles 
of the \cdsplit \ residuals. 
Given this partition's reliance on a single point, 
\cdsplit \ might be unstable in some scenarios.
An alternative in \citet{Izbicki2020} is to 
build a partition based on the full estimate
of the cdf of the \cdsplit \ residuals, $\hH(z|\x)$:

\begin{definition_2}[Profile distance]
 \label{def:profile_dist}
 \label{def:profile}
 The profile distance\footnote{
  The profile distance is a metric on
  the quotient space $\mathcal{X}/\sim$, 
  where $\sim$ is the equivalence relation
  $\x_a \sim \x_b \iff 
  \hg_{\x^a} \equiv \hg_{\x^b}$.
 }
 between $\x_a, \x_b \in \mathcal{X}$ is
 \begin{align*} 
  d^2(\x_a,\x_b) 
  &:= \int_{-\infty}^\infty 
  \left(\hH(z|\x_a)-\hH(z|\x_b)\right)^2 dz,
 \end{align*}
\end{definition_2}
The profile distance is chosen so that
two goals are satisfied.
First, if two instances are close, then
their split residuals have 
approximately the same conditional distribution.
As a result, if one chooses a partition
in such a way that all instances 
are close in the profile distance, then
the instances are approximately exchangeable
conditionally on $\X_{n+1}$.
Second, two points can be 
close in the profile distance even though
they are far apart in the Euclidean distance.
As a result, the profile distance
avoids the curse of dimensionality and
has a large number of instances even
in partitions composed of small neighborhoods.
This idea is illustrated in 
\cref{ex:location,ex:irrelevant}:

\begin{example}\textbf{[Location family]}
 \label{ex:location}
 Let $\hf(y|\x) = h(y-\mu(\x))$, where
 $h(y)$ is a density and
 $\mu(\x)$ an arbitrary function.
 In this case, $d(\x_a,\x_b) = 0$,
 for every $\x_a,\x_b \in \mathbb{R}^d$.
 For instance, if
 $Y|\X \sim N(\beta^t \X, \sigma^2)$, then
 all split residuals have the same conditional distribution.
 A partition based on the profile distance
 would have a single element.
\end{example}

\begin{example}\textbf{[Irrelevant features]}
 \label{ex:irrelevant}
 If \ $\x_S$ is a subset of the features such that
 $\hf(y|\x) = \hf(y|\x_S)$, then
 $d(\x_a,\x_b)$ does not depend
 on the irrelevant features, $S^c$.
 While irrelevant features do not
 affect the profile distance,
 have a large impact
 in the Euclidean distance in
 high-dimensional settings.
\end{example}

The profile distance is also related to \cdsplit.
Note that $d(\x_a,\x_b) = 0$
if and only if the split residuals
$\hf(Y|\x_a)$ and $\hf(Y|\x_b)$ have
the same estimated conditional cdfs.
That is, $\hqa(\x_a) = \hqa(\x_b)$,
for every $\alpha$. In this sense,
while \cdsplit \ generates a partition
that compares $\hqa(\x_a)$ and $\hqa(\x_b)$
for a single $\alpha$,
the profile distance can be used
to create a partition that
compares these values for 
every $\alpha \in [0,1]$,
as formalized in \cref{thm:equivalance}:
\begin{theorem_2}[Theorem 3.9 in \citet{Izbicki2020}]
 \label{thm:equivalance}
 Let $\hf(\cdot|\x)$ be a density
 w.r.t. the Lebesgue measure,
 for every $\x \in \sX$.
 The equivalence relation 
 $\x_a \sim \x_b \iff d(\x_a, \x_b) = 0$ is
 the minimal equivalence relation s.t.,
 if $\hqa(\x_a) = \hqa(\x_b)$ 
 for every $\alpha \in [0,1]$, then
 $\x_a \sim \x_b$.
\end{theorem_2}

\begin{figure}
 \centering
 \hspace{-10mm}\includegraphics[scale=0.32]
 {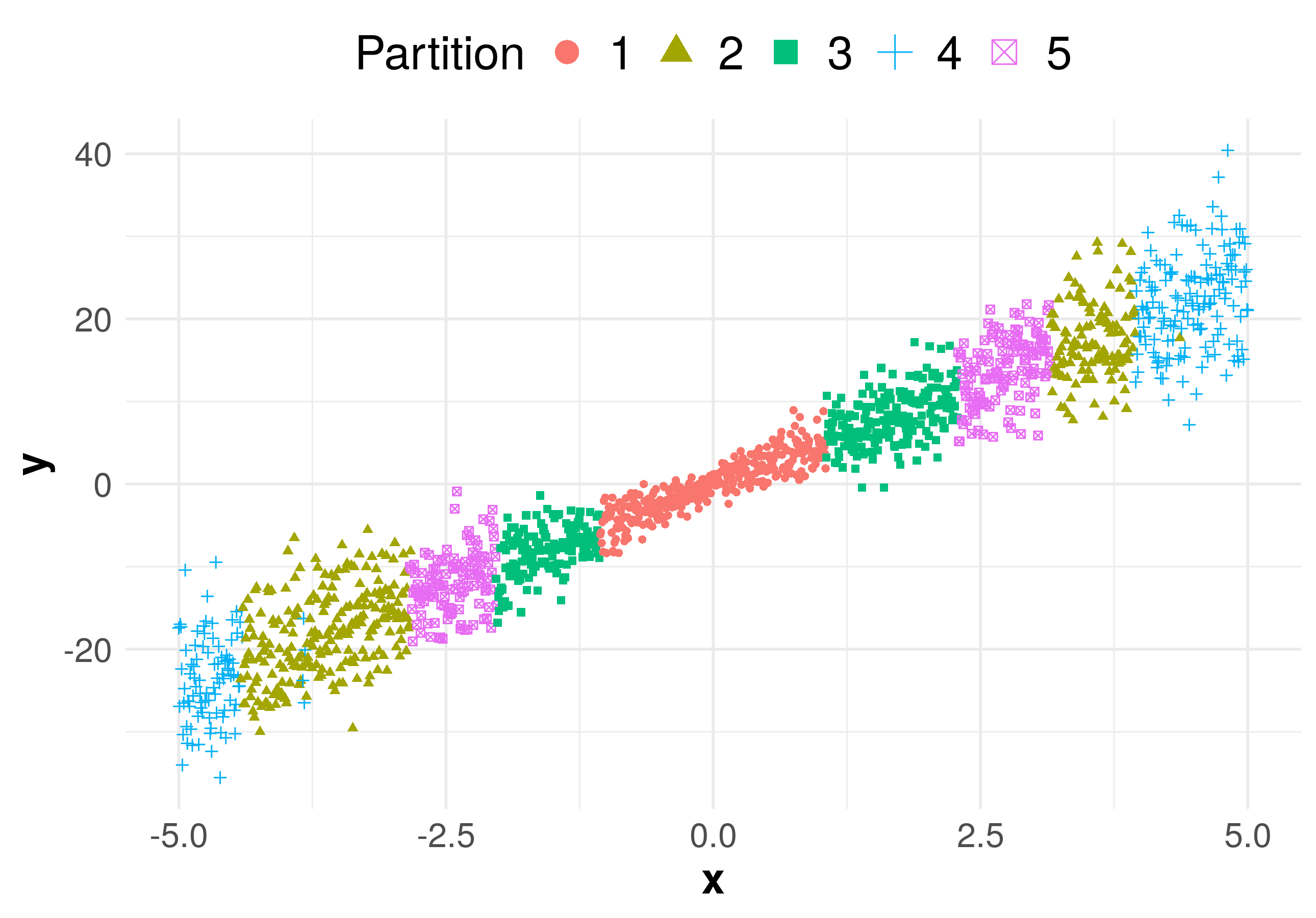}
 \caption{Scatter plot of data generated 
 according to $Y|x \sim N(5x, 1+|x|)$. 
 Colors indicate partitions that 
 were obtained from \cdsplitp.
 Note that points that are far from each other on 
 the $x$-axis can have similar densities and 
 belong to the same element of the partition.
 This allows larger partition elements while
 preserving the optimal cutoff.}
 \label{fig:partitions}
\end{figure}

The profile distance induces
a new partition over $\sX$, which
is used to define \cdsplitp:

\begin{definition_2}[\cdsplitp]
 \label{def:cd_split_p}
 Let $\C_1^{'},\ldots,\C_J^{'}$ be
 centroids for $\X_1^{'}, \ldots, \X_n^{'}$
 from $\D^{'}$ according to $d$,
 the profile distance 
 (\cref{def:profile_dist}).
 Let $\sA$ be a partition of $\sX$ 
 such that $\x \in A_j$ if and only if
 $d^2(\x, \C_j^{'}) < d^2(\x, \C_k^{'})$,
 for every $k \neq j$.
 That is, $\sA$ is the Voronoi partition
 generated from 
 $\C_1^{'},\ldots,\C_J^{'}$ and $d^2$. 
 \cdsplitp is defined in the same way 
 as \cdsplit \ (\cref{def:cd_split}), 
 but using this Voronoi partition
 instead of the one in
 \cref{def:cd_split_partition}.
\end{definition_2}
In practice, several algorithms might
be used for determining the centroids in
\cref{def:cd_split_p}. Here,
for each $i \in \D$, we let 
$\w_i$ be a discretization
obtained by evaluating $\hH(\cdot|\x_i)$
on a finite grid of values.
The clustering algorithm k-means++
\citep{arthur2007k} determines
centroids over these $\w_i$.
Finally, the values of $\x_i$ that
generated the $\w_i^c$ are chosen
as the centroids in \cref{def:cd_split_p}.
Algorithm \ref{alg:cdp} shows 
pseudo-code for this implementation.
\Cref{fig:partitions} illustrates 
a partition used in \cdsplitp .
Instances that are far apart 
in the Euclidean distance but
have similar split residuals are
put in the same partition.
\begin{algorithm}
 \caption{ \small \cdsplitp}
 \label{alg:cdp}
 \algorithmicrequire \ {\small 
 Data $(\x_{i},Y_{i})$, $i=1,...,n$, 
 coverage level $1-\alpha \in (0,1)$, 
 algorithm $\mathcal{B}$ for 
 fitting conditional density function, 
 number of elements of the partition $J$.} \\ 
 \algorithmicensure \ {\small Prediction band 
 for $\x_{n+1}\in\mathbb{R}^d$}
 \begin{algorithmic}[1]
  \STATE Randomly split $\{1,2,...,n\}$ into 
  two subsets $\D$ and $\D'$
  \STATE Fit $\hat{f} = 
  \mathcal{B}(\{(\x_i,Y_i):i \in \D' \})$
  \codecomment{Estimate conditional density function}
  \STATE Compute $\mathcal{A}$, 
  the partition of $\mathcal{X}$, 
  by applying  k-means++ to the cdf of the split residuals in $\D$'
  \STATE Compute $\hH(z|\x_{n+1})$, 
  for all $z \in \mathbb{R}$
  \codecomment{cdf of the split residual 
  (\cref{def:split_cdf})}
  \STATE Determine $A(\x_{n+1}) \in \mathcal{A}$, the   element of $\mathcal{A}$ that
  $\x_{n+1}$ belongs to
     \STATE Let $U_{\lfloor\alpha\rfloor}(\x_{n+1})$
   be the $\alpha$-quantile of
   $\left\{\hf(y_i|\x_i): 
   (\x_i,y_i) \in A(\x_{n+1})\right\}$
  \STATE Build a finite grid 
   over $\mathcal{Y}$ and,
   by interpolation,
   \textbf{return} 
   $\left\{y:\hf(y|\x_{n+1})
   \geq U_{\lfloor\alpha\rfloor}(\x_{n+1})\right\}$
 \end{algorithmic}
\end{algorithm}

In the following, we
present theoretical properties of
\cdsplit, \cdsplitp, and \hpdsplit.
\section{Theoretical properties of \cdsplit, \cdsplitp, and \hpdsplit}
\label{sec:assumptions}

\subsection{Marginal and local validity}

Since \cdsplit \ and
\cdsplitp \ are based directly
on the split method and
on adaptive cutoffs, it
follows from 
\cref{thm:local_Validity} that,
as long as the instances in $\D$
are exchangeable,
both methods satisfy
local and marginal validity.
This result is rephrased in
\cref{thm:cd_split_control}.

\begin{theorem_2}
 \label{thm:cd_split_control}
 If the instances in $\D$
 are exchangeable, then
 \cdsplit \ and \cdsplitp \ satisfy 
 local validity
 (\cref{def:local_validity}) with 
 respect to  the partition in 
 \cref{def:cd_split_partition}.
 In particular,
 \cdsplit \ and 
 \cdsplitp \ also 
 satisfy marginal validity
 (\cref{def:marginal_validity}).
\end{theorem_2}

$\hpdsplit$ uses
the split method without
adaptive cutoffs. Therefore,
\cref{thm:local_Validity} guarantees
solely that,
if the instances in $\D$
are exchangeable, then
\hpdsplit \ satisfies 
marginal validity.
This result is rephrased in
\cref{thm:hpd_marginal}.

\begin{theorem_2}
 \label{thm:hpd_marginal}
 If the instances in $\mathcal{D}$
 are exchangeable, then
 \hpdsplit \ satisfies marginal validity.
\end{theorem_2}
This result follows directly from
the fact that \hpdsplit \ uses
the standard split method.

\subsection{Convergence to the highest predictive density set and asymptotic conditional validity}
\label{sec:theory}

Although \cref{thm:local_Validity} shows
that it is possible to obtain
marginal and local validity 
requiring solely that 
the data is exchangeable,
further assumptions are required to
obtain convergence to
the highest predictive density set and
asymptotic conditional validity.
First, all instances are 
assumed to be i.i.d.:
\begin{assumption_2}
 \label{ass:iid}
 $\{(\X_1,Y_1),\ldots,(\X_{n+1},Y_{n+1})\}$
 are i.i.d.
\end{assumption_2}
Also, $\hf(y|\x)$ is 
assumed to be consistent:
\begin{assumption_2}[Consistency of $\hf$]
 \label{ass:consistent_cde_2}
 There exist $\eta_n = o(1)$ and
 $\rho_n = o(1)$ s.t.
 \begin{align*}
  \P\left(\E\left[\sup_{y \in \sY}
  \left(\hf(y|\X)-f(y|\X)\right)^2
  \bigg|\hf\right]
  \geq \eta_n \right) \leq \rho_n
 \end{align*}
\end{assumption_2}

Similarly, the conditional
cumulative density function of
$f(Y|\X)$, $H(z|\x)$, may be 
assumed to be well behaved, that is,
be smooth (bounded density) and
have no plateau close the $\alpha$ quantile:

\begin{assumption_2}
 \label{ass:continuous_hpd}
 For every $\x$, $H(u|\x)$ is continuous,
 differentiable and
 $\frac{dH(u|\x)}{du} \leq M_1$.
 Also $\frac{dH(u|\x)}{du} \geq M_2 > 0$
 in a neighborhood of $\qa(\x)$.
\end{assumption_2}

Finally, for some results,
$\sY$ is assumed to be bounded.
This is a weak assumption, since
there exist continuous 
bijective functions that
map $\Re^d$ onto $(-1,1)$. Also,
this assumption could probably
be removed by using stronger 
bounds in the proof of \cref{lemma:hpd_cdf}
in the Appendix.
\begin{assumption_2}
 \label{ass:bounded}
 $\sY$ is bounded.
\end{assumption_2}

Under the above assumptions
\hpdsplit \ converges to
the hpd set 
(\cref{def:hpd_conv}) and
satisfies asymptotic conditional
(\cref{def:as_conditional_validity}).
The same results are also obtained
for \cdsplit \ and \cdsplitp \ as
the partitions used in these methods
become thinner.
These results are presented in
\cref{thm:hpd_hpd,thm:cd_hpd,thm:cdp_hpd}:

\begin{theorem_2}
 \label{thm:hpd_hpd}
 \label{thm:converge_hpd_hpd}
 Under \cref{ass:iid,ass:bounded,ass:consistent_cde_2,ass:continuous_hpd},
 \hpdsplit \ converges to
 the hpd set and
 satisfies asymptotic conditional validity.
\end{theorem_2}
\begin{theorem_2}
 \label{thm:cd_hpd}
 \label{thm:convergeHPD}
 If $|I| = o(1)$, for every $I \in \mathcal{I}$
 (\cref{def:cd_split_partition}), then,
 under \cref{ass:iid,ass:bounded,ass:consistent_cde_2,ass:continuous_hpd}, \cdsplit \ converges to 
 the hpd set and
 satisfies asymptotic conditional validity.
\end{theorem_2}
\begin{theorem_2}
 \label{thm:cdp_hpd}
 If $|A| = o(1)$ for every $A \in \sA$
 (\cref{def:cd_split_p}), then,
 under \cref{ass:iid,ass:bounded,ass:consistent_cde_2,ass:continuous_hpd}, 
 \cdsplitp \ converges to 
 the hpd set  and
 satisfies asymptotic conditional validity.
\end{theorem_2}

The following section presents
simulation studies that
give further support to
the effectiveness of
the proposed methods.

\section{Simulation Studies}
\label{sec:exp}
In order to study the performance of proposed methods
in a conformal regression setting,
this section presents several simulations.
Whenever nothing else is specified,
the experiments are such that
$\X=(X_1,\ldots,X_d)$, with 
$X_i \overset{\text{iid}}{\sim} \mbox{Unif}(-1.5,1.5)$
and $d=20$. The simulated are scenarios
are the following:
\begin{itemize}[wide, labelwidth=!, labelindent=0pt]
 \item\textbf{[Homoscedastic]}
 $Y|\x \sim \mbox{N}(0.3x_1, 1)$.
 
 \item\textbf{[Bimodal]}
 $Y|\x \sim 0.5\mbox{N}(f(\x)-g(\x),
 \sigma^2(\x))+0.5\mbox{N}(f(\x)+g(\x),\sigma^2(\x)),$
 with $f(\x)=(x_1-1)^2 (x_1+1),$ $g(\x)=2\I(x_1 \geq -0.5) \sqrt{x_1+0.5}$, and $\sigma^2(\x)=0.25+|x_1|$.
 This is the example from \citet{Lei2014} with 
 added irrelevant variables.
 
 \item\textbf{[Heteroscedastic]}
 $Y|\x \sim \mbox{N}(0.3x_1, 1 + 0.3|x_1|)$.
 
 \item\textbf{[Asymmetric]}
 $Y|\x = 1.5x_1 + \epsilon$, where $\epsilon 
 \sim \mbox{Gamma}(1 + 0.6|x_1|, 1 + 0.6|x_1|)$.
\end{itemize}

Each scenario runs 5,000 times and
each predictive method uses 
a coverage level of $1-\alpha=90\%$.
Since the implementations of 
all methods obtain marginal coverage 
very close to the nominal $90\%$ level, 
this information is not displayed.
In all scenarios, the implementation of both
\cdsplit \ and \hpdsplit \ use \flexcode \ \citep{Izbicki2017,dalmasso2020} 
to estimate $f(y|\x)$.
\flexcode \ converts directly estimating $f$ into 
estimating regression functions that 
are the coefficients of the expansion of $f$ 
on a Fourier basis and often gives good results. 
The regression functions
are estimated with random forests \citep{Breiman2001}.
Also, unless otherwise stated, 
the feature space is divided in
a partition of size $\ceil{\frac{n}{100}}$,
so that on average 100 instances fall into 
each element of the partition.

\Cref{sec:tuning_cd} discusses how
to choose the tuning parameters in
\cdsplit \ and \cdsplitp .
\Cref{sec:comparison} compares
\cdsplitp \ to \hpdsplit \ and other conformal prediction
methods in the literature.
In both sections, the control of 
the conditional coverage is
measured through the conditional coverage
absolute deviation, that is,
$\E[|\P(Y^* \in C_{\alpha}(\X^*)|\X^*)-(1-\alpha)|]$.
\Cref{sec:high_dim} studies the 
effect of dimensionality over
\cdsplitp \ and \hpdsplit.
\Cref{sec:classification} uses 
a conformal classification setting to
compare \cdsplitp \ to Probability-split 
\citep[Sec. 4.3]{Sadinle2019}.

\subsection{Tuning \cdsplit}
\label{sec:tuning_cd}

\begin{figure}
 \centering
 \includegraphics[width=0.45\textwidth]
 {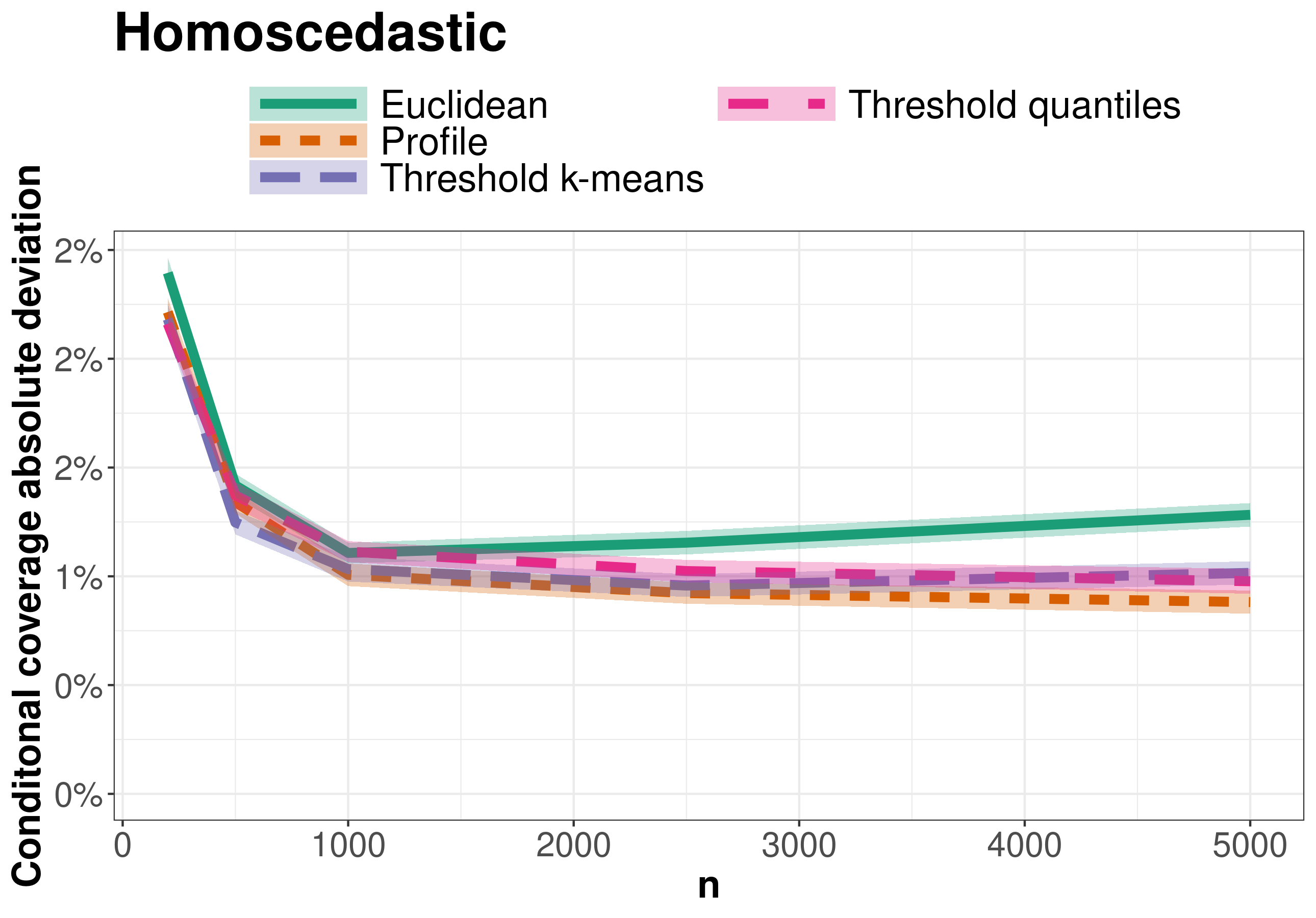}
 \hspace{2mm}
 \includegraphics[width=0.45\textwidth]
 {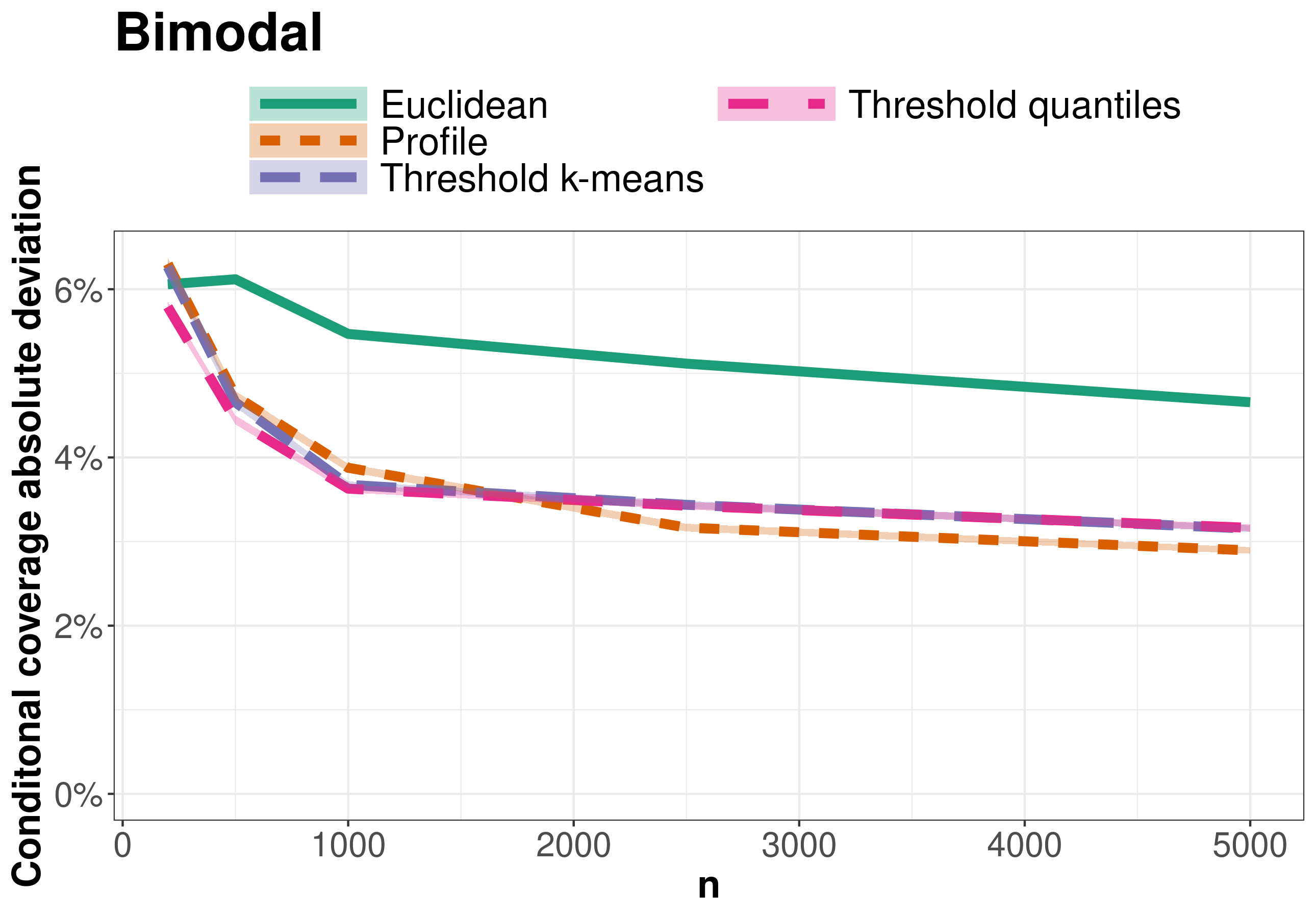} \\
 \includegraphics[width=0.45\textwidth]
 {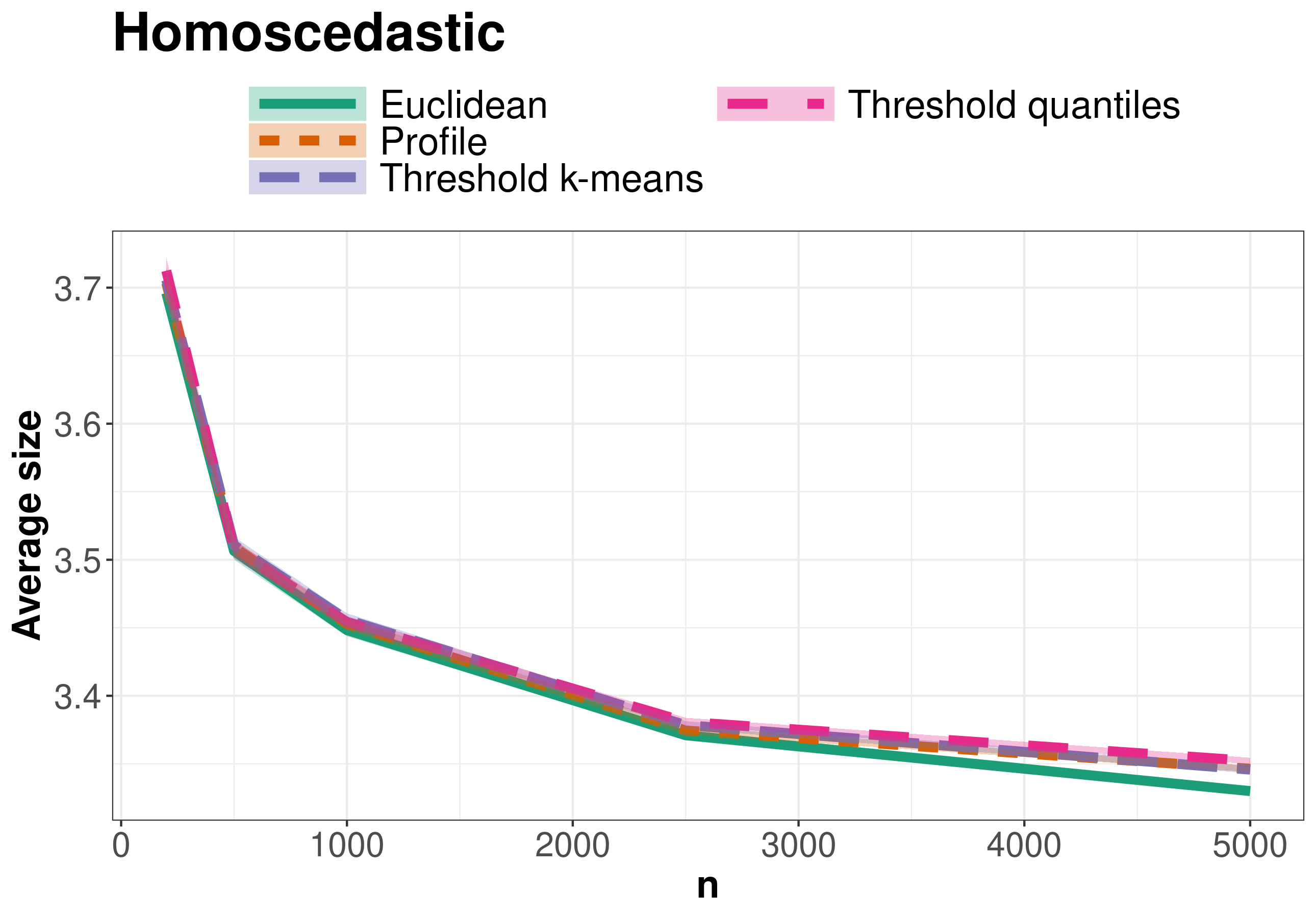}
 \hspace{2mm}
 \includegraphics[width=0.45\textwidth]
 {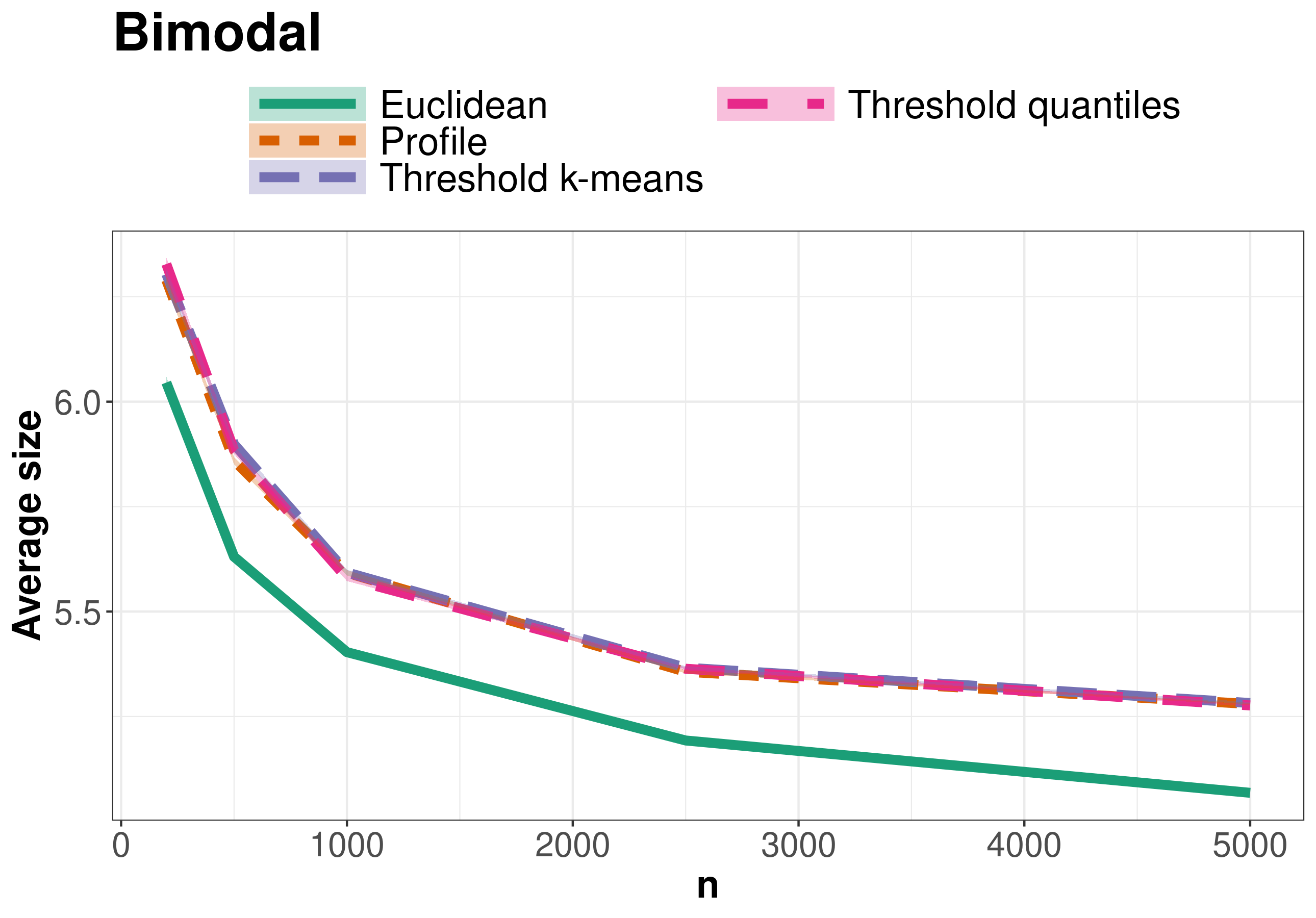}
 \caption{Conditional coverage (top panel)
 and predictive region size (bottom panel) 
 for different partitions in \cdsplit.} 
 \label{fig:partition_coverage_size} 
\end{figure}

Does the performance of \cdsplit \ depend
on the choice of the partitions in
\cref{def:cd_split_partition,def:cd_split_p}?
In order to approach this question,
we consider some variants of \cdsplit:
Euclidean distance partitions, such as in
\citet{Lei2014} (Euclidean),
\cdsplit \ with a partition
that is induced by intervals of
estimated threshold values with
the same number of instances (Threshold quantiles),
\cdsplit \ with a partition chosen
according to k-means over
the estimated quantiles (Threshold k-means), and
the standard \cdsplitp \ (Profile).
The upper panel of 
\Cref{fig:partition_coverage_size}
compares these methods according
to conditional coverage and region size
in the homoscedastic and bimodal scenarios.
In the bimodal scenario
the Euclidean partition 
has worse conditional coverage than
other partitions and \cdsplitp 's 
conditional coverage is
slightly better than that of 
threshold methods.
The heteroscedastic and asymmetric scenarions
behave similarly to the bimodal scenario,
as shown in
\Cref{fig:partition_coverage_size_2} 
in the Appendix.

Besides conditional coverage, 
one might also wish to compare 
the above methods according to
the expected predictive region size.
The lower panel of
\Cref{fig:partition_coverage_size}
allows this comparison.
Generally, all methods yield similar
predictive region sizes.
While in homoscedastic scenario all
partitions yield similar region sizes,
in the bimodal scenario the Euclidean
partition yields considerably smaller regions.
\Cref{fig:partition_coverage_size_2} in
the Appendix shows that the
asymmetric and heteroscedastic scenarios
have similar behaviors.
Since the smaller regions in
the Euclidean partition come 
at the cost of a larger conditional coverage deviation,
it does not indicate
a positive aspect of the Euclidean partition.

The above conclusion can be understood
through a simple toy example.
Consider that $X \sim \text{Bernoulli}(0.025)$,
$Y|X = 0 \sim N(0,1)$, and
$Y|X = 1 \sim N(0, 10^8)$.
In this case, the small predictive region
$C_1(X) \equiv [-2.25, 2.25]$ attains
marginal coverage at the expense of
conditional coverage. Although
$C_2(0) = [-2, 2]$ and
$C_2(1) = [-2 \cdot 10^4, 2 \cdot 10^4]$ yields
intervals that are much larger on average,
the fact that it satisfies conditional coverage
makes it better represent 
the uncertainty about $Y$ given each value of $X$.
$C_2$ is also the smallest region given $X$.

Given the above considerations,
we treat conditional coverage as a primary goal
and region size as a secondary goal.
Since \cdsplitp \ has better conditional coverage
than \cdsplit ,
we compare only \cdsplitp \ to
other methods suggested in the literature.

\begin{figure}
 \centering
 \includegraphics[width=0.45\textwidth]
 {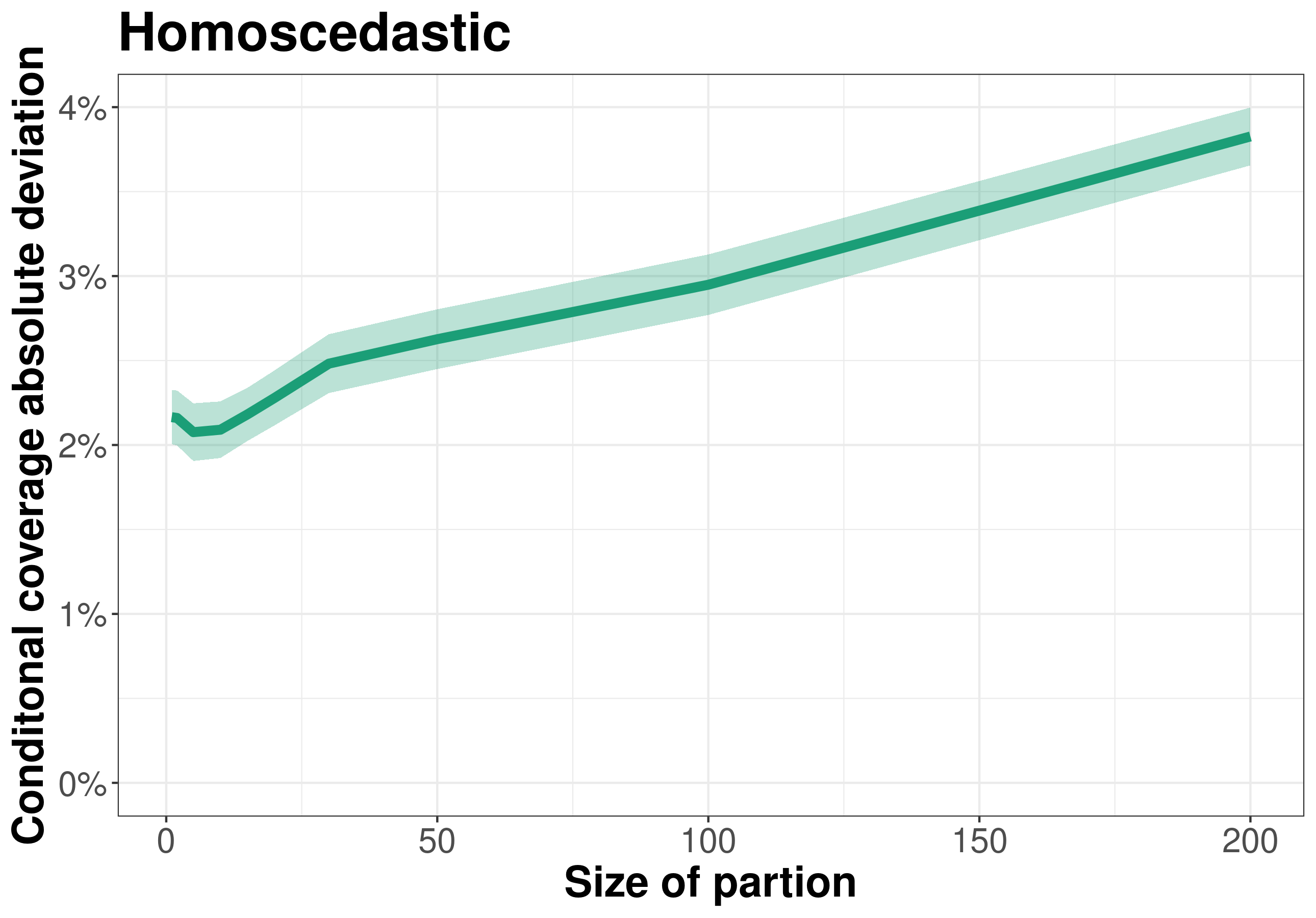} \hspace{2mm}
 \includegraphics[width=0.45\textwidth]
 {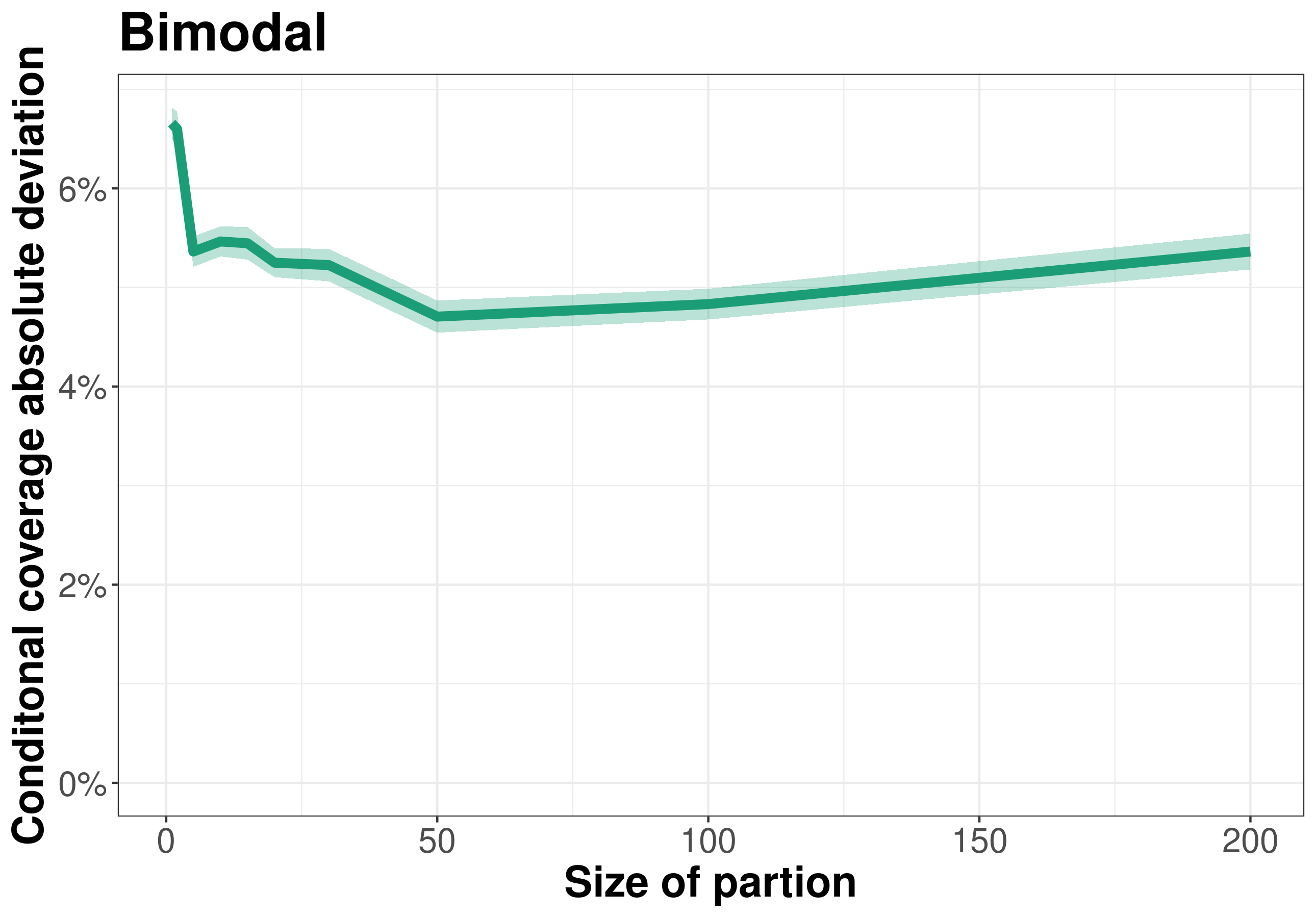} \\
 \includegraphics[width=0.45\textwidth]
 {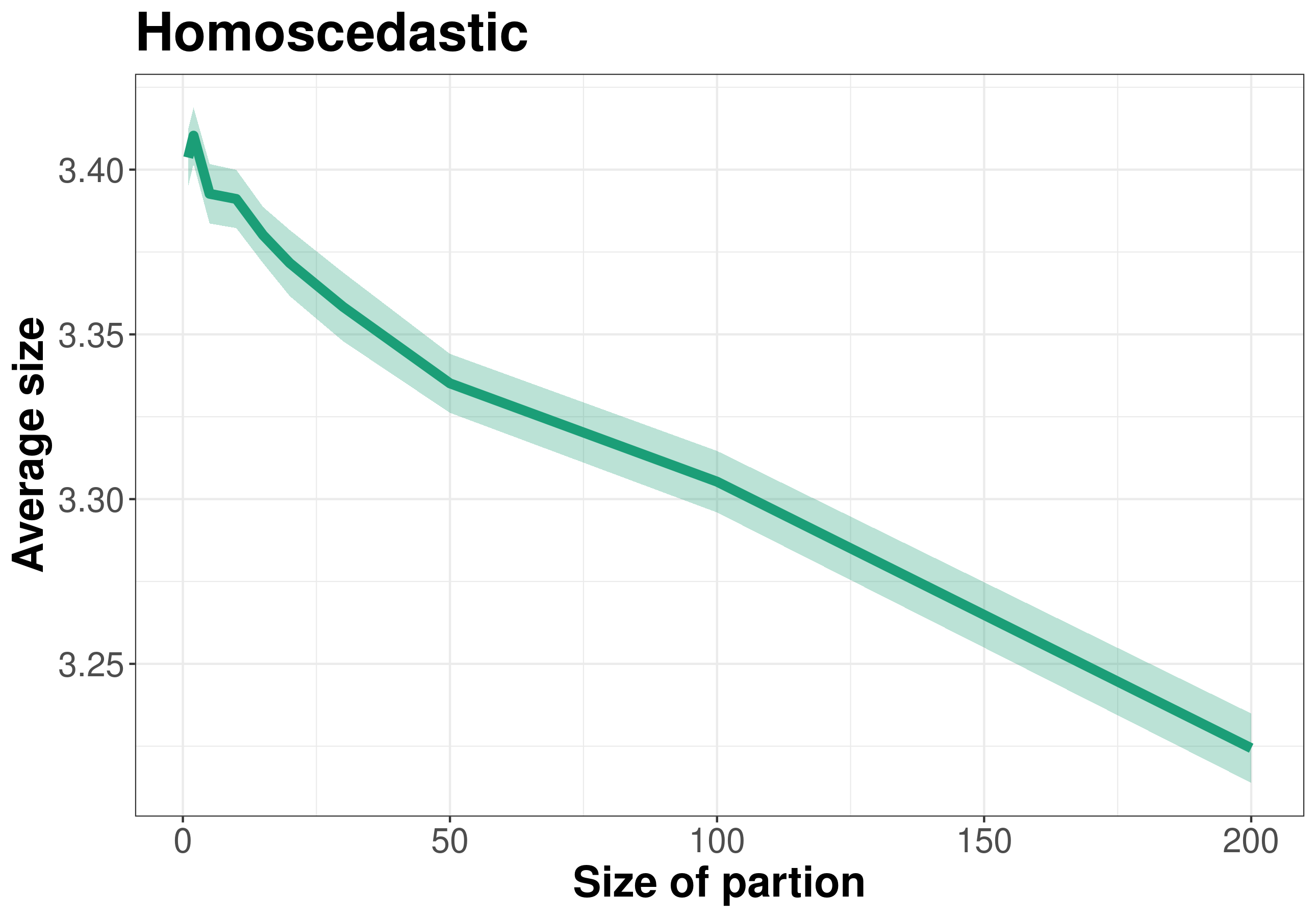}
 \includegraphics[width=0.45\textwidth]
 {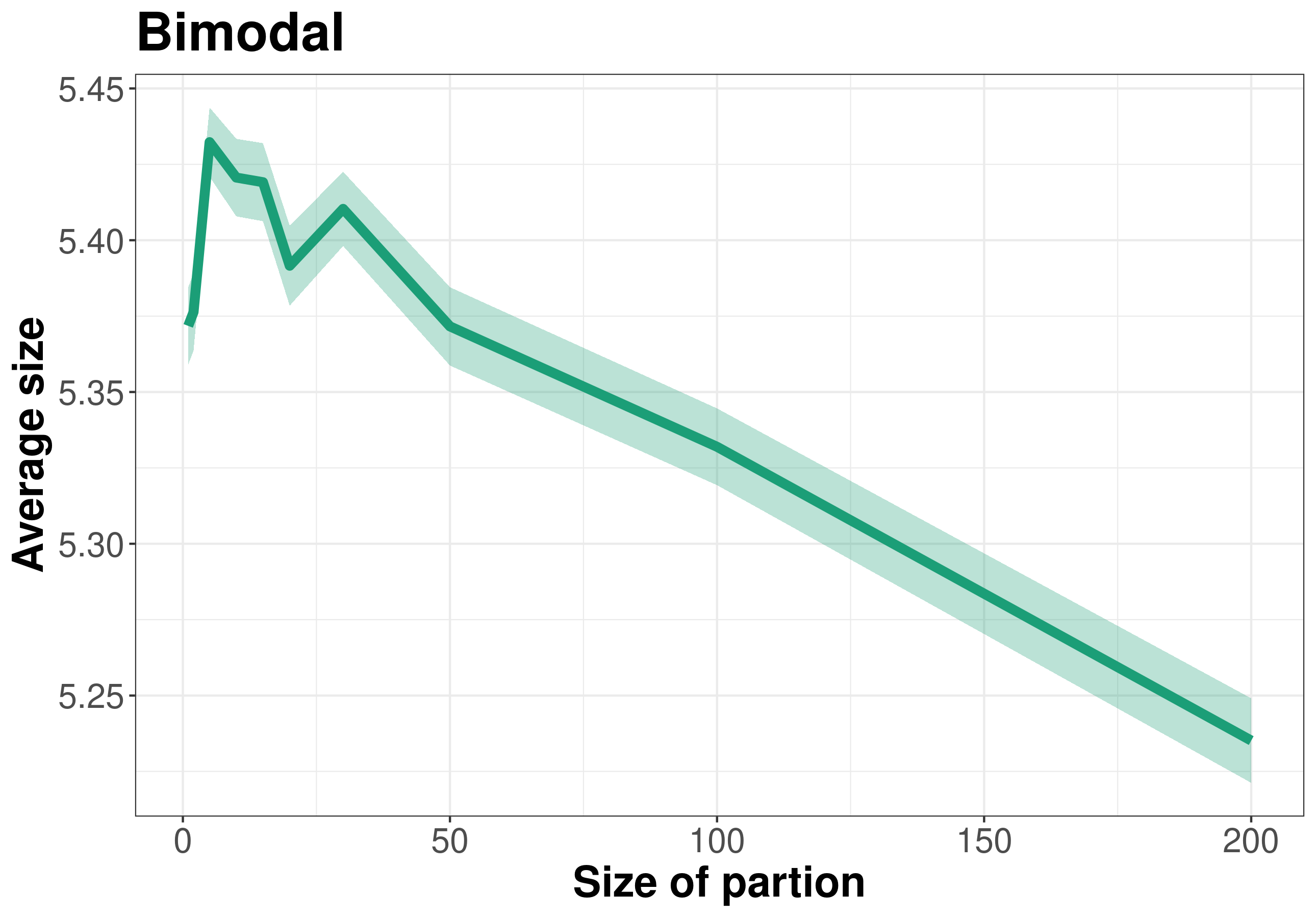}
 \caption{Conditional coverage (top panel) and
 predictive region size (bottom panel) for
 different partition sizes in \cdsplitp .} 
\label{fig:k} 
\end{figure}

Besides choosing the type of partition,
it is also necessary to choose its size.
\Cref{fig:k} shows how the size of the partition
affects the conditional coverage and
region size of \cdsplitp \ in
the homoscedastic and bimodal scenarios.
The upper panel shows that
in the homoscedastic scenario
conditional coverage worsens as
the partition size increases.
This result is compatible with the fact that,
if $f(y|\x)$ were known, then in this scenario
a single partition element would
be required (\cref{ex:location}).
On the other hand, in the bimodal scenario
conditional coverage decreases until
a partition of size $50$ and then
it increases. This behavior represents
the tradeoff between the
number of elements in the partition and
how close each element is to $\x_{n+1}$.
The bottom panels show that, 
in both scenarios, the region size
generally decreases with the partition size.
\Cref{fig:k_2} in the Appendix shows that
the heteroscedastic and asymmetric scenarios
are similar to the bimodal scenario.

\begin{figure}
 \centering
 \includegraphics[width=0.45\textwidth]
 {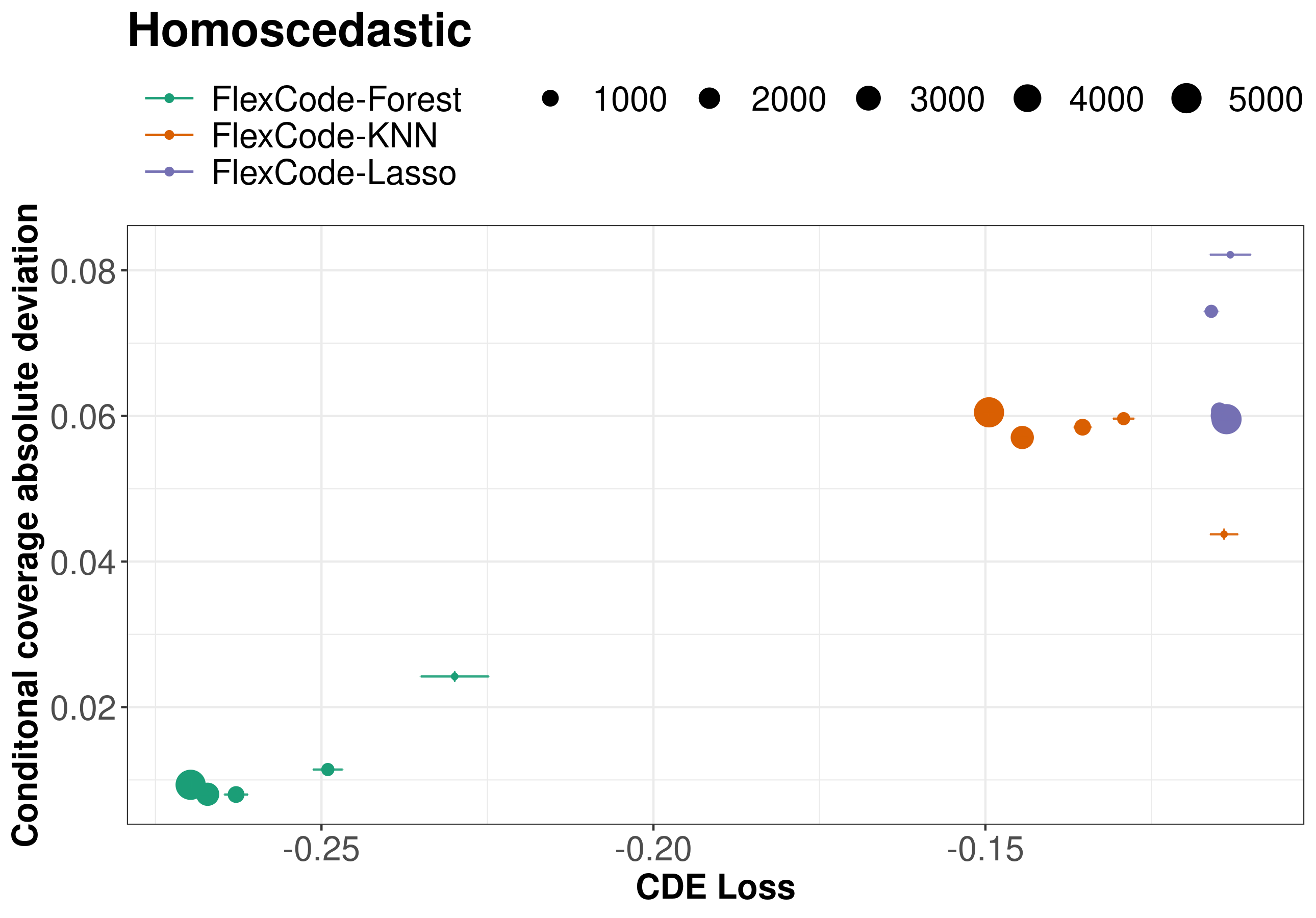}
 \hspace{2mm} 
 \includegraphics[width=0.45\textwidth]
 {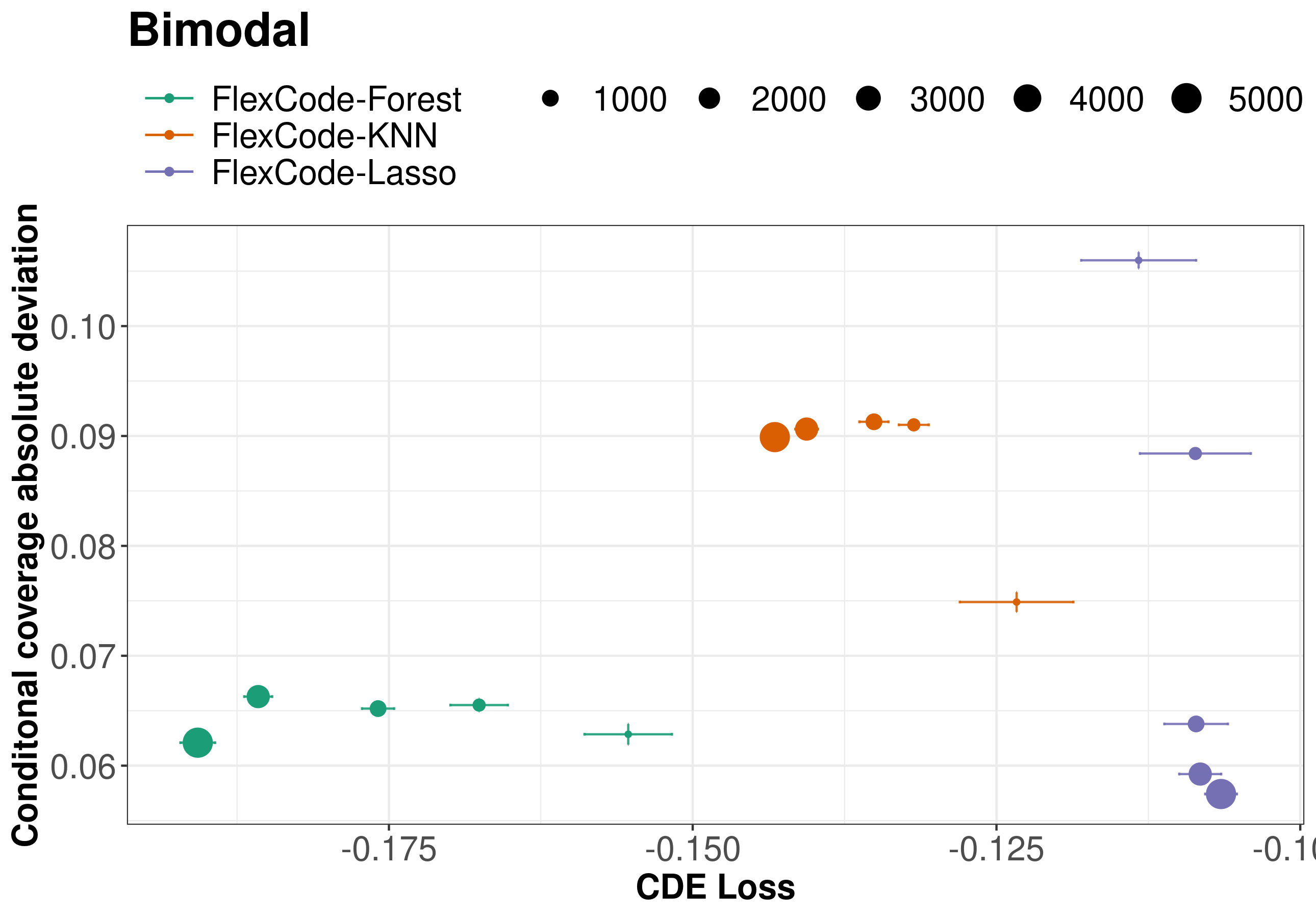}\\
 \includegraphics[width=0.45\textwidth]
 {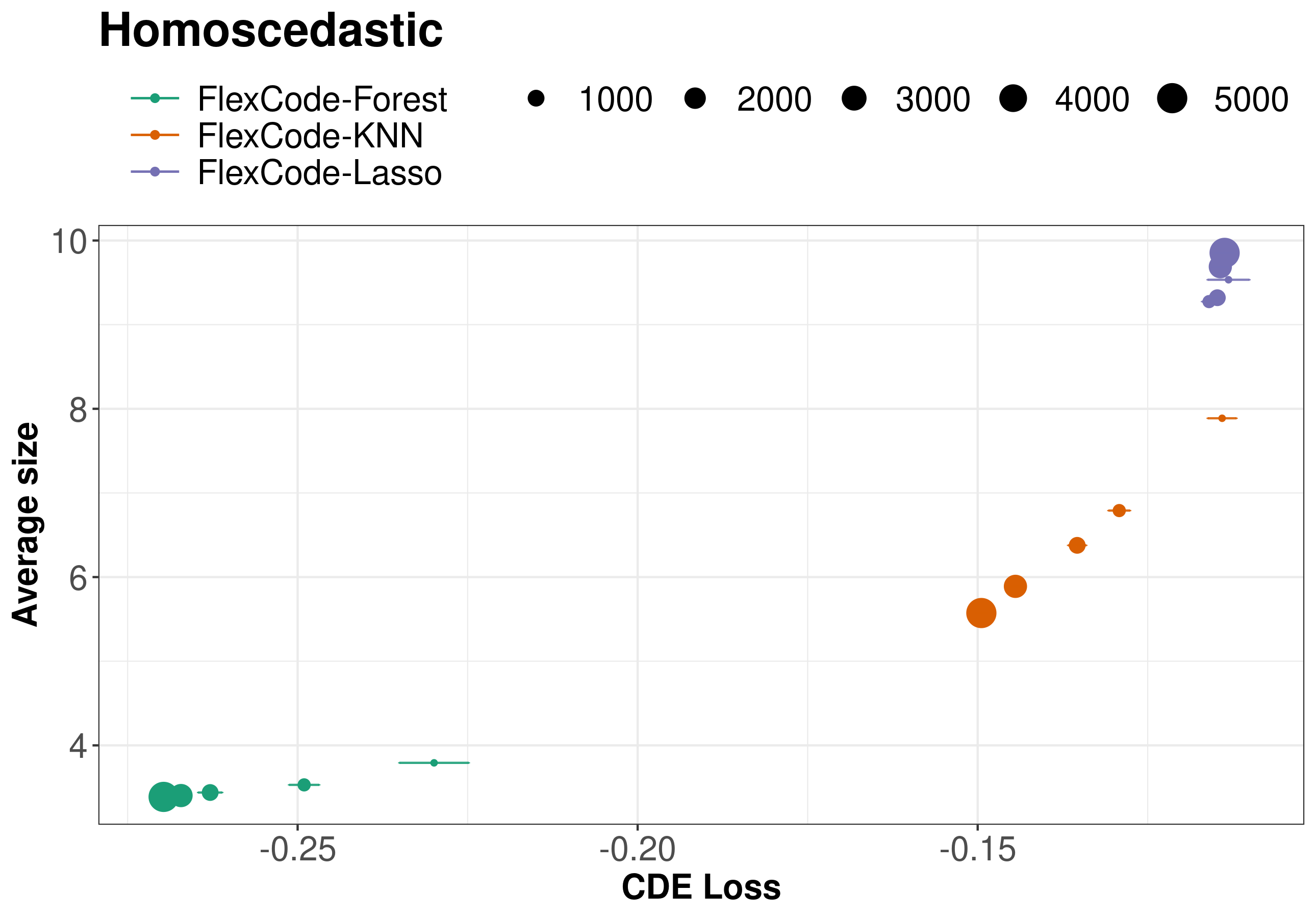}
 \hspace{2mm} 
 \includegraphics[width=0.45\textwidth]
 {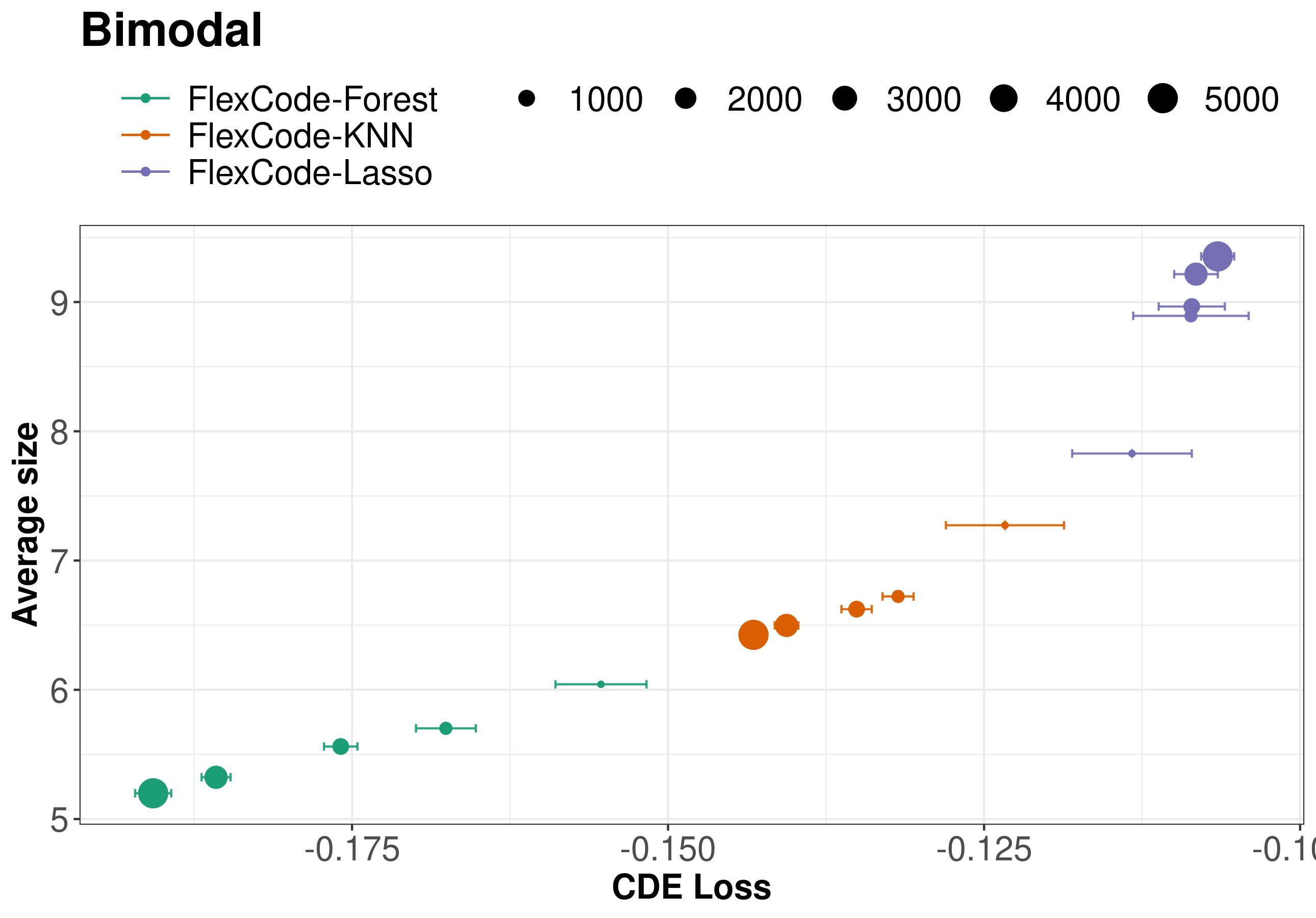}
 \vspace{-2mm} 
 \caption{Performance of the \cdsplitp \ with
 respect to conditional coverage (upper panel)
 and region size (lower panel)
 as a function of the estimated 
 conditional density loss (CDE loss).
 Each point is a different combination of 
 sample size and estimator.} 
 \label{fig:regression_th} 
\end{figure}

\cdsplitp \ also requires tuning with
respect to the conditional density estimator.
We test this type of tuning by fitting 
\flexcode \ coupled with
the following regression methods:
random forests, knn, and lasso.
We also investigate 5 different sample sizes.
For each density estimator, we estimated
the conditional density loss (CDE loss),
$\int \left(\hat f(y|\x)-f(y|\x) \right)^2 dP(\x)dy$ \citep{Izbicki2016}.
\Cref{fig:regression_th} shows that,
in homoscedastic and bimodal scenarios, 
the CDE loss is strongly associated with
the conditional coverage and 
region size of \cdsplitp .
That is, conditional density estimates with
a smaller loss lead to smaller prediction bands
with a better conditional coverage.
The only exception occurs in the bimodal scenario,
in which although for large sample sizes
\flexcode-lasso has a high CDE loss,
it also has a small conditional coverage deviation.
\Cref{fig:regression_th_2} in the Appendix shows that
the heteroscedastic and asymmetric scenarios
behave similarly as the bimodal scenario.
These observations lead to the conclusion that
a practical procedure for
obtaining good prediction bands is to
choose the conditional density estimator
with the smallest estimated CDE loss.

\subsection{Comparison to Other Conformal Methods}
\label{sec:comparison}

Next, we compare  \cdsplitp
\ and \hpdsplit \ to some
previously proposed methods:
\begin{itemize}[wide, labelwidth=!, labelindent=0pt]
 \item \textbf{[Reg-split]} 
 The regression-split method \citep{Lei2018}, 
 based on the conformal score 
 $|Y_i-\widehat{r}(\x_i)|$, where 
 $\widehat{r}$ is an estimate of
 the regression function.
 
 \item \textbf{[Local Reg-split]} 
 The local regression-split method \citep{Lei2018}, 
 based on the conformal score $\frac{|Y_i-\widehat{r}(\x_i)|}{\widehat{\rho}(\x_i)}$,
 where $\widehat{\rho}(\x_i)$ is
 an estimate of the conditional mean
 absolute deviation of $|Y_i-r(\x_i)|\x_i$.

 \item \textbf{[Quantile-split]} 
 The conformal quantile regression method \citep{Romano2019,Sesia2019}, based on
 conformalized quantile regression. 

 \item \textbf{[Dist-split]} 
 The conformal method from 
 \citet{Izbicki2020} that uses 
 the cumulative distribution function,
 $F(y|\x)$, to create prediction intervals.
 
 \item \textbf{[CD-split$^+$]} 
 From \cref{sec:cdsplitp} with
 partitions of size $\ceil{\frac{n}{100}}$.
\end{itemize}

Each experiment is performed with
comparable settings in each method. 
For instance, random forests \citep{Breiman2001}
estimate all required quantities:
the regression function in Reg-split,
the conditional mean absolute deviation 
in Local Reg-split, 
the conditional quantiles via 
quantile forests \citep{Meinshausen2006} 
in Quantile-split,
and the conditional density 
via FlexCode \citep{Izbicki2017}
in \distsplit\ and \cdsplitp .
A conditional cumulative distribution estimate,
$\widehat{F}(y|\x)$ is obtained by 
integrating the conditional density estimate,
that is, $\widehat{F}(y|\x) = 
\int_{-\infty}^y \widehat{f}(y|\x)dy$.
The tuning parameters of all methods were 
the default of the packages.

\begin{figure}
 \centering
 \includegraphics[scale=0.25]{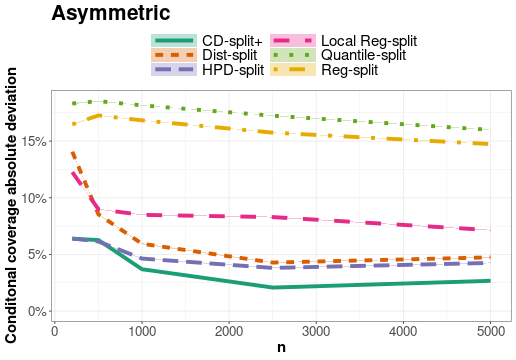}
 \hspace{2mm}
 \includegraphics[scale=0.25]{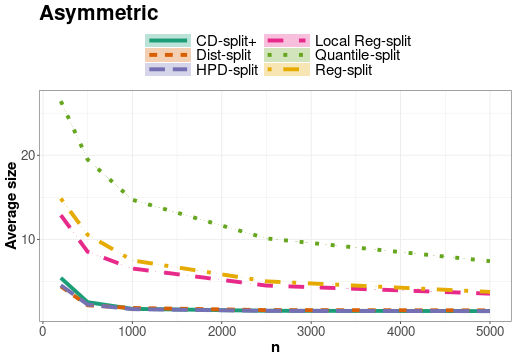} \\ [2mm]
 \includegraphics[scale=0.25]{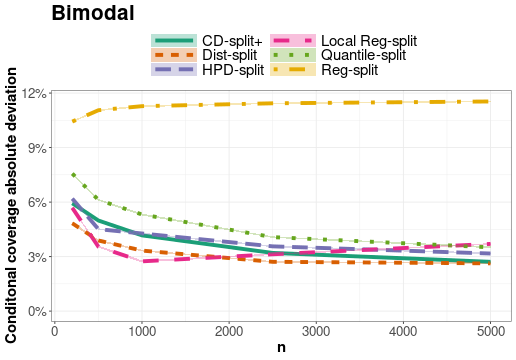}
 \hspace{2mm} 
 \includegraphics[scale=0.25]{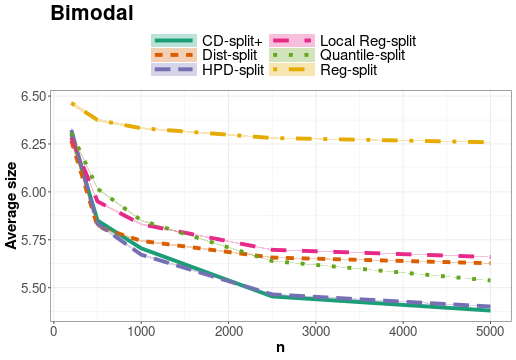}\\[2mm]
 \includegraphics[scale=0.25]{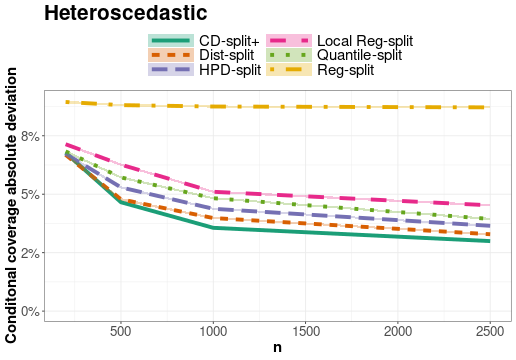}
 \hspace{2mm} 
 \includegraphics[scale=0.25]{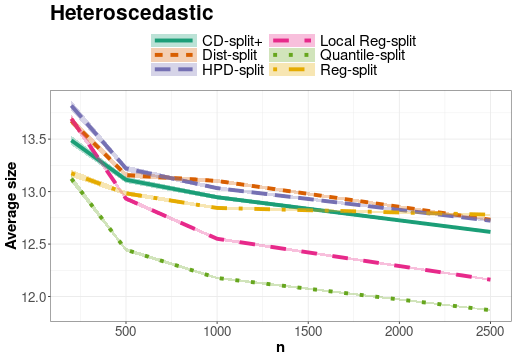} \\[2mm]
 \includegraphics[scale=0.25]{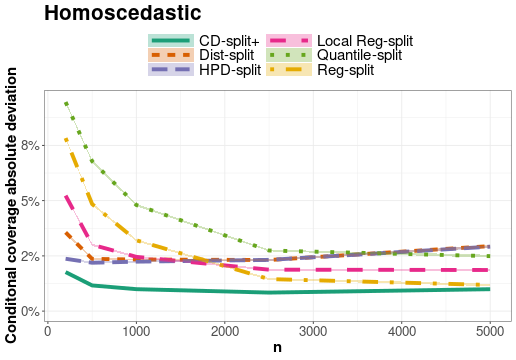}
 \hspace{2mm} 
 \includegraphics[scale=0.25]{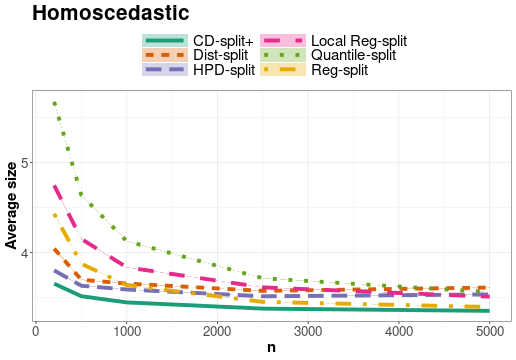}
 \vspace{-2mm} 
 \caption{Conditional coverage (left panel)
 and average size of prediction bands (right panel)
 for each conformal method
 as a function of the sample size.} 
\label{fig:regression} 
\end{figure}

\Cref{fig:regression} shows the performance of 
each method as a function of the sample size.
While the left side figures display how well
each method controls conditional coverage,
the right side displays the average size
of the obtained predictive regions.
\Cref{fig:regression} shows that,
in all settings, \cdsplitp \ is
the method which best 
controls conditional coverage.
Also, in most cases its 
prediction bands also have the smallest size.
The only exception occurs 
in the heteroscedastic scenario,
in which \cdsplitp \ trades 
a larger prediction band for
improved conditional coverage.
In general, \hpdsplit\ is also very competitive, 
although it is slightly outperformed 
in all scenarios by \cdsplitp .

\subsection{Effect of dimensionality on performance}
\label{sec:high_dim}

How are the \cdsplitp \ and 
\hpdsplit \ affected by
an increase in the dimension of the feature space?
We study this question by fixing the sample size at
$n = 1000$ and studying the performance of
the proposed methods in the
scenarios used in the previous sections as
the number of irrelevant features
increases from $d = 0$ to $5000$.
These simulations allow a better understanding
of the performance of the proposed methods in high-dimensional settings than
in the previous sections, in which
the number of irrelevant features was fixed at $d = 20$.

\begin{figure}
 \centering
 \includegraphics[scale=0.25]{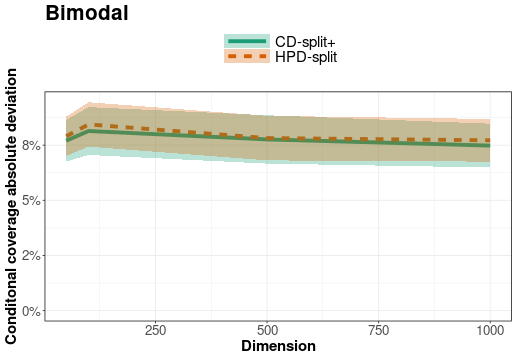}
 \hspace{2mm} 
 \includegraphics[scale=0.25]{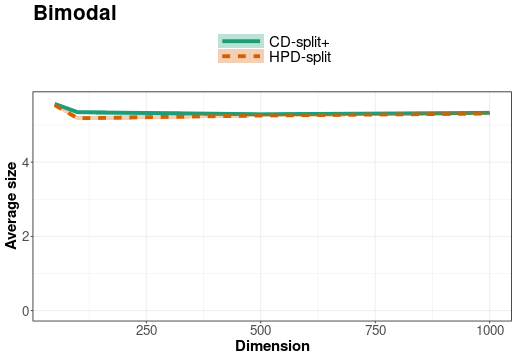}\\[2mm]
 \includegraphics[scale=0.25]{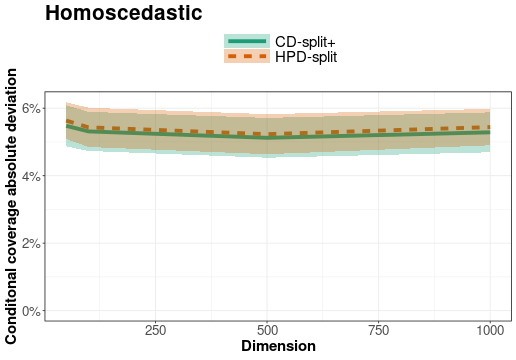}
 \hspace{2mm} 
 \includegraphics[scale=0.25]{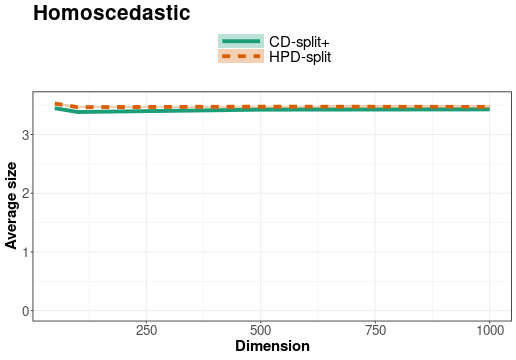}
 \vspace{-2mm} 
 \caption{Conditional coverage (left panel)
 and average size of prediction bands (right panel)
 for \cdsplitp\ and \hpdsplit \
 as a function of the number of features. None of the methods are heavily affected by increasing dimensionality.} 
 \label{fig:dimension}
\end{figure}

\Cref{fig:dimension} shows that 
neither \cdsplitp \ or \hpdsplit \ are
heavily affected by 
the dimensionality of the feature space
in the bimodal and homoscedastic scenarios.
The same conclusion also applies to
the asymmetrical and heteroscedastic scenarios,
as shown in \cref{fig:dimension_2} 
in the Appendix.
This observation can be explained by
the fact that the performance of these methods
relies mainly on the quality of
the conditional density estimator.
In this experiment, this estimation is performed by 
FlexCode-Random Forest, which
automatically performs variable selection.
As a result,  the density estimates and
conformal predictions are reasonable
even when there is a large number of
irrelevant features, as
expected by the empirical and theoretical findings 
in \citet{Izbicki2017}.

\subsection{Classification}
\label{sec:classification}

This section studies the performance of
\cdsplitp \ and \hpdsplit \ when
applied to conformal classification.

\begin{figure}
 \centering
 \includegraphics[width=0.45\textwidth]
 {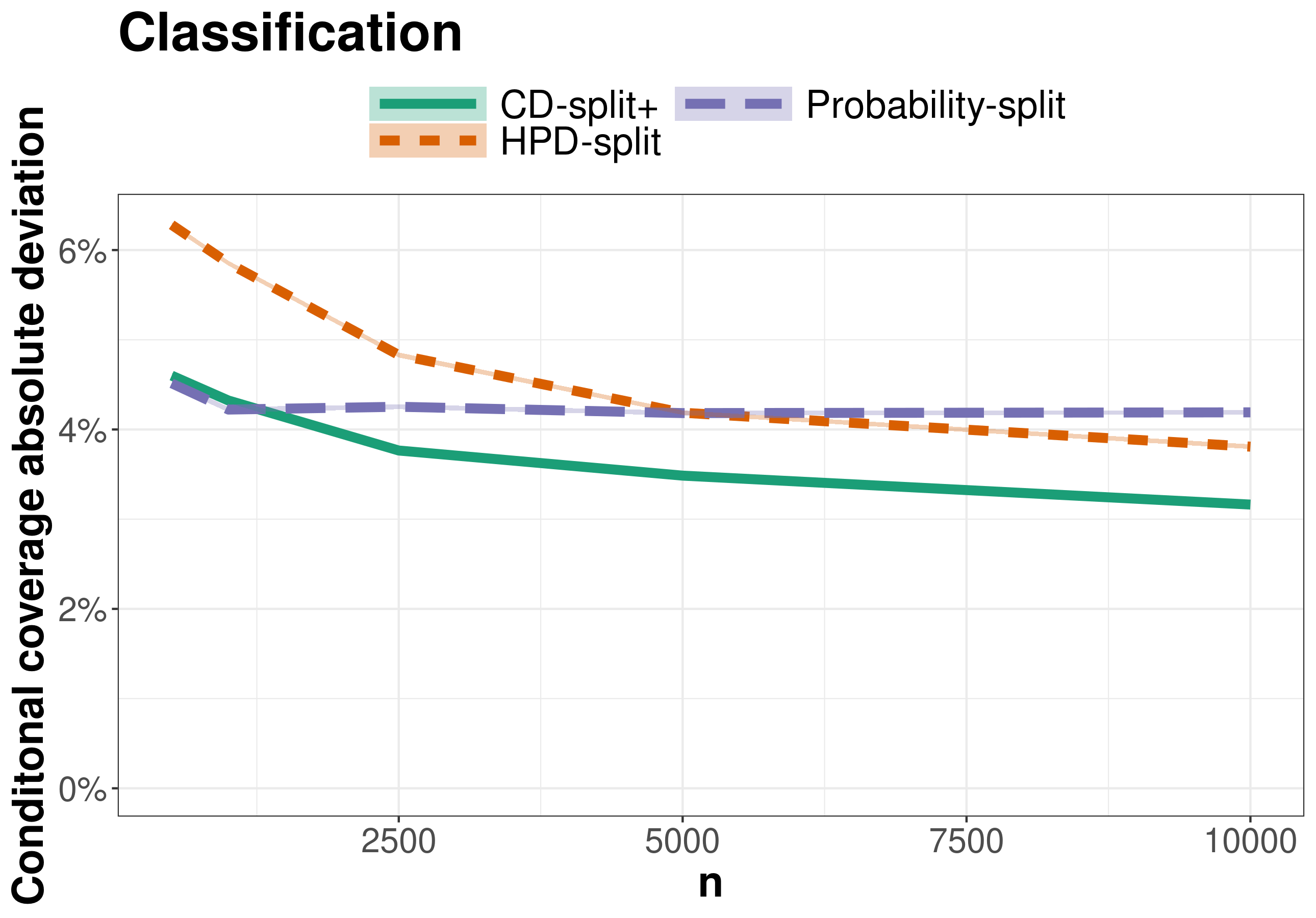}
 \hspace{2mm}
 \includegraphics[width=0.45\textwidth]
 {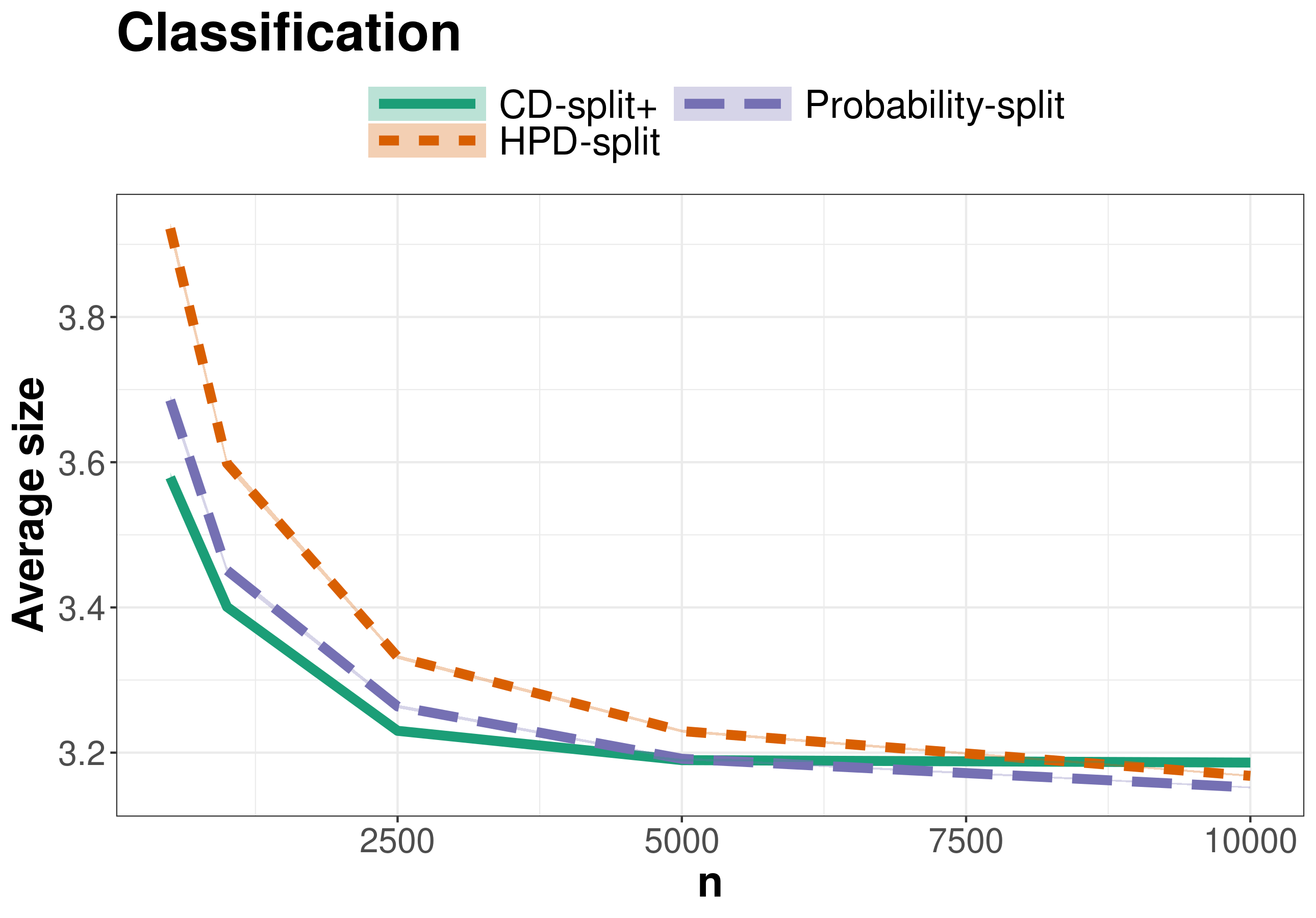}
 \includegraphics[width=0.45\textwidth]
 {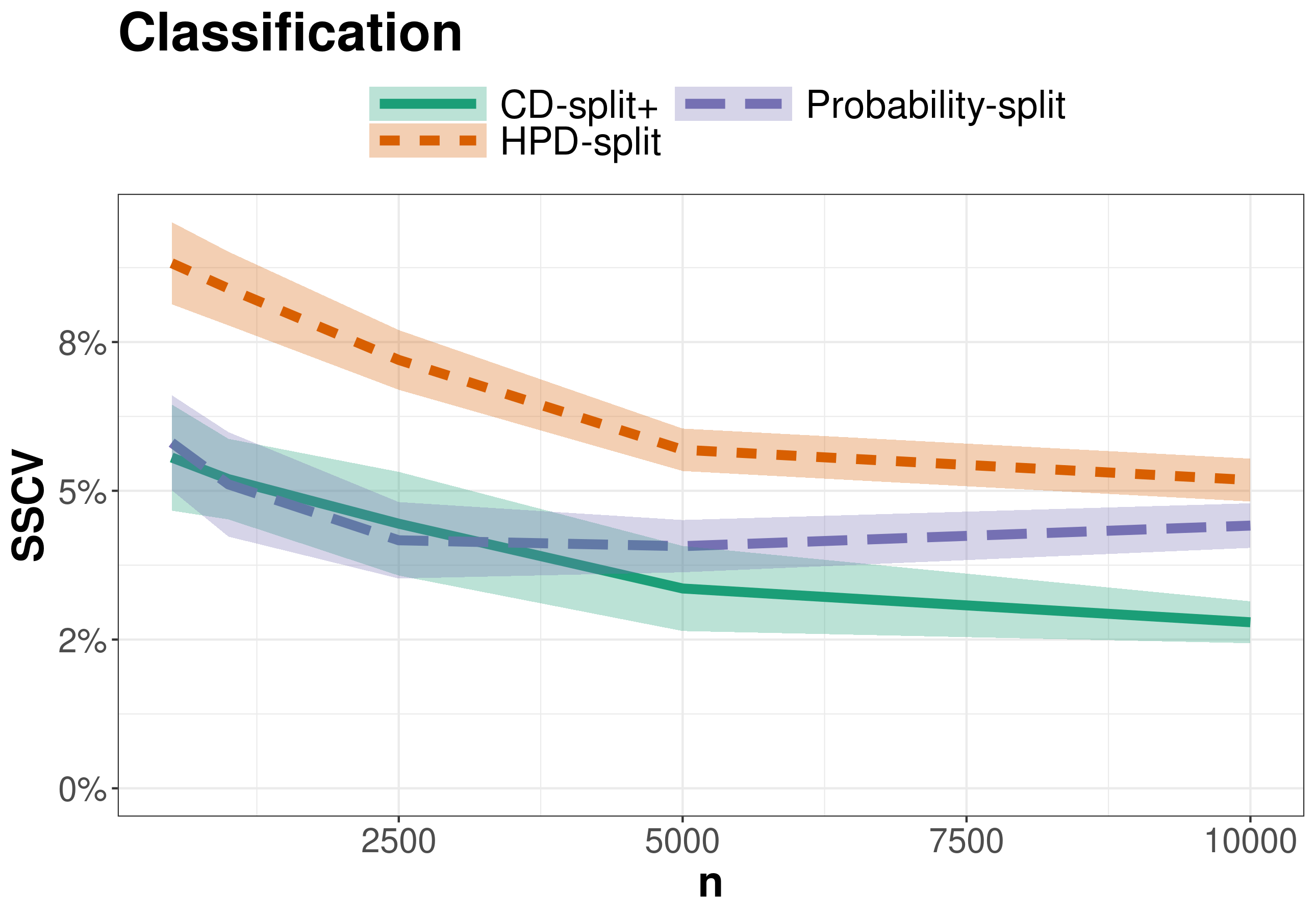}
 \vspace{-2mm}
 \caption{
  Conditional coverage (top left),
   average predictive region size (top right), and size
stratified coverage violation (SSCV; bottom)
  of each conformal method as 
  a function of the sample size.} 
 \label{fig:classification} 
\end{figure}

First,
a simulation study
compares \cdsplitp \ and \hpdsplit \ to
Probability-split \citep[Sec. 4.3]{Sadinle2019},
a particular case of \cdsplitp \ 
with a unitary partition.
We consider that $\X=(X_1,\ldots,X_d)$,
with $X_i \overset{\text{iid}}{\sim} N(0,1)$
and $Y|\X$ follows the logistic model,
$\P(Y=i|\x) \propto 
\exp\left\{\beta_i \cdot x_1 \right\}$,
where $\boldsymbol{\beta}
= (-6, -5, -1.5, 0, 1.5, 5, 6)$. 
\Cref{fig:classification} shows that,
while Probability-split can attain
slightly smaller predictive bands
than the other methods,
However, \cdsplitp \ yields 
better conditional coverage as measured by 
the conditional coverage absolute deviation (top left panel) and 
also by the size stratified coverage violation
(\citet{Angelopoulos2020}, bottom panel). 
In this scenario, \hpdsplit \ gives 
larger prediction bands and 
does not control conditional coverage 
as well as the other methods.

\begin{figure}
 \centering
 \includegraphics[width=0.3\textwidth,page=1]
 {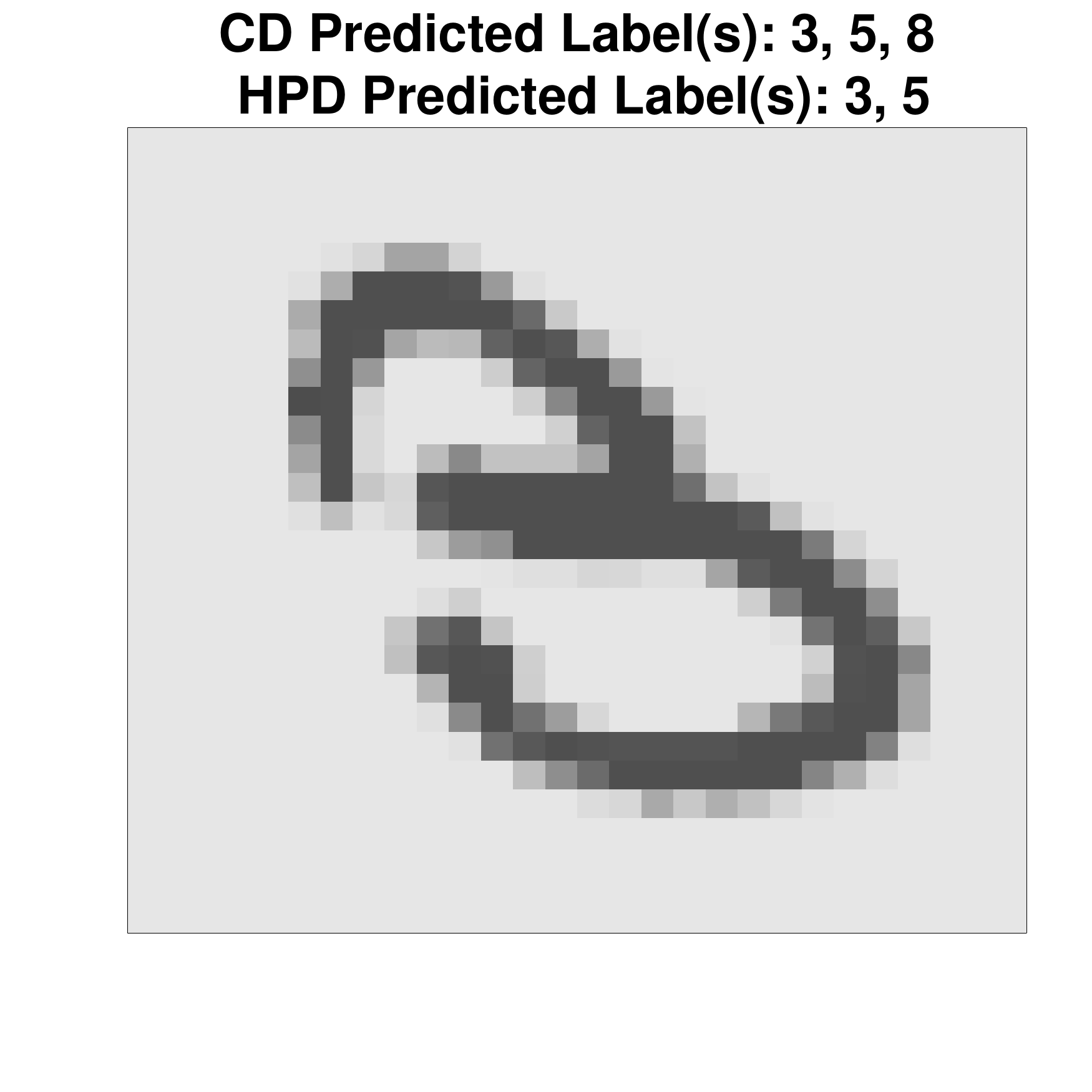}
 \includegraphics[width=0.3\textwidth,page=2]
 {figures/mnist_hpd}
 \includegraphics[width=0.3\textwidth,page=3]
 {figures/mnist_hpd}
 \vspace{-4mm}
 \includegraphics[width=0.3\textwidth,page=4]
 {figures/mnist_hpd}
 \includegraphics[width=0.3\textwidth,page=5]
 {figures/mnist_hpd}
 \includegraphics[width=0.3\textwidth,page=6]
 {figures/mnist_hpd}
 \caption{Prediction bands given by \cdsplitp \ and \hpdsplit \
 for some instances on MNIST.} 
 \label{fig:mnist} 
\end{figure}

\cdsplitp \ and \hpdsplit \ are also applied to 
the MNIST data set \citep{Lecun1995}.
The data is divided in three sets:
9\% as potential future samples,
70\% to estimate $\P(y|\x)$, and
21\% to calculate split residuals.
The conditional density, $\P(Y=y|\x)$, 
is estimated using a convolutional neural network.
\Cref{fig:mnist} shows six examples of images and 
their respective predictive bands. 
The top row displays examples where 
two labels were assigned to each data point.
These instances generally seem 
ambiguous for humans.
Both approaches lead to similar results.

\section{Application to photometric redshift prediction}
\label{sec:photo-z}

We apply \cdsplitp \ and \hpdsplit \ to 
a key problem in cosmology: estimating the redshift of a galaxy ($y$) based on 
its photometric features and 
$r$-magnitude ($\x$). 
This is an important problem since,
for instance, redshift is
a proxy for the distance between
the galaxy and Earth.
In this type of problem 
$f(y|\x)$ is expected to be 
highly asymmetrical and often multimodal and,
therefore, regression-based predictors are 
are uninformative  
\citep{sheldon2012photometric,carrasco2013tpz,izbicki2017photo,Freeman2017}. 
This challenge has motivated 
the development of several methods for 
photo-z density estimation 
\citep{schmidt2020evaluation}.

Here we construct prediction bands for redshifts using   the Happy A dataset \citep{beck2017realistic}, 
which is designed to 
compare photometric 
redshift prediction algorithms. 
Happy A contains
74,950 galaxies based on
the Sloan Digital Sky Survey DR12.
We use 64,950 galaxies as the training set,
5,000 as the prediction set, and
5,000 for comparing 
the performance of conformal methods.
In the training set,
the conformity of
\regsplit \ and \quantilesplit\ is
fit using random forests.
Also, for density-based methods 
(\hpdsplit, \cdsplitp \ and \distsplit),
the conformal score is fit
using a Gaussian mixture density 
neural network \citep{bishop1994mixture} 
with three components, 
one hidden layer and 500 neurons.
Using this mixture model,
\cdsplitp \ and \hpdsplit \ yield
predictive regions that
are the union of at most $3$ intervals.
All methods use a 
marginal coverage level of $1-\alpha=80\%$.

\Cref{tab:photoz} shows the 
local coverage and average size of 
the prediction bands in 
two regions of the feature space:
bright galaxies and faint galaxies.
This classification is based on
the galaxy's $r$-magnitude value:
low for bright galaxies and
high for faint ones.
The table shows that, 
while on bright galaxies
all methods obtain a
local coverage close to the 
nominal 80\% level,
the same does not happen 
for faint galaxies. 
Nominal coverage on faint galaxies
is only obtained for
\hpdsplit\ and \cdsplitp,
which therefore better control conditional coverage
in this application.
The table also shows that, except for
\regsplit, all methods yield
regions with similar size.
\regsplit \ obtains smaller regions,
at the cost of a worse 
local coverage among faint galaxies.

\begin{figure}[hp!]
 \centering
 \includegraphics[scale=0.4]{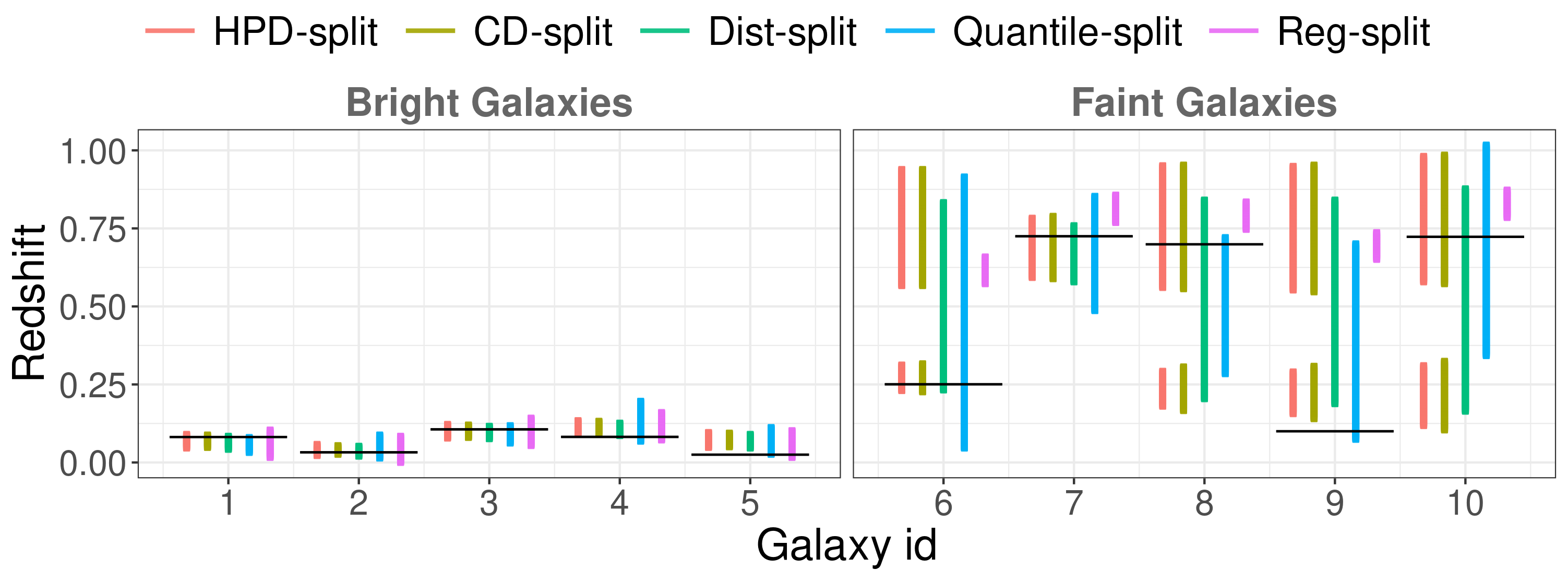}
 \caption{Prediction bands obtained for 5 bright and 5 faint  galaxies from the test set. Horizontal lines indicate the true redshift of each galaxy. \hpdsplit\  and \cdsplitp\ offer smaller bands than the other methods for faint galaxies  (which is where typically $f(z|\x)$ is multimodal).} 
\label{fig:photo-z} 
\end{figure}

\Cref{fig:photo-z} shows examples 
of the prediction regions of each method
along with an horizontal line indicating
the true redshift in each instance. 
All prediction regions increase in size 
on faint galaxies since they
typically have multimodal densities
\citep{wittman2009lies,kugler2016spectral,polsterer2016dealing,dalmasso2020}.
Indeed, in the selected
instances of faint galaxies, 
the true redshift falls either 
close to $0.25$ or to $0.75$,
but never close to $0.5$.
This multimodal behavior indicates that
the central portion of the
prediction intervals obtained in
$\distsplit$ and in $\quantilesplit$ 
commonly contains unlikely estimates
for the true redshift.

\begin{table}[H]
\centering
\caption{Coverage and average size of the prediction bands for the photometric redshift prediction problem, along with their standard errors.  \hpdsplit \ and \cdsplitp \ yield smaller prediction sets among the methods that have better control of conditional coverage.}
{\small
\begin{tabular}{ll|lllll}
            & Galaxy  & \hpdsplit & \cdsplitp       & \distsplit & \quantilesplit & \regsplit       \\ \hline
\multirow{2}{*}{Coverage} & Bright  &  0.800 (0.006)&0.795 (0.006)&0.802 (0.006)&0.808 (0.006)&0.807 (0.006)
\\
&   Faint      & 0.788 (0.018)&0.792 (0.018)&0.746 (0.019)&0.754 (0.019)&0.658 (0.021) \\ \hline 
\multirow{2}{*}{\shortstack[c]{Average \\ Size}} & Bright  &0.050 (0.001)&0.049 (0.001)&0.050 (0.001)&0.051 (0.001)&0.044 (0.000) \\
&   Faint      & 0.065 (0.001)&0.066 (0.001)&0.064 (0.002)&0.074 (0.002)&0.045 (0.000)
\end{tabular}
\label{tab:photoz}
}
\end{table}

\section{Final Remarks}
\label{sec:final}

We propose two new conformal methods:
\cdsplit \ and \hpdsplit. 
One of the features of these methods is
that they can yield 
arbitrary prediction regions.
Therefore, whenever 
the target variable is
for instance multimodal,
the proposed prediction regions
can be smaller and have
better conditional coverage than 
existing alternatives.
Both methods are
based on conditional density estimators.
Since these estimators are available
either when the target variable is
discrete (conditional pmf) or 
continuous (conditional pdf),
\cdsplit \ and \hpdsplit \ 
provide a unified framework to
conformal regression and to
conformal classification.

From a theoretical perspective,
we provide two types of results.
If the instances are exchangeable, then
the proposed methods achieve
marginal validity and 
\cdsplit \ obtains local validity.
Under additional assumptions, 
such as the consistency of
the conditional density estimator,
we prove that,
even in high-dimensional feature spaces,
\cdsplit \ obtains 
asymptotic conditional coverage and
converges to the highest predictive
density set.
Neither of these results require 
assumptions about the type of
dependence between
the target variable and the features.

Simulations and applications to
real data also point out that
the proposed methods perform well.
In most of the simulations
the proposed methods obtain
smaller region sizes and
better conditional coverage than
other methods in the literature.
This high performance occurs both
in conformal regression and
in conformal classification settings.
It is also occurs when
the feature is high-dimensional.
Furthermore, the proposed methods
also yield satisfactory results when
predicting photometric redshift.
In this application, 
the target variable is often bimodal.
As a result, \cdsplit \ and
\hpdsplit \ yield prediction regions which
are often a union of two intervals.
These predictions are more informative than
the ones obtained by a single interval.

Although \cdsplit \ and \hpdsplit \ can
generally outperform interval-based
predictions with respect to
region size and conditional coverage,
this performance can come at
the cost of a more difficult interpretation.
It is usually much easier to interpret that
the target falls between two values than
that it falls in an arbitrary region.
A middle ground can be obtained by
using, for instance, density estimators
based on mixture models, which
restrict the prediction regions in
\cdsplit \ and \hpdsplit \ to
the union of a fixed number of intervals.

We also show that \cdsplit \ can
be made more stable by considering
different partitions, 
such as in \cdsplitp .
This method is based on
a novel data-driven metric on 
the feature space that
defines neighborhoods for conformal methods
which perform well even in
high-dimensional settings.
It might be possible to use this metric with
other conformal methods to obtain
better conditional coverage.

R code for implementing 
\cdsplit , \cdsplitp \ and \hpdsplit \ 
is available at:
\begin{center}
\url{https://github.com/rizbicki/predictionBands}
\end{center}

\section*{Acknowledgements}

This study was financed in part by 
the Coordena\c{c}\~{a}o de Aperfei\c{c}oamento 
de Pessoal de N\'{\i}vel Superior - Brasil (CAPES) 
- Finance Code 001. 
Rafael Izbicki is grateful for 
the financial support of FAPESP (grant 2019/11321-9) and 
CNPq (grant 309607/2020-5).
Rafael B. Stern produced this work as
part of the activities of FAPESP Research, Innovation 
and Dissemination Center for Neuromathematics
(grant 2013/07699-0).

\bibliography{main}

\appendix
\section{Proofs}

The proofs are organized 
in subsections.
\Cref{sec:hpd_to_as} 
proves that
convergence to the hpd implies
asymptotic conditional validity
(\cref{thm:triangle_as_equiv}).
\Cref{sec:marginal_validity} proves
all results related to
marginal and local validity
(\cref{thm:local_Validity,thm:cd_split_control,thm:hpd_marginal}).
\Cref{sec:aux} proves several
auxiliary lemmas that
are used for proving
the asymptotic validity of
\cdsplit, \cdsplitp, and
\hpdsplit.
\cref{sec:hpd_hpd} proves
that \hpdsplit \ converges
to the hdp set and satisfies
asymptotic conditional validity
(\cref{thm:hpd_hpd}).
Similar results
(\cref{thm:cd_hpd,thm:cdp_hpd})
are proved for
\cdsplit \ and \cdsplitp \ 
respectively in
\cref{sec:cd_hpd,sec:cdp_hpd}.

\subsection{Proofs of \cref{thm:triangle_as_equiv}}
\label{sec:hpd_to_as}

\begin{lemma_2}
 \label{lemma:triangle_as_equiv}
 If there exists $C_{\alpha}^*(\X_{n+1})$ such that
 $\P(Y_{n+1} \in C_{\alpha}^*(\X_{n+1})|\X_{n+1}) \equiv 1-\alpha$
 and also such that
 $\P(Y_{n+1} \in C_{\alpha}^*(\X_{n+1}) \triangle C_{\alpha}(\X_{n+1}))
 = o(1)$, then $C_{\alpha}^*(\X_{n+1})$ satisfies
 asymptotical conditional validity.
\end{lemma_2}

\begin{proof}
 Since $\P(Y_{n+1} \in C_{\alpha}^*(\X_{n+1}) \triangle C_{\alpha}(\X_{n+1}))
 = o(1)$, it follows from Markov's inequality and
 the dominated convergence theorem that
 $\P(Y_{n+1} \in C_{\alpha}^*(\X_{n+1}) 
 \triangle C_{\alpha}(\X_{n+1})|\X_{n_1}) = o_{\P}(1)$.
 Therefore, there exists $\eta_n = o(1)$ such that,
 for $\Lambda_n^c = \{\P(Y_{n+1} \in C_{\alpha}^*(\X_{n+1}) 
 \triangle C_{\alpha}(\X_{n+1})|\X_{n+1}) > \eta_n\}$,
 one obtains $\P(X_n \in \Lambda_n^c) = o(1)$.
 Conclude that $\P(X_n \in \Lambda_n) = 1-o(1)$ and that
 \begin{align*}
  &\sup_{\x_{n+1} \in \X_{n+1}[\Lambda_n]} \bigg| \P(Y_{n+1} \in C_{\alpha}(\x_{n+1})|\X_{n+1}=\x_{n+1})-1-\alpha \bigg| \\
  \leq& 
  \sup_{\x_{n+1} \in \X_{n+1}[\Lambda_n]}
  \bigg| \P(Y_{n+1} \in C_{\alpha}^*(\x_{n+1}) \triangle C_{\alpha}(\x_{n+1})|\X_{n+1}=\x_{n+1}) \bigg|
  \leq \eta_n = o(1)
 \end{align*}
\end{proof}

\begin{proof}[Proof of \cref{thm:triangle_as_equiv}]
 Follows directly from \cref{lemma:triangle_as_equiv}.
\end{proof}

\subsection{Proofs related to
local and marginal validity}
\label{sec:marginal_validity}

\begin{proof}[Proof of   \cref{thm:local_Validity}]
 Note that $\hat{g}$ is 
 a function of $\D'$.
 Therefore, $\hat{g}(\X_i,Y_i)$
 can be written as $u(\X_i,Y_i,\D')$.
 Since the instances in 
 $\D$ are exchangeable, 
 it follows that
 $U_i := u(\X_i,Y_i,\D')$
 are exchangeable.
 Therefore, 
 for every $A \in \mathcal{A}$,
 $\{U_i: \X_i \in A \}$ are
 exchangeable.
 In particular, given that
 $\X_{n+1} \in A$,
 $\{U_i: \X_i \in A(\X_{n+1})\}$
 are exchangeable.
 That is, 
 for every $A \in \mathcal{A}$,
 \begin{align*}
  \P\left(U_{n+1} \geq  U_{\lfloor\alpha\rfloor}(\X_{n+1})|
  \X_{n+1} \in A\right) 
  &\geq 1-\alpha \\
  \P\left(\hat{g}(\X_{n+1}, Y_{n+1}) \geq  U_{\lfloor\alpha\rfloor}(\X_{n+1})|
  \X_{n+1} \in A\right) 
  &\geq 1-\alpha \\
  \P\left(Y_{n+1} \in
  \{y: \hat{g}(\X_{n+1}, y) \geq  U_{\lfloor\alpha\rfloor}(\X_{n+1})\}
  |\X_{n+1} \in A\right)
  &\geq 1-\alpha \\
  \P(Y_{n+1} \in C(\X_{n+1})|
  \X_{n+1} \in A)
  &\geq 1-\alpha
 \end{align*}
 Conclude that
 $C(\X_{n+1})$ satisfies
 local validity with
 respect to $\mathcal{A}$ and
 also satisfies, in particular, 
 marginal validity.
\end{proof}

\begin{proof}[Proof \cref{thm:cd_split_control}]
 Follows directly from
 \cref{thm:local_Validity}.
\end{proof}

\begin{proof}[Proof of \cref{thm:hpd_marginal}]
 Follows directly from
 \cref{thm:local_Validity} by
 taking $\mathcal{A} = \{\sX\}$.
\end{proof}

\subsection{Auxiliary results}
\label{sec:aux}

\begin{lemma_2}
 \label{lemma_cdsplit_1}
 Under \cref{ass:consistent_cde_2},
 \begin{align*}
 \P\left(\sup_{y \in \sY} |\hf(y|\X)-f(y|\X)| 
 \geq \eta_n^{1/3}\right) = o(1).
 \end{align*}
\end{lemma_2}

\begin{proof}
 Let $B_n = \left\{\sup_{y \in \sY} |\hf(y|\X)-f(y|\X)| 
 \geq \eta_n^{1/3}\right\}$ and
 $A_n = \left\{\E\left[\sup_{y \in \sY}
 \left(\hf(y|\X)-f(y|\X)\right)^2 \big|\hf\right] \geq \eta_n\right\}$.
 \begin{align*}
  \P(B_n) 
  &= \E[\P(B_n|\hf) \I(A_n)] 
  + \E[\P(B_n|\hf) \I(A_n^c)] \\
  &\leq \P(A_n) + \E\left[\frac{\E[\sup_{y \in \sY} (\hf(y|\X)-f(y|\X))^2|\hf]}
  {\eta_n^{2/3}}\I(A_n^c)\right] \\
  &\leq \rho_n + \eta_n^{1/3} = o(1)
 \end{align*}
\end{proof}

\begin{lemma_2}
 \label{lemma:hpd_cdf}
 Under \cref{ass:bounded,ass:consistent_cde_2,ass:continuous_hpd},
 $\sup_u |H(u|\X_i)-\hH(u|\X_i)| = o_{\P}(1)$. Also,
 there exists a neighborhood of $\alpha$, 
 $N_{\alpha}$, such that
 $$\sup_{u \in N_\alpha}\
 |H^{-1}(u|\X_i)-\hH^{-1}(u|\X_i)| = o_{\P}(1).$$
\end{lemma_2}

\begin{proof}
 Define that $D = \left\{\sup_{y \in \sY} |\hf(y|\X_i)-f(y|\X_i)|
 \geq \eta_n^{1/3}\right\}$ and
 $A_u = \left\{y: f(y|\X_i) \leq u, \hf(y|\X_i) 
 \leq u\right\}$. 
 If $D^c$ holds, then
 \begin{align}
  \label{eq:cd_dist_1}
  \sup_u\int_{A_u}{|f(y|\X_i)-\hf(y|\X_i)|dy}
  &\leq \bigg|\int_{\sY} \eta_n^{1/3} dy \bigg|
  & D^c \nonumber \\
  &= o_\P(1) & \text{\cref{ass:bounded}}
 \end{align}
 Also let 
 $B_u = \{y: f(y|\X_i) \leq u, \hf(y|\X_i) > u\}$.
 Under $D^c$,
 \begin{align}
  \label{eq:cd_dist_2}
  \sup_u\int_{B_u}f(y|\X_i)dy 
  &\leq \sup_u \int_{\{y: 
  u-\eta_n^{1/3} \leq f(y|\X_i) \leq u\}}
  f(y|\X_i)dy & D^c \nonumber \\
  &= \sup_u |H(u|\X_i)-H(u-\eta_n^{1/3}|\X_i)| \nonumber \\
  &= o_\P(1)
  & \text{\cref{ass:continuous_hpd}}
 \end{align}
 Finally, let $C_u = \{y: f(y|\X_i) > u, \hf(y|\X_i) \leq u\}$.
 Under $D^c$,
 \begin{align}
  \label{eq:cd_dist_3}
  \sup_u \int_{C_u}\hf(y|\X_i)dy
  &\leq \sup_u \int_{\{y: 
  u \leq f(y|\X_i) \leq u + \eta_n^{1/3}\}}
  (f(y|\X_i) + \eta_n^{1/3}) dy 
  & D^c \nonumber \\
  &= \sup_u |H(u + \eta_n^{1/3}|\X_i)-H(u|\X_i)| + o(1)
  & \text{\cref{ass:bounded}} \nonumber \\
  & o_\P(1)
  & \text{\cref{ass:continuous_hpd}}
 \end{align}
 Using the above derivations, observe that under $D^c$
 \begin{align*}
  \sup_u |H(u|\X_i)-\hH(u|\X_i)|
  &= \sup_u \bigg|\int_{\{y:f(y|\x) \leq u\}}{f(y|\x)dy}
  -\int_{\{y:\hf(y|\x) \leq u\}}{\hf(y|\x)dy}\bigg| \\
  &\leq \sup_u \int_{A_u}{|f(y|\X_i)-\hf(y|\X_i)|dy} 
  + \sup_u \int_{B_u}f(y|\X_i)dy 
  + \sup_u \int_{C_u}\hf(y|\X_i)dy \\
  &= o_\P(1), \hspace{5mm}
  \text{\cref{eq:cd_dist_1,eq:cd_dist_2,eq:cd_dist_3}.}
 \end{align*}
 Since \cref{lemma_cdsplit_1} shows that
 $\P(D) = o(1)$, conclude that
 $\sup_u |H(u|\X_i)-\hH(u|\X_i)| = o_{\P}(1)$.
 It follows from \cref{ass:continuous_hpd} that
 there exists a neighborhood of $\alpha$, 
 $N_{\alpha}$, such that
 $$\sup_{u \in N_\alpha}\
 |H^{-1}(u|\X_i)-\hH^{-1}(u|\X_i)| = o_{\P}(1).$$
\end{proof}

\begin{lemma_2}
 \label{lemma:cd_hpd}
 Let $C^*_{\alpha} = \{y:f(y|\X_{n+1}) \geq q_{\alpha}(\X_{n+1})\}$ and
 $C_{\alpha} = \{y:\hf(y|\X_{n+1}) \geq \hat{q}_{\alpha}\}$.
 Under \cref{ass:consistent_cde_2,ass:continuous_hpd},
 if $|q_{\alpha}(\X_{n+1})-\hat{q}_{\alpha}| = o_{\P}(1)$,
 then
 \begin{align*}
  \P\left(Y \in C^*_{\alpha} \triangle C_{\alpha}| \X_{n+1}\right)
  = o_{\P}(1)
 \end{align*}
\end{lemma_2}

\begin{proof}
 Let $B_n$ be such as in \cref{lemma_cdsplit_1}.
 It follows from \cref{lemma_cdsplit_1} that
 $\P(B_n) = o(1)$. 
 Similarly, since $|\qa-\hqa| = o_\P(1)$,
 there exists $\lambda_n = o(1)$ such that
 $D_n := \P(|\qa - \hqa| > \lambda_n) = o(1)$.
 Finally, note that if $E_n^c := (B_n \cup D_n)^c$ holds,
 then $C_{\alpha}^* \triangle C_{\alpha}
 \subseteq \{y: |f(y|\X_{n+1})-\qa| \leq \lambda_n + \eta_n^{1/3}\}$
 Therefore,
 \begin{align*}
  \P\left(Y_{n+1} \in C_{\alpha}^* \triangle C_{\alpha}\right)
  &\leq \P\left(Y_{n+1} \in C_{\alpha}^* \triangle C_{\alpha}, E_n^c\right)
  + \P(E_{n}) \\
  &\leq \P(|f(Y_{n+1}|\X_{n+1})-\qa| \leq \lambda_n + \eta_n^{1/3}\}) + o(1) \\
  &= \E[H(\qa+o(1)|\X_{n+1})
  - H(\qa-o(1)|\X_{n+1})] + o(1) \\
  &= o(1)
  & \text{\cref{ass:continuous_hpd}},
 \end{align*}
 where the last equality uses
 the dominated convergence theorem.
 Since $\P\left(Y_{n+1} \in C_{\alpha}^* \triangle C_{\alpha}\right) = o(1)$,
 it follows from Markov's inequality that
 $\P\left(Y_{n+1} \in C_{\alpha}^* \triangle C_{\alpha}|\X_{n+1}\right) 
 = o_\P(1)$.
\end{proof}

\subsection{Proof of \cref{thm:hpd_hpd}}
\label{sec:hpd_hpd}

\begin{lemma_2}
 \label{lemma:hpd_1}
 There exists $\lambda_n = o(1)$ such that,
 under \cref{ass:iid,ass:bounded,ass:consistent_cde_2,ass:continuous_hpd},
 $|I_1| = n + o_\P(n)$ and
 $|I_2| = o_\P(n)$, where
 $I_1 := \left\{i \leq n: 
 |\hH(\hf(Y_i|\X_i))-H(f(Y_i|\X_i))| \leq \lambda_n\right\}$ and
 $I_2 = \{1,2,\ldots,n\}-I_1$.
\end{lemma_2}

\begin{proof}
 We start by proving that
 $\sup_y |H(f(y|\X_i))-\hH(\hf(y|\X_i))| = o_{\P}(1)$.
 From the triangular inequality, it
 is sufficient to show that
 $\sup_y|H(f(y|\X_i))-H(\hf(y|\X_i))| = o_{\P}(1)$ and
 $\sup_y|H(\hf(y|\X_i))-\hH(\hf(y|\X_i))| = o_{\P}(1)$.
 From \cref{lemma_cdsplit_1} it follows that
 $\sup_y|f(y|\X_i) - \hf(y|\X_i)| = o_\P(1)$ and,
 therefore, from \cref{ass:continuous_hpd},
 $\sup_y|H(f(y|\X_i))-H(\hf(y|\X_i))| = o_{\P}(1)$.
 Also, it follows from \cref{lemma:hpd_cdf} that
 $\sup_u |H(u|\X_i)-\hH(u|\X_i)| = o_{\P}(1)$ and,
 therefore, $\sup_y|H(\hf(y|\X_i))-\hH(\hf(y|\X_i))| = o_{\P}(1)$.
 Conclude that $\sup_y |H(f(y|\X_i))-\hH(\hf(y|\X_i))| = o_{\P}(1)$.
 
 Next, we show that there exists $\lambda_n = o(1)$
 such that $|I_1| = n + o_\P(n)$ and
 $|I_2| = o_\P(n)$.
 Since $\sup_y |H(f(y|\X_i))-\hH(\hf(y|\X_i))| = o_{\P}(1)$,
 conclude that $|H(f(Y_i|X_i))-\hH(\hf(Y_i|X_i))| = o_{\P}(1)$
 and there exists $\lambda_n = o(1)$ such that,
 for $B_n := \{|H(f(Y_i|X_i))-\hH(\hf(Y_i|X_i))| > \lambda_n\}$,
 $\P(B_n) = o(1)$.
 Therefore, from \cref{ass:iid},
 $|I_2| \sim \text{Binomial}(n, \P(B_n))$,
 $|I_2| = o_\P(n)$, and
 $|I_1| = n + o_\P(n)$.
\end{proof}

\begin{lemma_2}
 \label{lemma:hpd_2}
 Let $U_i = \hH(\hf(Y_i|\X_i))$.
 Under \cref{ass:iid,ass:bounded,ass:consistent_cde_2,ass:continuous_hpd},
 for every $\alpha \in (0,1)$,
 $U_{\lfloor \alpha \rfloor} = \alpha + o_{P}(1)
 = U_{\lceil \alpha \rceil}$.
\end{lemma_2}

\begin{proof}
 Let $I_1$ and $I_2$ be 
 such as in \cref{lemma:hpd_1}.
 Also, let $\hat{G}_1$, $G_1$ and $G_0$ be,
 the empirical quantiles of, respectively,
 $\{U_i: i \in I_1\}$,
 $\{H(f(Y_i|\X_i)): i \in I_1\}$, and
 $\{H(f(Y_i|\X_i)): i \leq n\}$.
 By definition of $I_1$,
 for every $\alpha^* \in [0,1]$,
 $\hat{G}_1^{-1}(\alpha^*) = G_1^{-1}(\alpha^*) + o(1)$.
 Also, $G_0^{-1}(\alpha^*) = \alpha^* + o_P(1)$.
 Therefore, since
 \begin{align*}
  G_0^{-1}\left(\frac{|I_1|\alpha^*}{n}\right)
  \leq G_1^{-1}(\alpha^*)
  \leq G_0^{-1}\left(\frac{|I_1|\alpha^*+|I_2|}{n}\right),
 \end{align*}
 conclude that $\hat{G}_1^{-1}(\alpha^*) 
 = \alpha^* + o_P(1)$.
 Finally, since
 \begin{align*}
  \hat{G}^{-1}\left(\frac{n\alpha-|I_2|}{|I_1|}\right)
  \leq U_{\lfloor \alpha \rfloor} \leq 
  U_{\lceil \alpha \rceil} \leq
  \hat{G}^{-1}\left(\frac{n\alpha}{|I_1|}\right),
 \end{align*}
 Conclude that $U_{\lfloor \alpha \rfloor} = \alpha + o_P(1)
 = U_{\lceil \alpha \rceil}$.
\end{proof}

\begin{proof}[Proof of \cref{thm:converge_hpd_hpd}]
 First, note that
 \begin{align}
  \label{eq:hpd_1}
  C(\X_{n+1}) 
  &= \left\{y: \hH(\hf(y|\X_{n+1})) 
  \geq  U_{\lceil \alpha \rceil}\right\} \nonumber \\
  &= \left\{y: \hf(y|\X_{n+1}) 
  \geq  \hH^{-1}(U_{\lceil \alpha \rceil}|\X_{n+1})\right\}.
 \end{align}
 Note that $q_{\alpha}(\x) = H^{-1}(\alpha|\x)$ and
 conclude from \cref{lemma:hpd_2,lemma:hpd_cdf} that
 $|\qa(\x) - \hH^{-1}(U_{\lceil \alpha \rceil}|\X_{n+1})| = o_{\P}(1)$.
 The proof follows from
 \cref{eq:hpd_1,lemma:cd_hpd,thm:triangle_as_equiv}.
\end{proof}

\subsection{Proof of \cref{thm:cd_hpd}}
\label{sec:cd_hpd}

\begin{lemma_2}
 \label{lemma:quantile}
 Let $M \sim \text{Bin}(n,p)$ and 
 $(\X_1,Z_1)$,\ldots,$(\X_M,Z_M)$ be
 i.i.d. continuous random variables such that
 $|F^{-1}_{Z_i}(\alpha|\X_i)-t| = o_{\P}(1)$ and
 $F_{Z_i}$ is continuous and 
 increasing in a neighborhood of $t$.
 If $\hF$ is the empirical cdf of
 $Z_1,\ldots,Z_M$, then 
 $|\hF^{-1}(\alpha)-t| = o_{\P}(1)$.
\end{lemma_2}

\begin{proof}
 Since $|F^{-1}_{Z_i}(\alpha|\X_i)-t| = o_{\P}(1)$,
 there exists $\rho_n = o(1)$ such that one obtains
 $\P(|F^{-1}_{Z_i}(\alpha)-t| > \rho_n) = o(1)$.
 Let $A_n = \{i \leq M: |F^{-1}_{Z_i}(\alpha|\X_i)-t| 
 \leq \rho_n\}$, $t_n$ be the $\alpha$-quantile
 of $Z_i$ given that $Z_i \in A_n$,
 and $\hF_A$ be the empirical cumulative distribution function
 of the $Z_i$ in $A_n$. By construction,
 \begin{align*}
  \hF_A^{-1}\left(\frac{\lfloor (M-|A_n^c|)\alpha \rfloor}
  {M} \right) \leq 
  \hF^{-1}\left(\alpha\right) \leq 
  \hF_A^{-1}\left(\frac{\lceil M\alpha \rceil}
  {M} \right). \text{ Therefore, }
 \end{align*}
 \begin{align*}
  \P\left(|\hF^{-1}(\alpha)-t| > \epsilon\right) 
  &\leq
  \P\left(\max\left(\bigg|\hF_A^{-1}\left(\frac{\lfloor (M-|A_n^c|)\alpha \rfloor}
  {M} \right)-t \bigg|, \bigg|\hF_A^{-1}\left(\frac{\lceil M\alpha \rceil}
  {M} \right)-t\bigg|\right) > \epsilon\right) \\
  &\leq 
  \P\left(\bigg|\hF_A^{-1}\left(
  \frac{\lfloor (M-|A_n^c|)\alpha \rfloor}{M} \right)-t \bigg| \geq 0.5\epsilon\right) + 
  \P\left(\bigg|\hF_A^{-1}\left(\frac{\lceil M\alpha \rceil}
  {M} \right)-t\bigg| > 0.5\epsilon\right) 
 \end{align*}
 Hence, it is enough to show that 
 $\bigg|\hF_A^{-1}\left(\frac{\lfloor (M-|A_n^c|)\alpha \rfloor}{M} \right)-t \bigg|$ and
 $\bigg|\hF_A^{-1}\left(\frac{\lceil M\alpha \rceil}{M} \right)-t\bigg|$ are $o_{\P}(1)$.
 Since the proofs are similar, we show
 only the first case. Furthermore,
 \begin{align*}
  \P\left(\bigg|\hF_A^{-1}\left(\frac{\lfloor (M-|A_n^c|)\alpha \rfloor}{M} \right)-t \bigg| > \epsilon \right)
  &\leq \P\left(\bigg|\hF_A^{-1}\left(\frac{\lfloor (M-|A_n^c|)\alpha \rfloor}{M} \right)-t_n \bigg| > \epsilon - |t-t_n| \right).
 \end{align*}
 It is enough to show that
 \begin{align*}
  \P\left(\hF_A^{-1}\left(\frac{\lfloor (M-|A_n^c|)\alpha \rfloor}{M} \right) < t_n - \epsilon + |t-t_n| \right) 
  &= o(1) & \text{ and } \\
  \P\left(\hF_A^{-1}\left(\frac{\lfloor (M-|A_n^c|)\alpha \rfloor}{M} \right) > t_n + \epsilon - |t-t_n| \right) 
  &= o(1)
 \end{align*}
 Since both cases are similar,
 we show only the former.
 Note that by construction, $|t-t_n| \leq \rho_n = o(1)$,
 and therefore there exists $m_1$ such that, 
 for every $n > m_1$,
 \begin{align}
  \label{eq:quantile-1}
  \P\left(\hF_A^{-1}\left(\frac
  {\lfloor (M-|A_n^c|)\alpha \rfloor}{M} \right) 
  < t_n - \epsilon + |t-t_n| \right)
  &\leq \P\left(\hF_A^{-1}\left(\frac
  {\lfloor (M-|A_n^c|)\alpha \rfloor}{M} \right) 
  \leq t_n - 0.5\epsilon \right) \nonumber \\
  &\leq \P\left(\bigg|\left\{i: Z_i \in A_n \cap 
  Z_i \leq t_n - 0.5\epsilon\right\}\bigg| 
  \geq \frac{\lfloor (M-|A_n^c|)\alpha \rfloor}{M} 
  \cdot |A_n| \right) \nonumber \\
  &= \P\left(Q_n \geq 
  \frac{\lfloor (M-|A_n^c|)\alpha \rfloor}{M} 
  \cdot |A_n| \right),
 \end{align}
 where $Q_n := \bigg|\left\{i: Z_i \in A_n \cap Z_i \leq t_n 
  - 0.5\epsilon\right\}\bigg|$.
 Let $F_{A}(z) := \P(Z_{i} \leq y|Z_i \in A_n)$.
 Since the $Z_i$ are i.id., given $A_n$,
 $Q_n \sim \text{Binomial}\left(|A_n|,
  F_A(t_n-0.5\epsilon)\right)$.
 
 It remains to show that $F_A(t_n-0.5\epsilon)$ is
 close to $F_{Z_i}(t_n-0.5\epsilon)$.
 Since $F_A(z) = 
 \frac{\P(Z_i \leq z \cap Z_i \in A_n)}{\P(Z_i \in A_n)}$,
 obtain 
 $\frac{F_{Z_i}(y)-o(1)}{1-o(1)}
  \leq F_{A}(z) 
  \leq \frac{F_{Z_i}(y)}{1-o(1)}$.
 From these inequalities and observing that
 $t_n = t + o(1)$ and that
 $F_{Z_i}(z)$ is continuous and increasing,
 conclude that
 $F_A(t_n-0.5\epsilon) + o(1) =
 F_{Z_i}(t-0.5\epsilon) < \alpha$.
 That is, there exists $m_2 > m_1$ such that
 for $n > m_2$,
 $F_A(t_n-0.5\epsilon) \leq \alpha_* < \alpha$ and,
 using \cref{eq:quantile-1},
 \begin{align*}
  \P\left(\bigg|\hF_A^{-1}\left(\frac{\lfloor (M-|A_n^c|)\alpha  \rfloor}{M} \right)-t \bigg| > \epsilon \right)
  &\leq \E\left[\P\left(Q_n \geq 
  \frac{\lfloor (M-|A_n^c|)\alpha \rfloor}{M} \cdot |A_n|
  \bigg|A_n\right)\right] \\
  &\leq \E\left[\exp\left(-2|A_n|\left(
  \alpha^* - \frac{\lfloor (M-|A_n^c|)\alpha \rfloor}{M}
  \right)^2\right)\right]
 \end{align*}
 Since $\P(M < 0.5np) = o(1)$ and
 $\frac{|A_n^c|}{M} = o_{\P}(1)$, conclude that
 \begin{align*}
  \P\left(\bigg|\hF_A^{-1}\left(
  \frac{\lfloor (M-|A_n^c|)\alpha\rfloor}{M} \right)-t \bigg| 
  > \epsilon \right) 
  &= o(1).
 \end{align*}
\end{proof}

\begin{proof}[Proof of \cref{thm:cd_hpd}]
 Let $Z_i = \hf(Y_i|\X_i)$,
 $t = H^{-1}(\alpha|\X_{n+1})$, and
 $M = |\mathcal{T}(\X_{n+1},\mathbb{D})|$.
 It follows from \cref{ass:continuous_hpd} that
 $\P(\X_i \in \mathcal{T}(\X_{n+1},\mathbb{D})) = p > 0$.
 Therefore, $M \sim \text{Bin}(n, p)$.
 Also, it follows from \cref{lemma:hpd_cdf} that
 $H^{-1}(\alpha|\X_i) - \hH^{-1}(\alpha|\X_i) 
 = o_{\P}(1)$. Therefore, for every
 $\X_i \in A(\X_{n+1})$,
 $H^{-1}(\alpha|\X_{i})-t = o_{\P}(1)$, that is,
 $F^{-1}_{Z_i}(\alpha)-t = o_{\P}(1)$.
 Since $U_{\lfloor\alpha\rfloor}(\X_{n+1})$
 is the $\alpha$-quantile of the empirical cdf of
 $\{Z_i: i \in A(\X_{n+1})\}$,
 conclude from \cref{lemma:quantile} that
 $|U_{\lfloor\alpha\rfloor}(\X_{n+1})-t| = o_{\P}(1)$.
 The rest of the proof follows directly from
 \cref{lemma:cd_hpd,thm:triangle_as_equiv}.
\end{proof}

\subsection{Proof of \cref{thm:cdp_hpd}}
\label{sec:cdp_hpd}

\begin{lemma_2}
 \label{lemma:l2_equiv_sup}
 Under \cref{ass:bounded,ass:continuous_hpd},
 there exist a universal constants, $C_1$,
 such that, for every $\x_0, \x_1 \in \sX$,
 $C_1 \|H(\cdot |\x_0)-H(\cdot |\x_1)\|_{\infty}^{1.5} 
 \leq \|H(\cdot |\x_0)-H(\cdot |\x_1)\|_{2}$.
 Furthermore, if $\hH$ has the same support as $H$
 and $\|\hH\|_{\infty} < 1$, then
 there exists a universal constant, $C_2$,
 such that
 $\|H(\cdot |\x_0)-\hH(\cdot |\x_1)\|_{2}
 \leq  C_2 \|H(\cdot |\x_0)-\hH(\cdot |\x_1)\|_{\infty}^{0.5}$.
\end{lemma_2}

\begin{proof}
 Let $z := 0.5\|H(\cdot |\x_0)-H(\cdot |\x_1)\|_{\infty}$ 
 and $y^*$ be such that $|H(y^*|\x_0)-H(y^*|\x_1)| \geq z$.
 It follows from \cref{ass:continuous_hpd} that
 $H(\cdot |\x_0)-H(\cdot |\x_1)$ is Lipschitz continuous and,
 in particular, for every 
 $y \in \left[y^* - z(2M)^{-1}, y^* + z(2M)^{-1} \right]$,
 $|H(y|\x_0)-H(y|\x_1)| \geq z - 2M|y^*-y|$. Therefore,
 there exists $C_1$ such that
 $\|H(\cdot |\x_0)-H(\cdot |\x_1)\|_{2} \geq
  C_1 \|H(\cdot |\x_0)-H(\cdot |\x_1)\|_{\infty}^{1.5}$.
 
 Next, since $|H|$ and $|\hH|$ are bounded by $1$,
 $\|H(\cdot |\x_0)-H(\cdot |\x_1)\|_{1} \leq 2|\sY|$.
 Therefore,
 \begin{align*}
  \|H(\cdot |\x_0)-H(\cdot |\x_1)\|_{2}^2
  &\leq \|H(\cdot |\x_0)-H(\cdot |\x_1)\|_{1} 
  \cdot  \|H(\cdot |\x_0)-H(\cdot |\x_1)\|_{\infty} 
  & \text{Hölders inequality} \\
 &\leq 2|\sY| \cdot \|H(\cdot |\x_0)-H(\cdot |\x_1)\|_{\infty}.
 \end{align*}
 The proof follows directly from
 \cref{ass:bounded}.
\end{proof}

\begin{lemma_2}
 \label{lemma:l2_to_sup}
 Under \cref{ass:iid,ass:bounded,ass:consistent_cde_2,ass:continuous_hpd},
 if $\|\hH(\cdot|\X_i)-\hH(\cdot|\X_{n+1})\|_2 = o(1)$, then
 $$\hH^{-1}(\alpha|\X_i)-H^{-1}(\alpha|\X_{n+1}) = o_\P(1).$$
\end{lemma_2}

\begin{proof}
 It follows from \cref{lemma:hpd_cdf} that
 $H^{-1}(\alpha|\X_i) - \hH^{-1}(\alpha|\X_i) = o_\P(1)$ and,
 therefore, it is sufficient to prove that
 $H^{-1}(\alpha|\X_{i}) - H^{-1}(\alpha|\X_{n+1}) = o_\P(1)$.
 Note that
 \begin{align*}
  \|H(\cdot|\X_{i}) - H(\cdot|\X_{n+1})\|_2
  \leq& \|H(\cdot|\X_{i}) - \hH(\cdot|\X_i) \|_2
  +  \|\hH(\cdot|\X_i) - \hH(\cdot|\X_{n+1}) \|_2 \\
  &+ \|\hH(\cdot|\X_{n+1}) - H(\cdot|\X_{n+1}) \|_2 \\
  \leq& C_2(\|H(\cdot|\X_{i}) - \hH(\cdot|\X_i) \|_\infty^{0.5}
  + \|H(\cdot|\X_{n+1}) - \hH(\cdot|\X_{n+1}) \|_\infty^{0.5}) + o(1)
  & \text{\cref{lemma:l2_equiv_sup}} \\
  &= o_{\P}(1)
  & \text{\cref{lemma:hpd_cdf}}
 \end{align*}
 Conclude from \cref{lemma:l2_equiv_sup} that
 $\|H(\cdot|\X_{i}) - H(\cdot|\X_{n+1})\|_\infty = o_\P(1)$.
 It follows from \cref{ass:continuous_hpd} that
 $H^{-1}(\alpha|\X_{i}) - H^{-1}(\alpha|\X_{n+1}) = o_\P(1)$,
 which completes the proof.
\end{proof}

\begin{proof}[Proof of \cref{thm:cdp_hpd}]
 Let $Z_i = \hf(Y_i|\X_i)$,
 $t = H^{-1}(\alpha|\X_{n+1})$, and
 $M = |\mathcal{T}(\X_{n+1},\mathbb{D})|$.
 It follows from \cref{ass:continuous_hpd} that
 $\P(\X_i \in \mathcal{T}(\X_{n+1},\mathbb{D})) = p > 0$.
 Therefore, $M \sim \text{Bin}(n, p)$.
 Also, it follows from \cref{lemma:l2_to_sup} that,
 for every $\X_i \in A(\x_{n+1})$,
 $H^{-1}(\alpha|\X_{i})-t = o_{\P}(1)$, that is,
 $F^{-1}_{Z_i}(\alpha)-t = o_{\P}(1)$.
 Since $U_{\lfloor\alpha\rfloor}(\X_{n+1})$
 is the $\alpha$-quantile of the empirical cdf of
 $\{Z_i: i \in A(\X_{n+1})\}$,
 conclude from \cref{lemma:quantile} that
 $|U_{\lfloor\alpha\rfloor}(\X_{n+1})-t| = o_{\P}(1)$.
 The rest of the proof follows directly from
 \cref{lemma:cd_hpd,thm:triangle_as_equiv}
\end{proof}

\section*{Appendix B. Additional figures}

\begin{figure}
 \centering
 \includegraphics[width=\figSize]
 {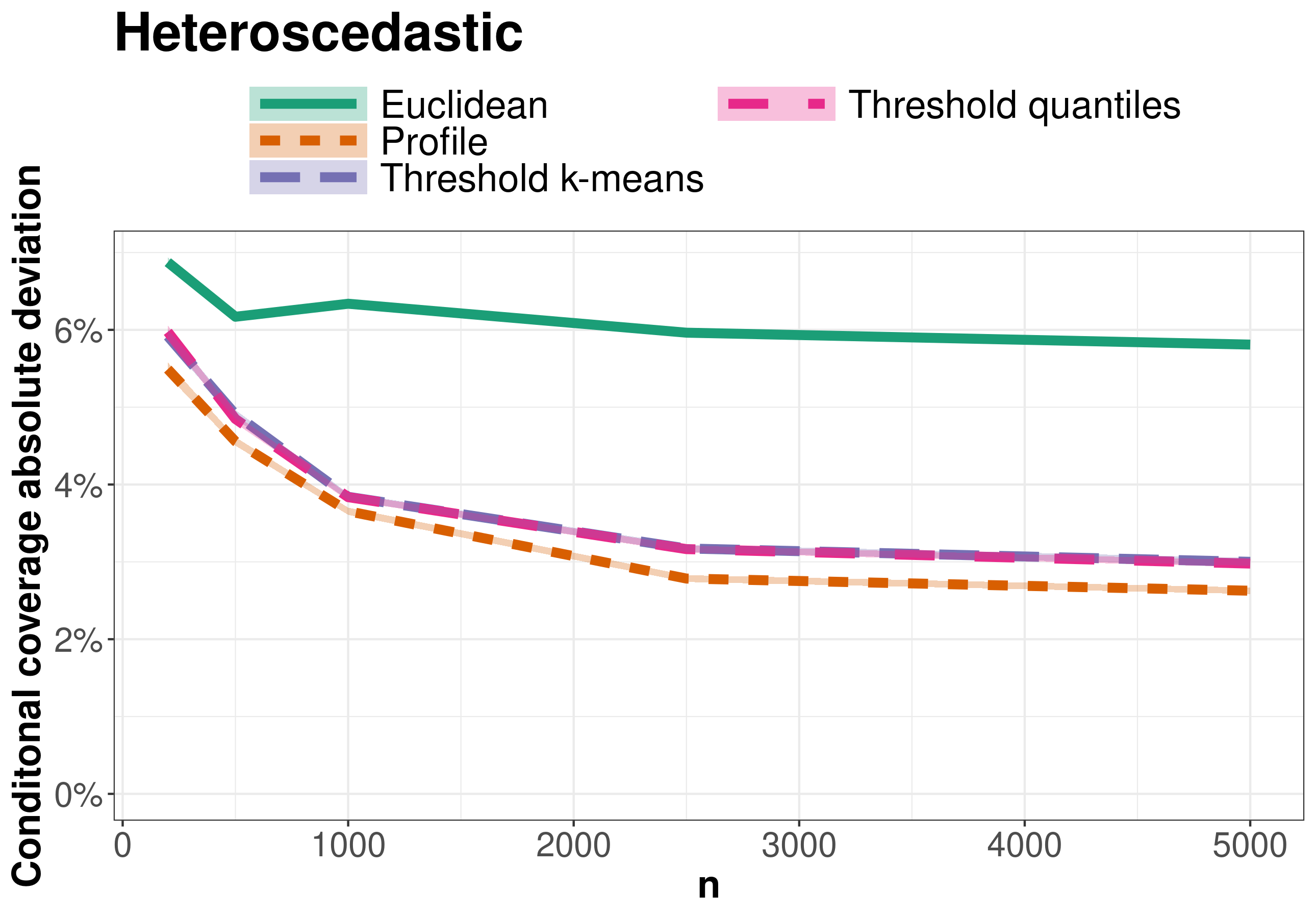} 
 \includegraphics[width=\figSize]
 {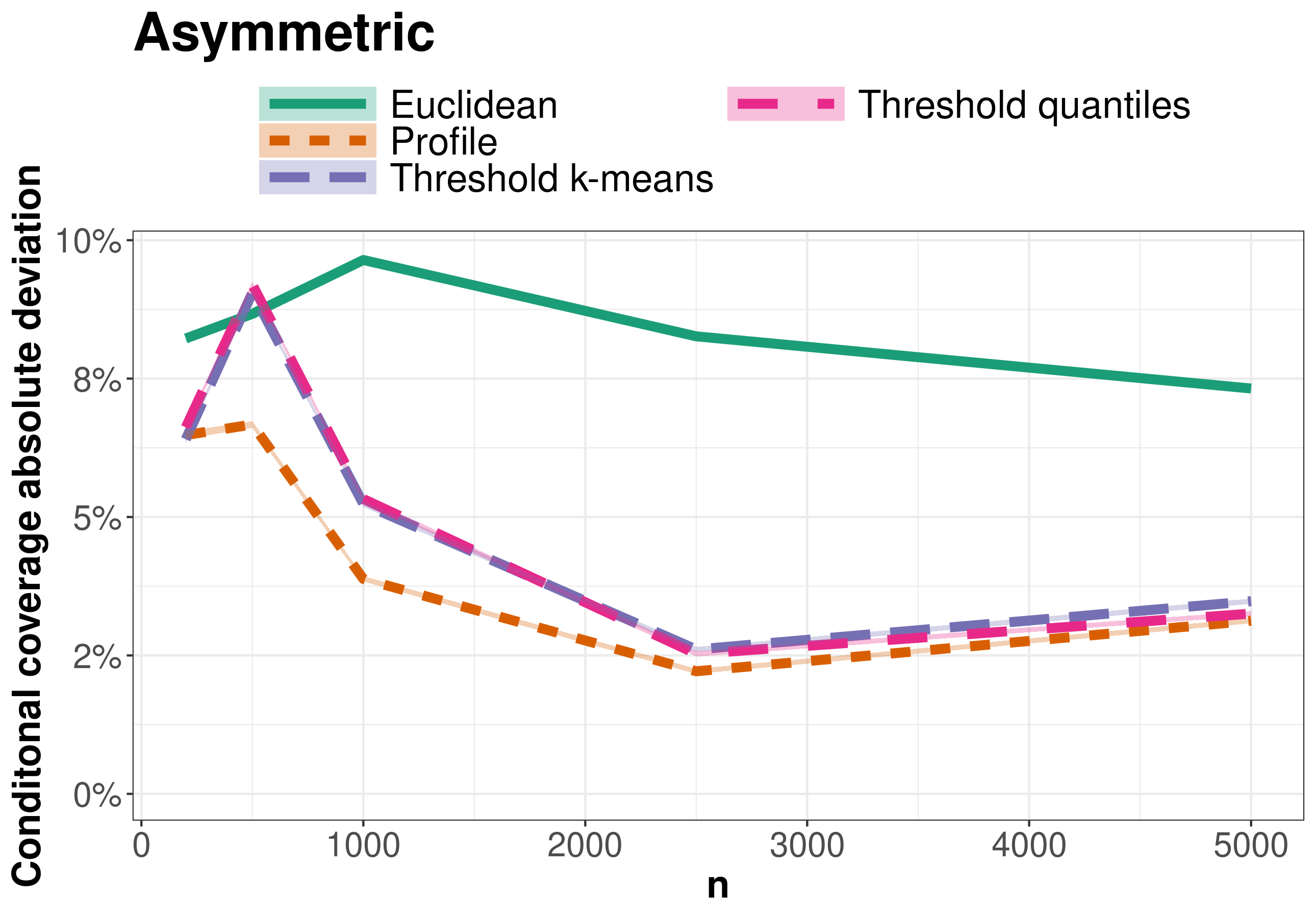}
 \includegraphics[width=\figSize]
 {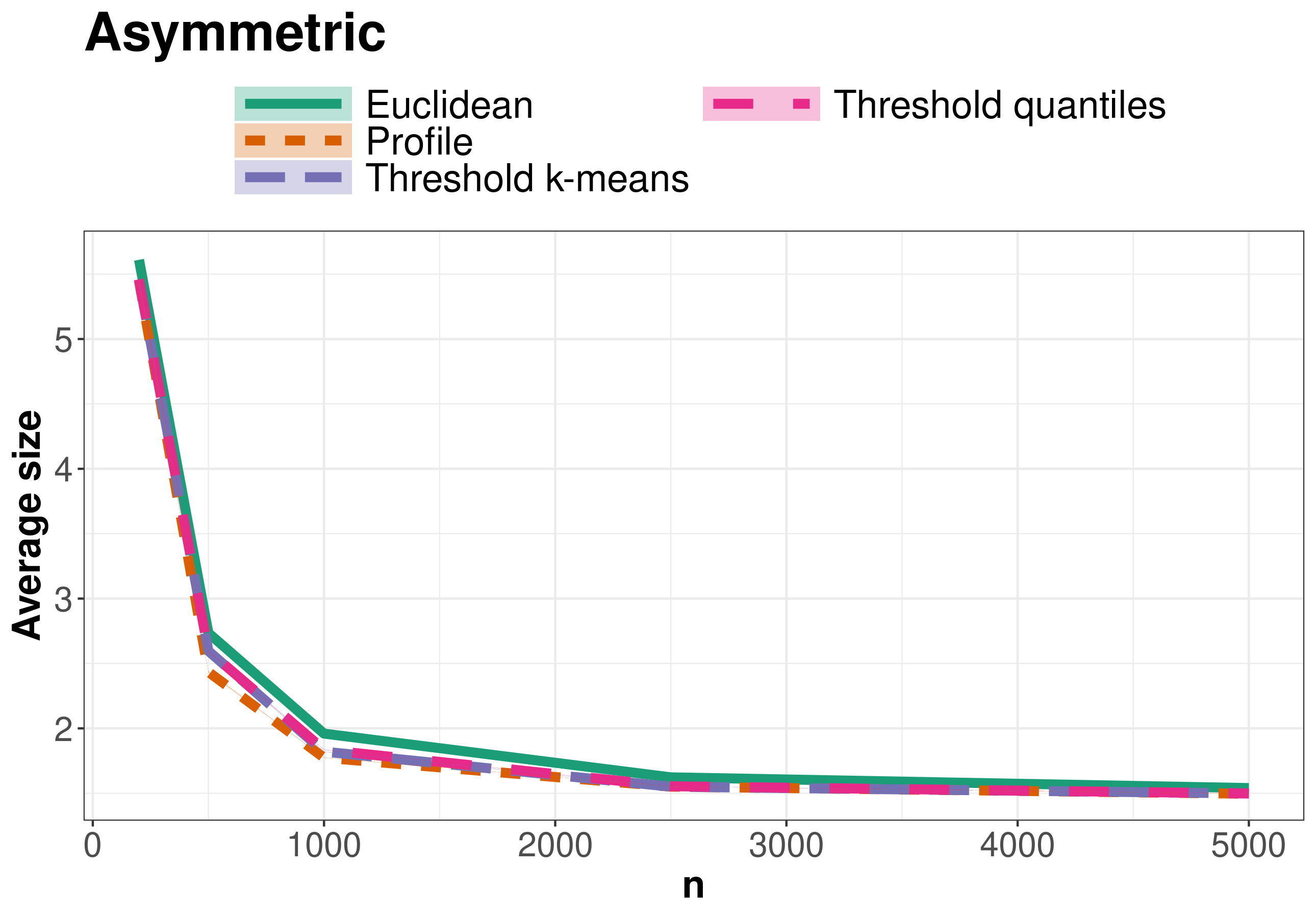}
 \includegraphics[width=\figSize]
 {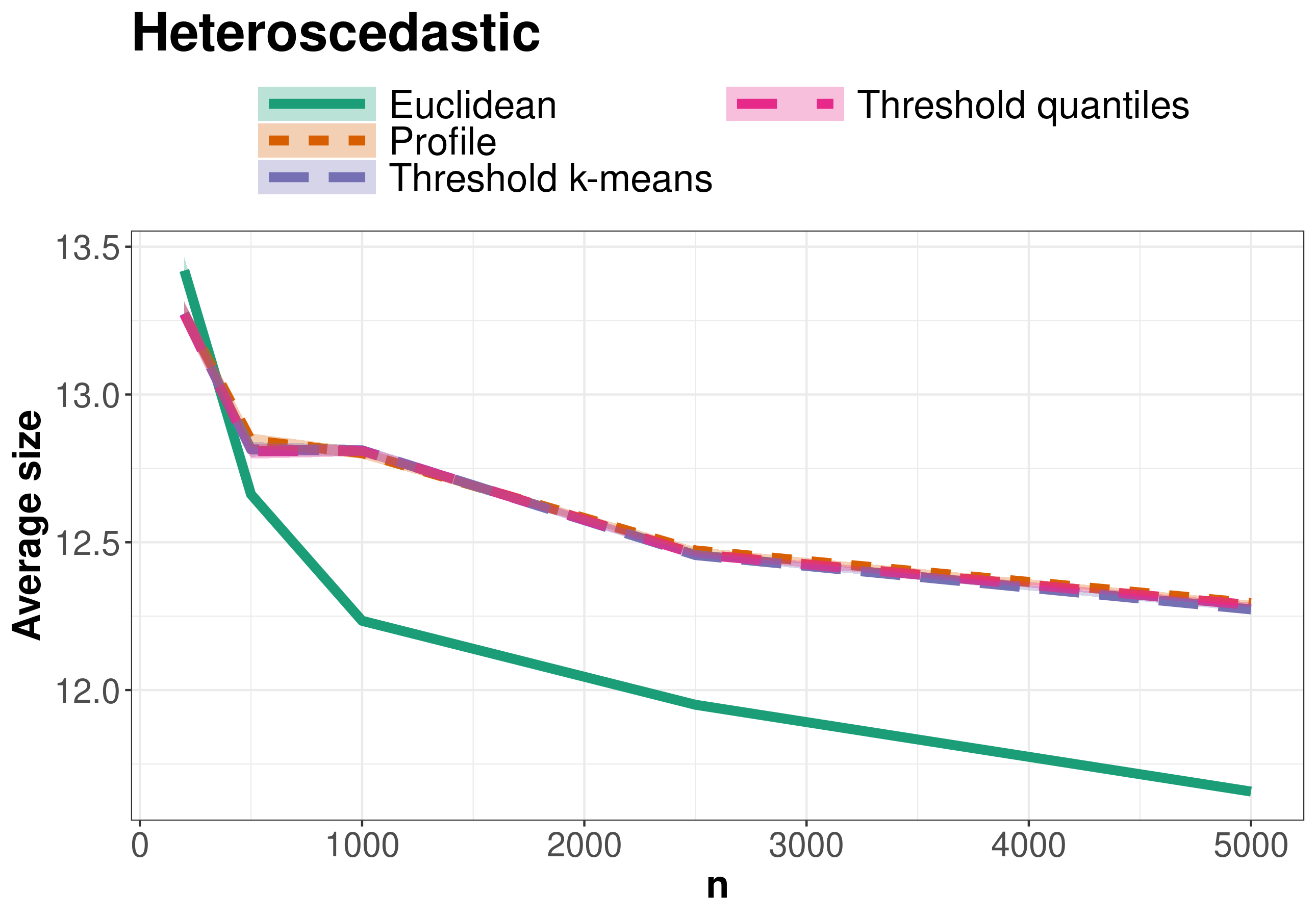}
 \vspace{-2mm}
 \caption{Conditional coverage (top panel) 
 and region size (lower panel) for
 different partitions in \cdsplit.} 
 \label{fig:partition_coverage_size_2}
\end{figure}

\begin{figure}
 \centering
 \includegraphics[width=\figSize]
 {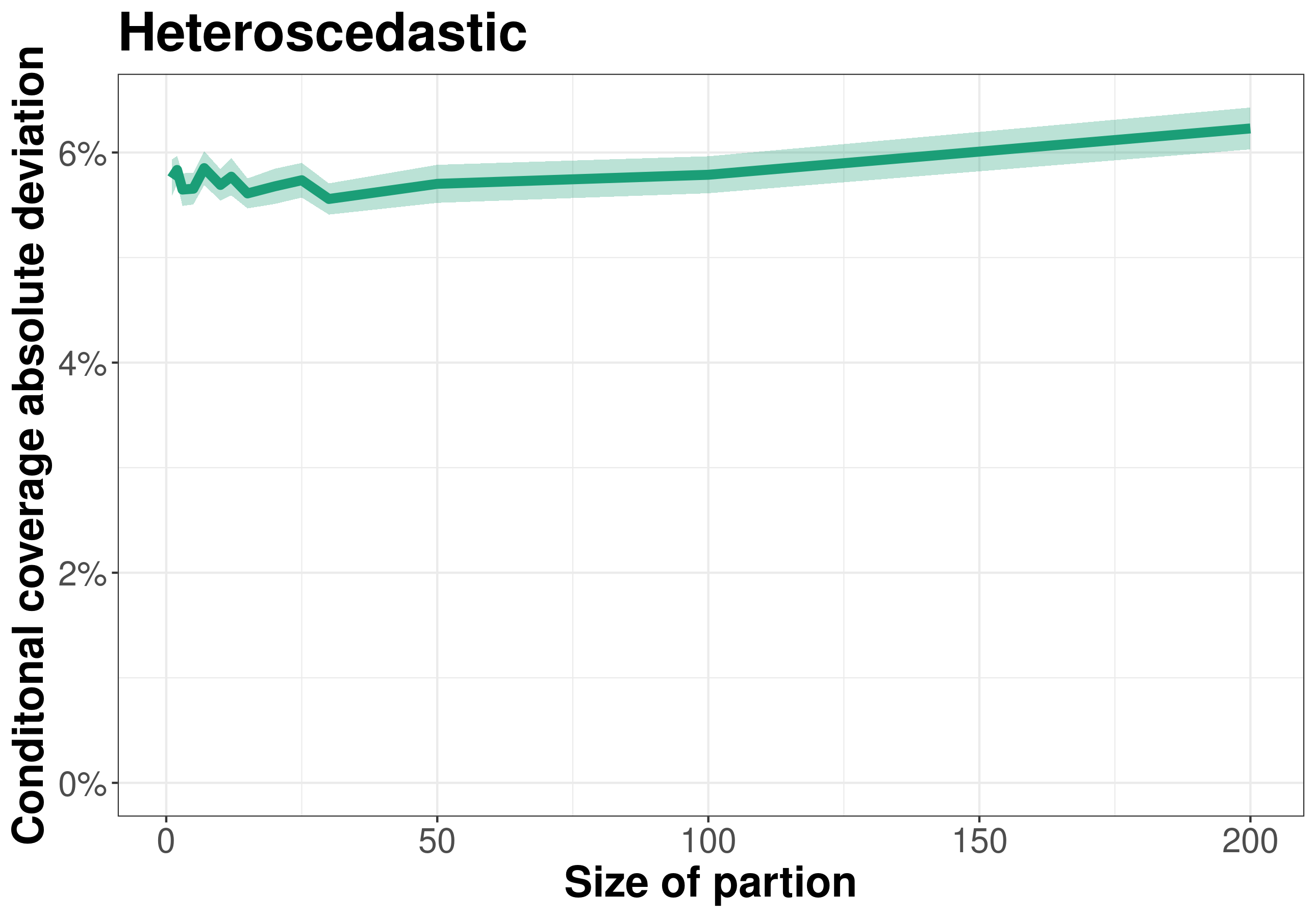}
 \hspace{2mm}
 \includegraphics[width=\figSize]
 {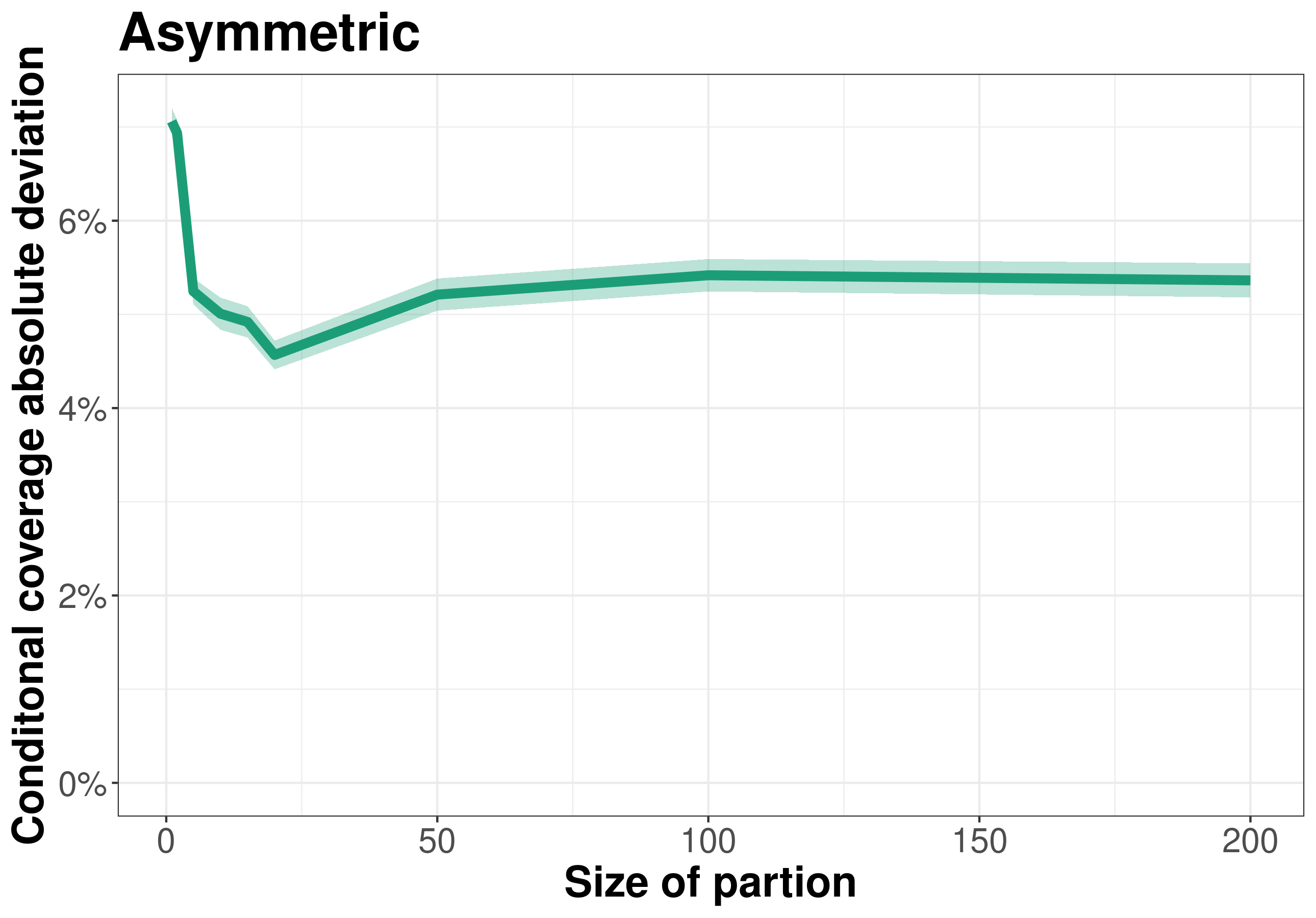} \\
 \includegraphics[width=\figSize]
 {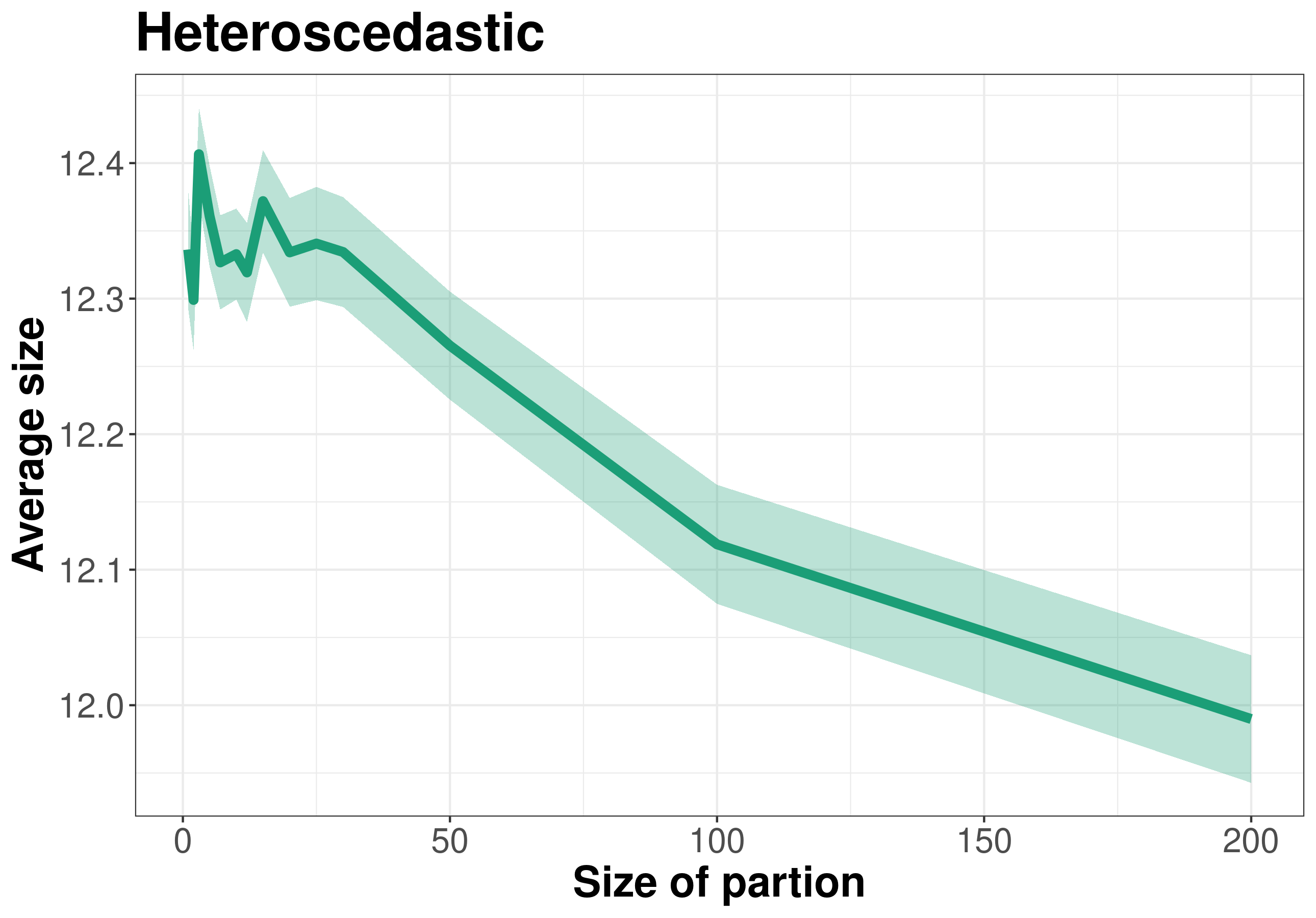}
 \hspace{2mm}
 \includegraphics[width=\figSize]
 {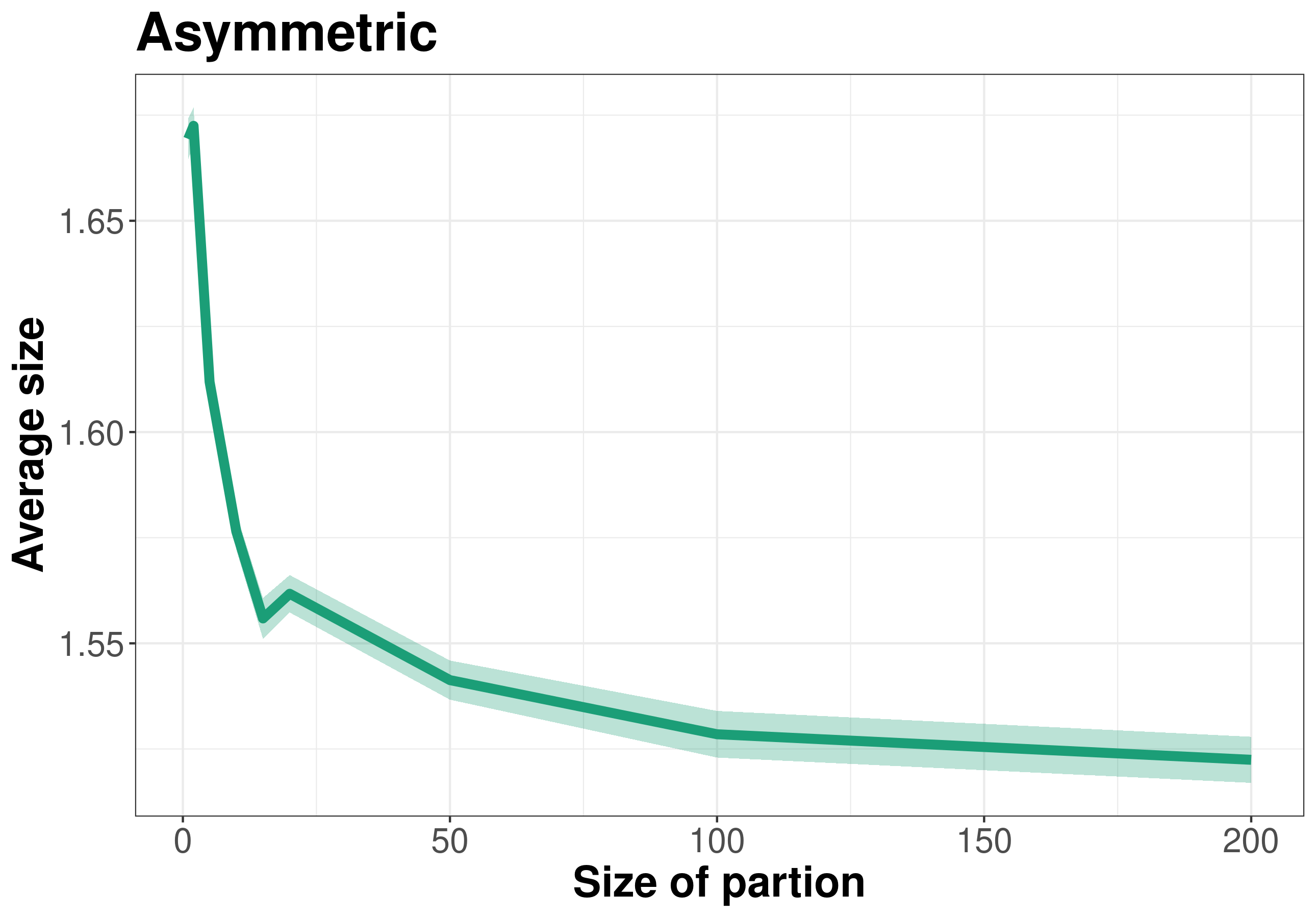}
 \vspace{-2mm}
 \caption{Conditional coverage (top panel) and
 predictive region size (bottom panel) for
 different partition sizes in \cdsplitp .} 
\label{fig:k_2} 
\end{figure}

\begin{figure}
 \centering
 \includegraphics[width=\figSize]
 {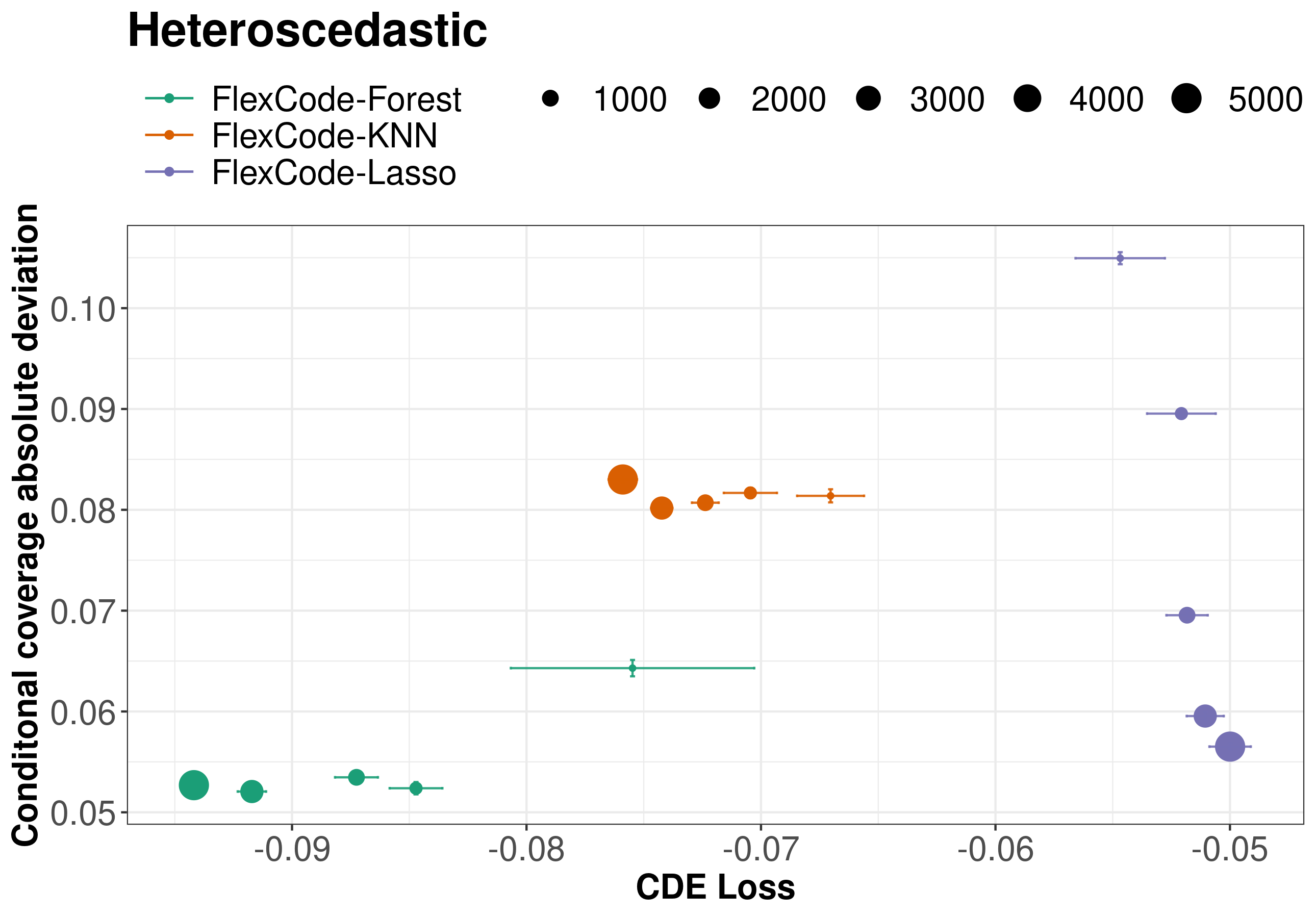}
 \hspace{2mm}
 \includegraphics[width=\figSize]
 {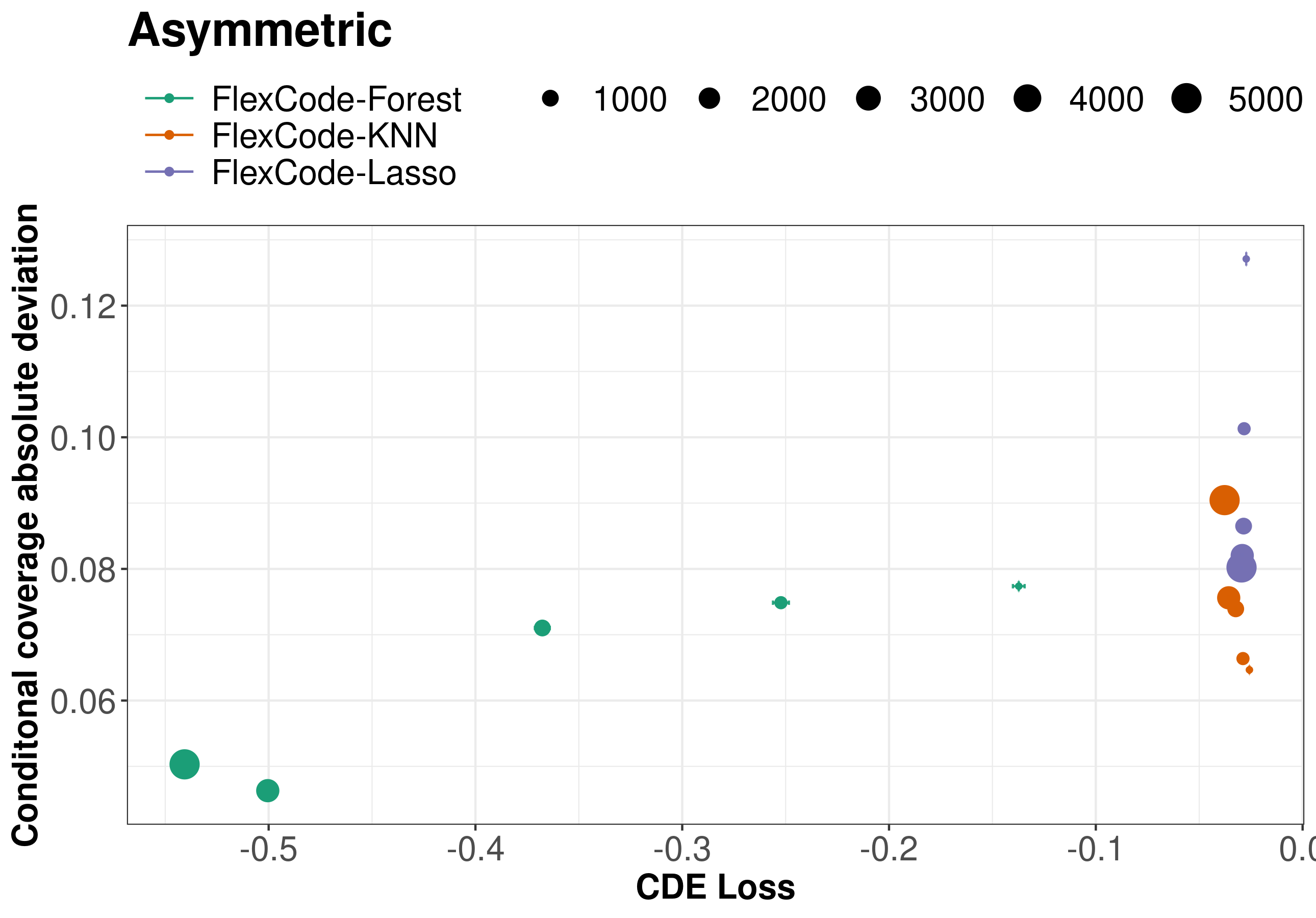} \\
 \includegraphics[width=\figSize]
 {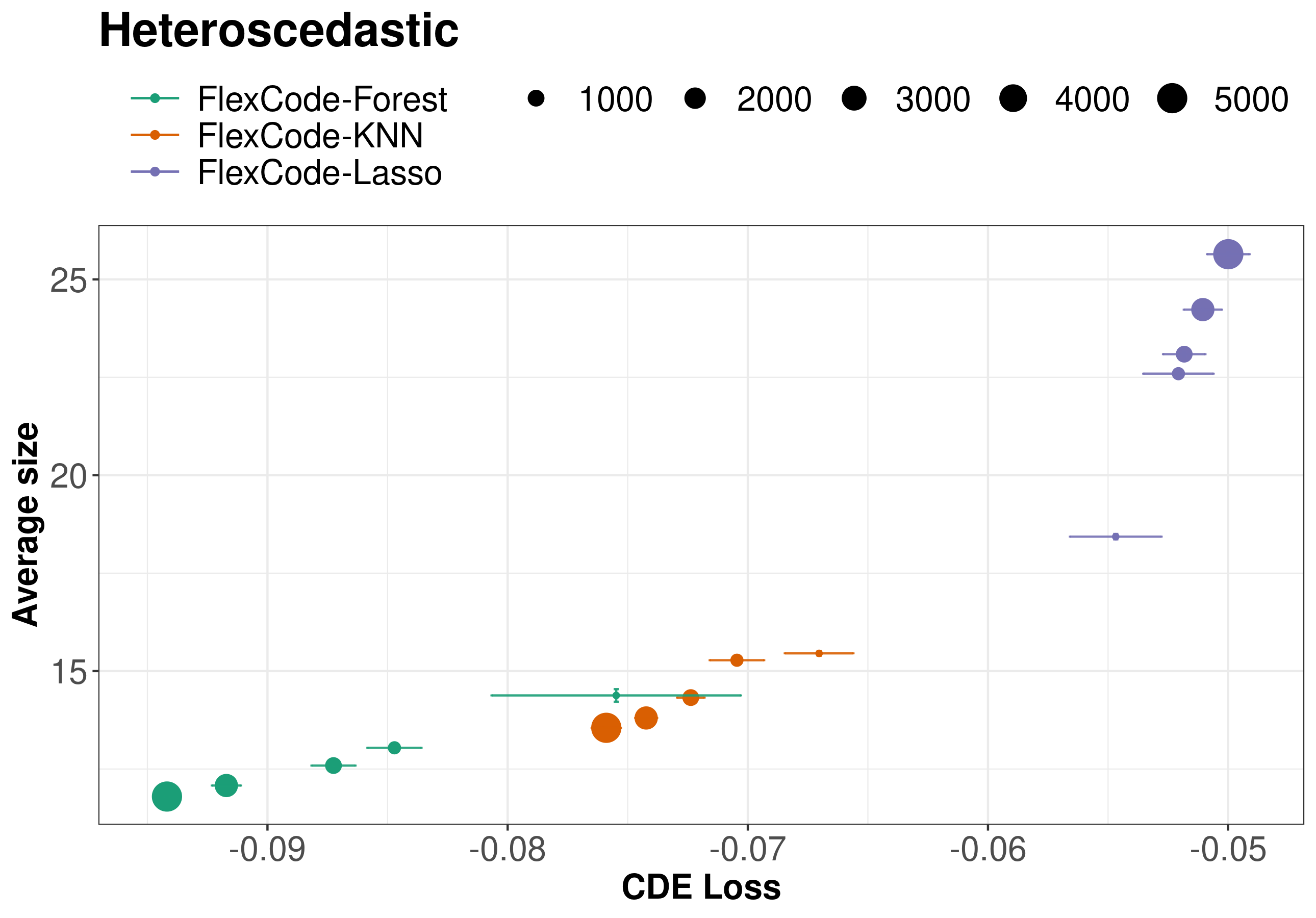}
 \hspace{2mm}
 \includegraphics[width=\figSize]
 {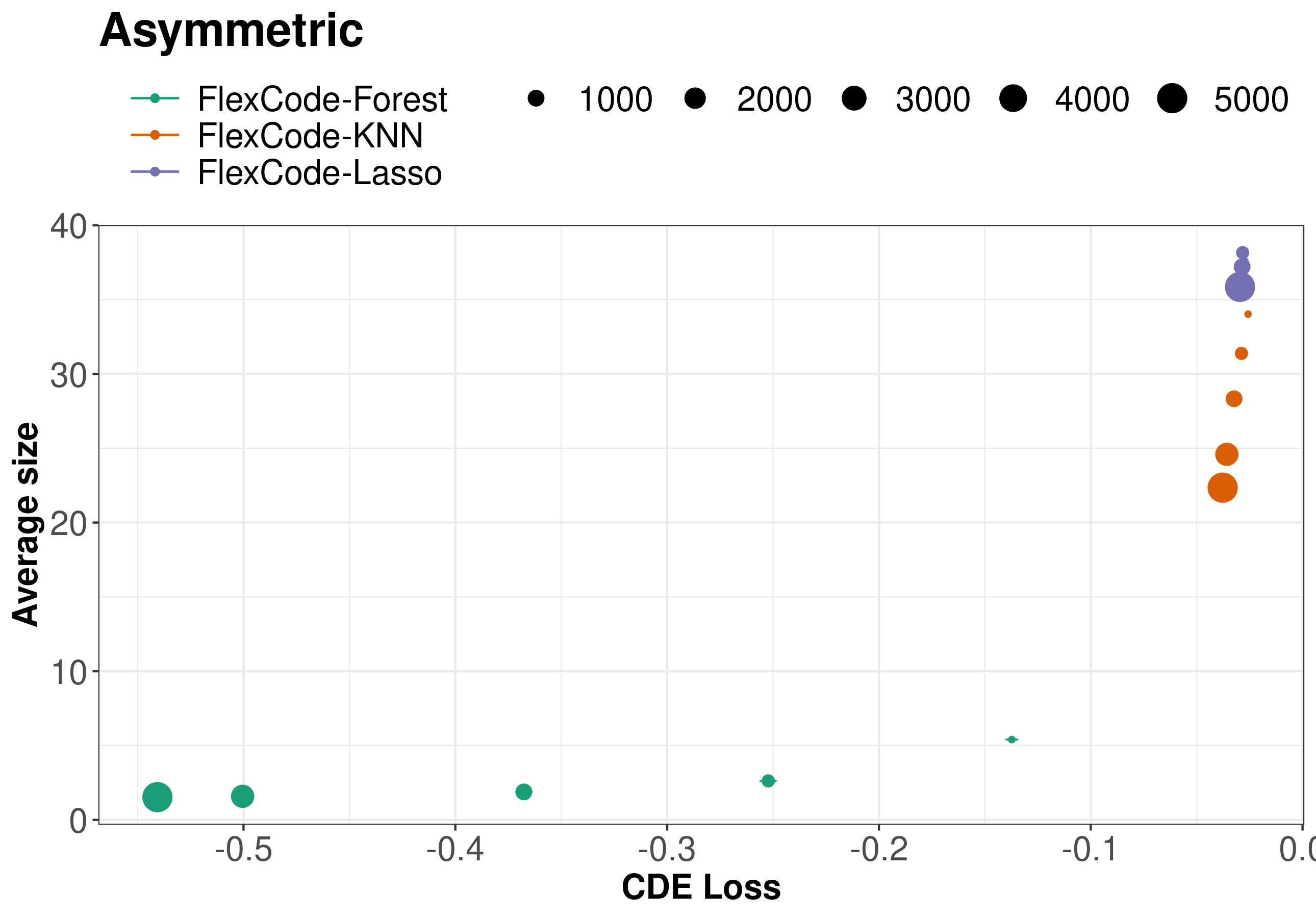}
 \vspace{-2mm}
 \caption{Performance of the \cdsplitp \ with
 respect to conditional coverage (upper panel)
 and region size (lower panel)
 as a function of the estimated 
 conditional density loss (CDE loss).
 Each point is a different combination of 
 sample size and estimator.} 
 \label{fig:regression_th_2}
\end{figure}

\begin{figure}
 \centering
 \includegraphics[scale=0.25]{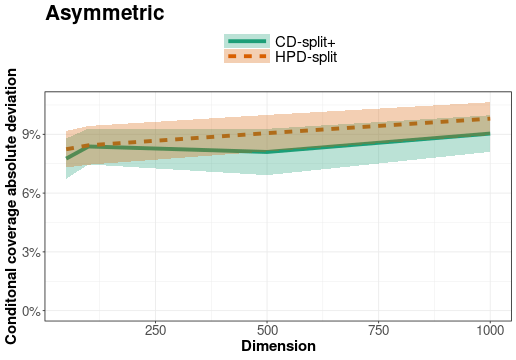}
 \hspace{2mm}
 \includegraphics[scale=0.25]{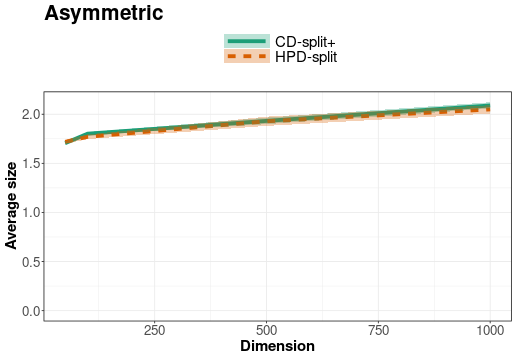} \\ [2mm]
 \includegraphics[scale=0.25]{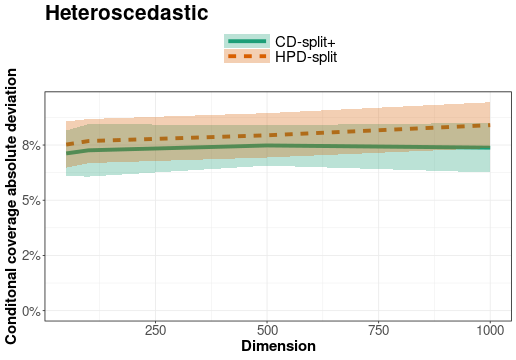}
 \hspace{2mm} 
 \includegraphics[scale=0.25]{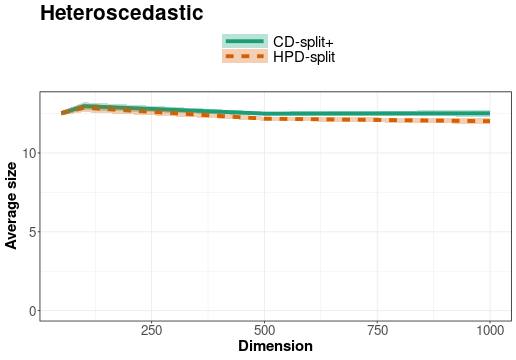} \\ [2mm]
 \vspace{-2mm} 
 \caption{Conditional coverage (left panel)
 and average size of prediction bands (right panel)
 for \cdsplitp\ and \hpdsplit \
 as a function of the number of features. None of the methods are heavily affected by increasing dimensionality.} 
 \label{fig:dimension_2}
\end{figure}

\end{document}